\documentclass{article} 

\usepackage{iclr2024_conference,times}
\iclrfinaltrue

\usepackage{fancyhdr}
\ificlrfinal
    \ifpreprint
    \else
    \fi
\else
\fi
\usepackage{wrapfig}
\usepackage{xcolor}
\usepackage[utf8]{inputenc}
\usepackage{epsfig}
\usepackage{graphicx}
\usepackage{amsmath,amsfonts,amssymb,amsthm}
\usepackage{color}
\usepackage{subcaption}
\usepackage{xspace}
\usepackage{array}
\usepackage{cite}
\usepackage{bm}
\usepackage[pagebackref=true,breaklinks=true,colorlinks,bookmarks=false]{hyperref}
\usepackage{multirow}
\usepackage{booktabs}
\usepackage{cancel}
\usepackage{bbm}

\usepackage[figurename=Fig.]{caption}

\usepackage[toc,page,header]{appendix}
\usepackage{minitoc}

\usepackage{algorithm}
\usepackage{algorithmic}

\usepackage{enumitem}
\setlist{leftmargin=5.5mm}

\newenvironment{tight_itemize}{
\begin{itemize}
  \setlength{\itemsep}{0pt}
  \setlength{\parskip}{0pt}
  \setlength{\topsep}{0pt}
  \setlength{\partopsep}{0pt}
}{\end{itemize}}

\newenvironment{tight_enumerate}{
\begin{enumerate}
  \setlength{\itemsep}{0pt}
  \setlength{\parskip}{0pt}
  \setlength{\topsep}{0pt}
  \setlength{\partopsep}{0pt}
}{\end{enumerate}}

\makeatletter
\DeclareRobustCommand\onedot{\futurelet\@let@token\@onedot}
\def\@onedot{\ifx\@let@token.\else.\null\fi\xspace}
 \def\Eg{E.g\onedot}
\def\ie{i.e\onedot}

\makeatother

\newcommand{\figref}[1]{Fig.~\ref{#1}}
\newcommand{\secref}[1]{Section~\ref{#1}}

\newcommand{\algref}[1]{Algorithm~\ref{#1}}
\newcommand{\eqnref}[1]{Eq.~\eqref{#1}}

\newcommand{\tabref}[1]{Table~\ref{#1}}

\newcommand{\boldparagraph}[1]{\vspace{0.0cm}\noindent{\bf #1:} }

\definecolor{darkgreen}{rgb}{0,0.7,0}
\definecolor{darkblue}{RGB}{31,119,180}
\definecolor{darkred}{RGB}{214,39,40}
\definecolor{mediumgray}{rgb}{0.5,0.5,0.5}
\definecolor{mediumteal}{rgb}{0,0.5,0.5}

\definecolor{ellisred}{rgb}{0.87,0.44,0.38} %
\definecolor{ellisgreen}{rgb}{0.69,0.90,0.52} %
\definecolor{elliscyan}{rgb}{0.29,0.77,0.74} %
\definecolor{ellisorange}{rgb}{0.89,0.55,0.28} %
\definecolor{ellisblue}{rgb}{0.41,0.61,0.86} %

\title{GTA: A Geometry-Aware Attention Mechanism for Multi-View Transformers}

\author{
Takeru Miyato\textsuperscript{1}, Bernhard Jaeger\textsuperscript{1}, Max Welling\textsuperscript{2}, Andreas Geiger\textsuperscript{1} \\
\textsuperscript{1} University of Tübingen, Tübingen AI Center~ \textsuperscript{2} University of Amsterdam \\
}

\newcommand{\brhog}{{\bf P_{\bf g}}}

\newcommand\blfootnote[1]{%
  \begingroup
  \renewcommand\thefootnote{}\footnote{#1}%
  \addtocounter{footnote}{-1}%
  \endgroup
}

\begin{document}
\doparttoc 
\faketableofcontents 

\maketitle
\vspace{-3mm}
\begin{abstract}
As transformers are equivariant to the permutation of input tokens, encoding the positional information of tokens is necessary for many tasks. However, since existing positional encoding schemes have been initially designed for NLP tasks, their suitability for vision tasks, which typically exhibit different structural properties in their data, is questionable. We argue that existing positional encoding schemes are suboptimal for 3D vision tasks, as they do not respect their underlying 3D geometric structure. Based on this hypothesis, we propose a geometry-aware attention mechanism that encodes the geometric structure of tokens as \textit{relative transformation} determined by the geometric relationship between queries and key-value pairs. By evaluating on multiple novel view synthesis (NVS) datasets in the sparse wide-baseline multi-view setting, we show that our attention, called \textit{Geometric Transform Attention (GTA)}, improves learning efficiency and performance of state-of-the-art transformer-based NVS models without any additional learned parameters and only minor computational overhead.
\ificlrfinal
\blfootnote{
Correspondence to \texttt{takeru.miyato@gmail.com}. Code: \url{https://github.com/autonomousvision/gta}.}
\fi
\end{abstract}

\section{Introduction}
\vspace{-1mm}

The transformer model~\citep{Vaswani2017NEURIPS}, which is composed of a stack of permutation symmetric layers, processes input tokens as a \textit{set} and lacks direct awareness of the tokens' structural information. Consequently, transformer models are not solely perceptible to the structures of input tokens, such as word order in NLP or 2D positions of image pixels or patches in image processing.

A common way to make transformers position-aware is through vector embeddings: in NLP, a typical way is to transform the position values of the word tokens into embedding vectors to be added to input tokens or attention weights~\citep{Vaswani2017NEURIPS, Shaw2018NAACL}. While initially designed for NLP, these positional encoding techniques are widely used for 2D and 3D vision tasks today~\citep{Wang2018CVPRd, Dosovitskiy2021ICLR, Sajjadi2022CVPR, Du2023CVPR}.

Here, a natural question arises: ``Are existing encoding schemes suitable for tasks with very different geometric structures?". Consider for example 3D vision tasks using multi-view images paired with camera transformations. The 3D Euclidean symmetry behind multi-view images is a more intricate structure than the 1D sequence of words. With the typical vector embedding approach, the model is tasked with uncovering useful camera poses embedded in the tokens and consequently struggles to understand the effect of non-commutative Euclidean transformations.

Our aim is to seek a principled way to incorporate the geometrical structure of the tokens into the transformer. To this end, we introduce a method that encodes the token relationships as transformations directly within the attention mechanism. 
More specifically, we exploit the relative transformation determined by the geometric relation between the query and the key-value tokens. We then apply those transformations to the key-value pairs, which allows the model to compute QKV attention in an aligned coordinate space.

We evaluate the proposed attention mechanism on several novel view synthesis (NVS) tasks with \textit{sparse and wide-baseline} multi-view settings, which are particularly hard tasks where a model needs to learn \textit{strong 3D geometric priors} from multiple training scenes.
We show that existing positional encoding schemes are suboptimal and that our geometric-aware attention, named \textit{geometric transform attention (GTA)}, significantly improves learning efficiency and performance of state-of-the-art transformer-based NVS models, just by replacing the existing positional encodings with GTA.

\section{Related work} \label{sec:background}
Given token features $X\in \mathbb{R}^{n\times d}$, the attention layer's outputs  $O \in \mathbb{R}^{n\times d}$ are computed as follows:
\begin{align}
    O  := {\rm Attn}(Q, K, V) = {\rm softmax}(Q K^{\rm T})V,
    \label{eq:vatt}
\end{align}
where $Q,K,V=XW^Q,XW^K,XW^V\in \mathbb{R}^{n\times d}, W^{\{Q, K, V\}}\in \mathbb{R}^{d\times d}$, and $(n, d)$ is the number of tokens and channel dimensions. We omit the scale factor inside the softmax function for simplicity. 
The output in \eqnref{eq:vatt} is invariant to the permutation of the key-value vector indices. 
To break this permutation symmetry, we explicitly encode positional information into the transformer, which is called positional encoding (PE).
The original transformer~\citep{Vaswani2017NEURIPS} incorporates positional information by adding embeddings to all input tokens. 
This \textit{absolute positional encoding} (APE) scheme has the following form:
\begin{align}
     {\rm softmax}\left((Q+\gamma({\bf P})W^Q)(K+\gamma({\bf P})W^K)^{\rm T}\right) \left(V+\gamma({\bf P})W^V\right), \label{eq:ape} 
\end{align}
where ${\bf P}$ denotes the positional attributes of the tokens $X$ and $\gamma$ is a PE function. From here, a bold symbol signifies that the corresponding variable consists of a list of elements.
$\gamma$ is typically the sinusoidal function, which transforms position values into Fourier features with multiple frequencies.
\citet{Shaw2018NAACL} proposes an alternative PE method, encoding the relative distance between each pair of query and key-value tokens as biases added to each component of the attention operation:
\begin{align}
    {\rm softmax}\left(QK^{\rm T} + \gamma_{\rm rel}({\bf P})\right)(V+\gamma'_{\rm rel}({\bf P})), \label{eq:rpe}
\end{align}
where $\gamma_{\rm rel}({\bf P})\in \mathbb{R}^{n\times n}$ and $\gamma'_{\rm rel}({\bf P})\in \mathbb{R}^{n\times d}$ are the bias terms that depend on the distance between tokens. This encoding scheme is called \textit{relative positional encoding}~(RPE) and ensures that the embeddings do not rely on the sequence length, with the aim of improving length generalization.

Following the success in NLP, transformers have demonstrated their efficacy on various image-based computer vision tasks ~\citep{Wang2018CVPRd, Ramachandran2019NEURIPS, Carion2020ECCV, Dosovitskiy2021ICLR, Ranftl2021ICCV, Romero2020ICML, Wu2021CVPRb, Chitta2022PAMI}. Those works use variants of APE or RPE applied to 2D positional information to make the model aware of 2D image structure. Implementation details vary across studies. 
Besides 2D-vision, there has been a surge of application of transformer-based models to 3D-vision~\citep{Wang2021CVPR, Liu2022ECCV, Kulhanek2022ECCV, Sajjadi2022CVPR, Watson2023ICLR, Varma2023ICLR, Xu2023PAMI, Shao2023CORL, Venkat2023ARXIV, Du2023CVPR, Liu2023ARXIV}. 

Various PE schemes have been proposed in 3D vision, mostly relying on APE- or RPE-based encodings.
In NVS \citet{Kulhanek2022ECCV, Watson2023ICLR, Du2023CVPR} embed the camera extrinsic information by adding linearly transformed, flattened camera extrinsic matrices to the tokens. In \citet{Sajjadi2022CVPR, Safin2023ARXIV}, camera extrinsic and intrinsic information is encoded through ray embeddings that are added or concatenated to tokens. \citet{Venkat2023ARXIV} also uses ray information and biases the attention matrix by the ray distance computed from ray information linked to each pair of query and key tokens.
An additional challenge in 3D detection and segmentation is that the output is typically in an orthographic camera grid, differing from the perspective camera inputs. Additionally, sparse attention \citep{Zhu2021ICLR} is often required because high resolution feature grids \citep{Lin2017CVPRb} are used. \citet{Wang2021CORL, Li2022ECCV} use learnable PE for the queries and no PE for keys and values. \citet{Peng2023WACV} find that using standard learnable PE for each camera does not improve performance when using deformable attention.
\citet{Liu2022ECCV, Liu2023ICCV} do add PE to keys and values by generating 3D points at multiple depths for each pixel and adding the points to the image features after encoding them with an MLP. 
\citet{Zhou2022CVPR} learn positional embeddings using camera parameters and apply them to the queries and keys in a way that mimics the relationship between camera and target world coordinates.
\citet{Shu20233ICCV} improves performance by using available depths to link image tokens with their 3D positions.
Besides APE and RPE approaches, \citet{Hong2023ARXIV, Zou2023ARXIV, Wang2023ARXIV} modulate tokens by FiLM-based approach~\citep{Perez2018AAAI}, where they element-wise multiply tokens with features computed from camera transformation.

In point cloud transformers, \citet{Yu2021ICCV} uses APE to encode 3D positions of point clouds. ~\citet{Qin2022CVPR} uses an RPE-based attention mechanism, using the distance or angular difference between tokens as geometric information.
Epipolar-based sampling techniques are used to sample geometrically relevant tokens of input views in attention layers~\citep{He2020CVPR, Suhail2022ECCV, Saha2022ICRA, Varma2023ICLR, Du2023CVPR}, where key and value tokens are sampled along an epipolar line determined by the camera parameters between a target view and an input view. 

%
%
\section{Geometric encoding by relative transformation}\label{sec:geo_enc}

In this work, we focus on novel view synthesis (NVS), which is a fundamental task in 3D-vision. The NVS task is to predict an image from a novel viewpoint, given a set of context views of a scene and their viewpoint information represented as $4\times 4$ extrinsic matrices, each of which maps 3D points in world coordinates to the respective points in camera coordinates. NVS tasks require the model to understand the scene geometry directly from raw image inputs. 

The main problem in existing encoding schemes of the camera transformation is that they do not respect the geometric structure of the Euclidean transformations.
In \eqnref{eq:ape} and \eqnref{eq:rpe}, the embedding is added to each token or to the attention matrix. However, the geometry behind multi-view images is governed by Euclidean symmetry. When the viewpoint changes, the change of the object's pose in the camera coordinates is computed based on the corresponding camera transformation.

Our proposed method incorporates geometric transformations directly into the transformer's attention mechanism through a \textit{relative transformation} of the QKV features. Specifically, each key-value token is transformed by a {\rm relative transformation} that is determined by the geometric attributes between query and key-value tokens. This can be viewed as a coordinate system alignment, which has an analogy in geometric processing in computer vision: when comparing two sets of points each represented in a different camera coordinate space, we move one of the sets using a relative transformation $c c'^{-1}$ to obtain all points represented in the same coordinate space. Here, $c$ and $c'$ are the extrinsics of the respective point sets. Our attention performs this coordinate alignment within the \textit{attention feature space}. This alignment allows the model not only to compare query and key vectors in the same reference coordinate space, but also to perform the addition of the attention output at the residual path in the aligned local coordinates of each token due to the value vector's transformation.

This direct application of the transformations to the attention features shares its philosophy with the classic transforming autoencoder~\citep{Hinton2011ICANN, Cohen2014ICLR, Worrall2017ICCV, Rhodin2018ECCV, Falorsi2018ARXIV, Chen2019CVPRc, Dupont2020ICML}, capsule neural networks~\citep{Sabour2017NEURIPS, Hinton2018ICLR}, and equivariant representation learning models~\citep{Park2022ICML, Miyato2022NEURIPS, Koyama2023ARXIV}. In these works, geometric information is provided as a transformation applied to latent variables of neural networks. Suppose $\Phi(x)$ is an encoded feature, where $\Phi$ is a neural network, $x$ is an input feature, and $\mathcal{M}$ is an associated transformation (e.g. rotation). Then the pair ($\Phi(x)$, $\mathcal{M}$) is identified with $\mathcal{M} \Phi(x)$. We integrate this feature transformation into the attention to break its permutation symmetry.

\label{sec:rta}

\boldparagraph{Group and representation} 
We briefly introduce the notion of a \textit{group} and a \textit{representation} because we describe our proposed attention through the language of group theory, which handles different geometric structures in a unified manner, such as camera transformations and image positions. 
In short, a group $G$ with its element $g$, is an associative set that is closed under multiplication, has the identity element and each element has an inverse. \Eg the set of camera transformations satisfies the axiom of a group and is called \textit{special Euclidean group}: $SE(3)$. 
A (real) \textit{representation} is a function $\rho: G \rightarrow GL_d(\mathbb{R})$ such that $\rho(g)\rho(g') = \rho(g g')$ for any $g, g' \in G$. The property $\rho(g)\rho(g') = \rho(g g')$ is called \textit{homomorphism}.
Here, $GL_d(\mathbb{R})$ denotes the set of $d \times d$ invertible real-valued matrices.
We denote by $\rho_g := \rho(g) \in \mathbb{R}^{d\times d}$ a representation of $g$. 
A simple choice for the representation $\rho_g$ for $g\in SE(3)$ is a $4\times4$ rigid transformation matrix $\left[\begin{smallmatrix} R & T \\ 0 & 1 \end{smallmatrix}\right] \in \mathbb{R}^{4\times 4}$ where $R \in \mathbb{R}^{3\times 3}$ is a 3D rotation and $T\in \mathbb{R}^{3 \times 1}$ is a 3D translation. A block concatenation of multiple group representations is also a representation. What representation to use is the user's choice. We will present different design choices of $\rho$ for several NVS applications in \secref{sec:gta}, \ref{sec:rho_design} and \ref{sec:kroneckergta}.
\subsection{Geometric transform attention}\label{sec:gta}
Suppose that we have token features $X\in \mathbb{R}^{n\times d}$ and a list of geometric attributes ${\bf g} = [g_1, \dots, g_n]$, where $g_i$ is an $i$-th token's geometric attribute represented as a group element. For example, each $X_i\in \mathbb{R}^d$ corresponds to a patch feature, and $g_i$ corresponds to a camera transformation and an image patch position. Given a representation $\rho$ and $Q, K, V = XW^Q, XW^K, XW^V\in\mathbb{R}^{n\times d}$, we define our geometry-aware attention given query $Q_i\in \mathbb{R}^d$ by:
\begin{align}
O_i =&\sum_j^{n} \frac{\exp ({Q_i^{\rm T} (\rho_{g_i g_j^{-1}} K_j))}}{\sum_{j'=1}^{n} \exp(Q_i^{\rm T} (\rho_{g_i g_{j'}^{-1}} K_{j'}))} (\rho_{g_i g_j^{-1}} V_{j}), \label{eq:rta}
\end{align}
Using the homomorphism property $\rho_{g_i g_j^{-1}} = \rho_{g_i} \rho_{g_j^{-1}}$, the above equation can be transformed into
\begin{align}
O_i =&\rho_{g_i} \sum_j^{n} \frac{\exp ((\rho_{g_i}^{\rm T} Q_i)^{\rm T} (\rho_{g_j^{-1}} K_j))}{\sum_{j'=1}^{n} \exp((\rho_{g_i}^{\rm T} Q_i)^{\rm T} (\rho_{g_{j'}^{-1}} K_{j'}))} (\rho_{g_j^{-1}} V_{j}). \label{eq:rta_2}
\end{align}
Note that the latter expression is computationally and memory-wise more efficient, requiring computation and storage of $n^2$ values of each ($\rho_{g_i g_j^{-1}} K_j$, $\rho_{g_i g_j^{-1}} V_j$) in \eqnref{eq:rta} versus only $n$ values for ($\rho_{g_i}^{\rm T} Q_i$, $\rho_{g_j}^{-1} K_j$, $\rho_{g_j}^{-1} V_j$) and $\rho_{g_i} \hat{O}_i$ in \eqnref{eq:rta_2}, where $\hat{O}_i$ is the output of the leftmost sum.
\begin{wrapfigure}{r}{0.26\textwidth}
    \vspace{-2mm}
    \centering
    \includegraphics[scale=0.32]{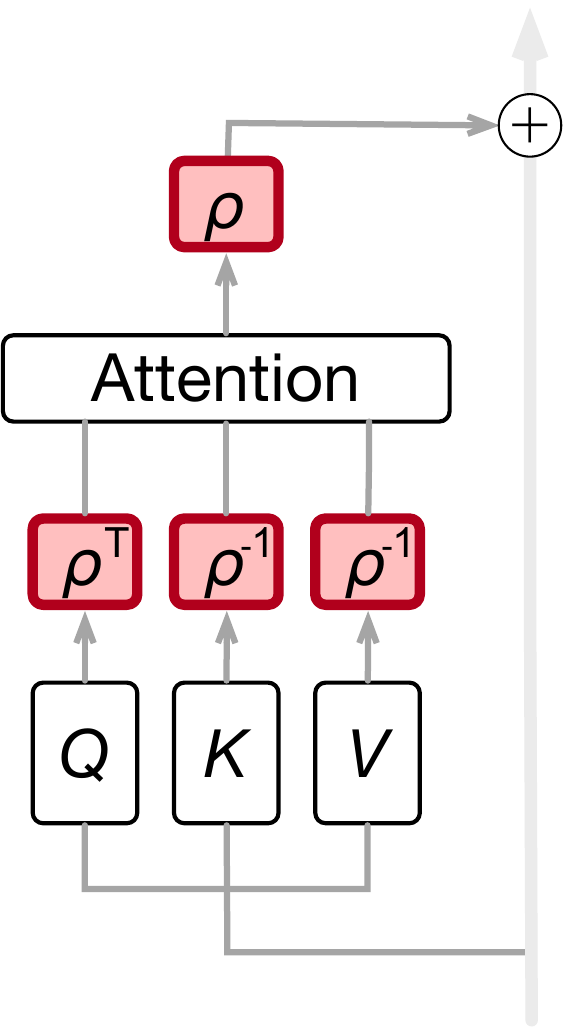}
    \caption{\textbf{GTA mechanism.} $\rho^{-1}$ and $\rho^{\rm T}$ together take $Q,K$ and $V$ to a shared coordinate space, and the $\rho$ gets the attention output back to each token's coordinate space. \label{fig:gta_mech}}
    \vspace{-5mm}
\end{wrapfigure}

\eqnref{eq:rta_2}, given all queries $Q$, can be compactly rewritten in an implementation-friendly form:
\begin{align}
O = \brhog \circledcirc {\rm Attn}\left( \brhog^{\rm T} \circledcirc Q, 
\brhog^{-1} \circledcirc K, \brhog^{-1} \circledcirc V\right),
\label{eq:rta_impl}
\end{align}
where $\brhog$ denotes a list of representations for different tokens: $\brhog:=[\rho_{g_1}, \dots, \rho_{g_n}]$, and ``$\circledcirc$" denotes token-wise matrix multiplication:
$
    \brhog \circledcirc K = \begin{bmatrix} \rho_{g_1} K_1 \cdots \rho_{g_n} K_n \end{bmatrix}^{\rm T} \in \mathbb{R}^{n \times d}
$.
Also, the transpose $^{\rm T}$ and the inverse $^{-1}$ operate element-wise on $\brhog$ (e.g., $\brhog^{\rm T} := [\rho_{g_1}^{\rm T}, \dots, \rho_{g_n}^{\rm T}]$). 
We call the attention mechanism in \eqnref{eq:rta_impl} \textit{geometric transform attention (GTA)} and show the diagram of \eqref{eq:rta_impl} in \figref{fig:gta_mech}.
Note that the additional computation of GTA is smaller than the QKV attention and the MLP in the transformer when constructing $\rho_g$ from a set of small matrices, which we will detail in~\secref{sec:rho_design} and in Appendix~\ref{sec:algo}.

\begin{figure}[t]
\centering
    \begin{tabular}{l|r}
        \includegraphics[scale=0.08]{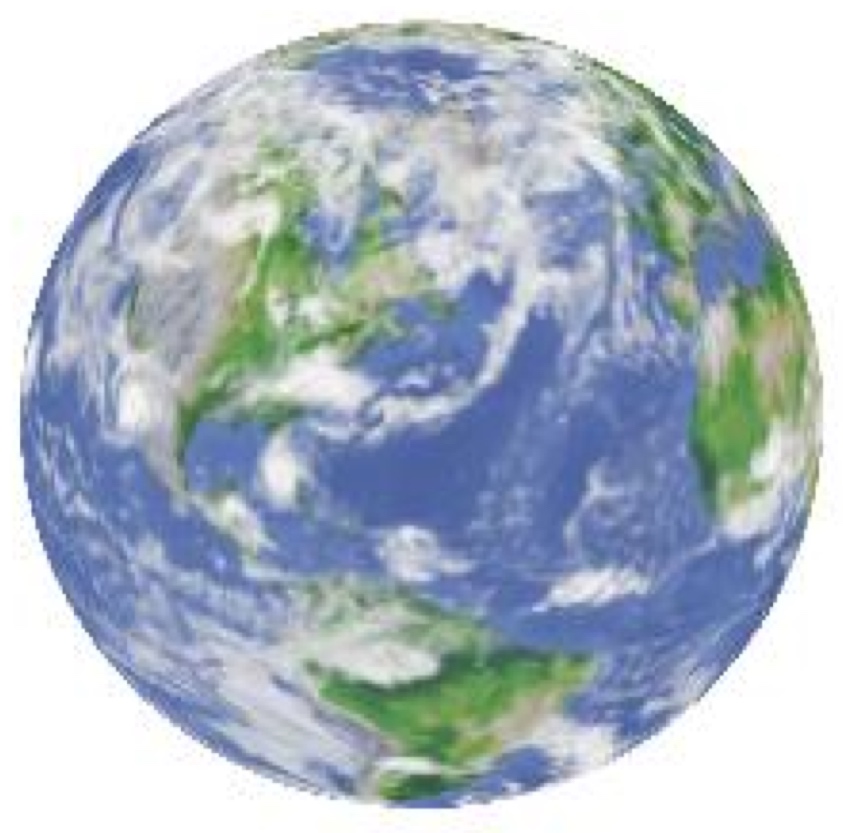}& 
        \includegraphics[scale=0.07]{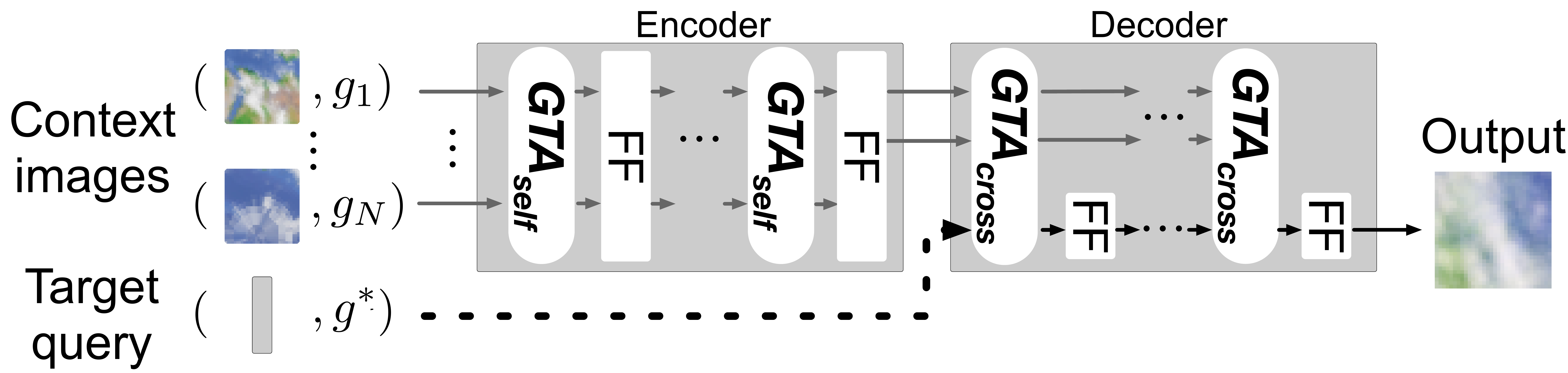}
    \end{tabular}
    \caption{\textbf{Synthetic experiment.} Left: Texture of the surrounding sphere. Right: Model architecture. The query pair consists of a learned constant value and a target extrinsic $g^*$.\label{fig:synth_exp} }
    \vspace{-0.0cm}
\end{figure}

\boldparagraph{A simple NVS experiment}
We first demonstrate that GTA improves learning as compared to APE and RPE in a simplified NVS experiment. We construct a setting where only camera rotations are relevant to show that the complexity of $\rho_g$ can be adapted to the problem complexity. A single empty scene surrounded by an enclosing sphere whose texture is shown in \figref{fig:synth_exp} left is considered. All cameras are placed in the center of the scene where they can be rotated but not translated. Each scene consists of 8 context images with 32x32 pixel resolution rendered with a pinhole camera model. The camera poses are chosen by randomly sampling camera rotations. We randomize the global coordinate system by setting it to the first input image. This increases the difficulty of the task and is similar to standard NVS tasks, where the global origin may be placed anywhere in the scene. 
The goal is to render a target view given its camera extrinsic and a set of context images.

We employ a transformer-based encoder-decoder architecture shown on the right of \figref{fig:synth_exp}.
Camera extrinsics in this experiment form the 3D rotation group: $SO(3)$. We choose $\rho_g$ to be a block concatenation of the camera rotation matrix: 
\begin{align}
    \rho_{g_i} := \underbrace{R_i \oplus \cdots \oplus R_i}_{d/3~\text{times}},
\end{align}\vspace{-4mm}%

where $R_i$ is the $3\times 3$ matrix representation of the extrinsic $g_i\in SO(3)$ linked to the $i$-th token. $A\oplus B$ denotes block-concatenation: $A\oplus B=[\begin{smallmatrix}A & 0 \\ 0 & B\end{smallmatrix}]$. Because here each $\rho_{g_i}$ is orthogonal, the transpose of $\rho_{g_i}$ becomes the inverse, thus the same transformation is applied across query, key, and value vector for each patch.

We compare this model to APE- and RPE-based transformers as baselines.
For the APE-based transformer, we add each flattened rotation matrix associated with each token to each attention layer's input. Since we could not find an RPE-based method that is directly applicable to our setting with rotation matrices, we use an RPE-version of our attention where instead of multiplying the matrices with the QKV features, we apply the matrices to 
\textit{biases}. More specifically, for each head, we prepare learned bias vectors $b^{Q}, b^{K}, b^{V} \in \mathbb{R}^9$ concatenated with each of the QKV vectors of each head and apply the representation matrix defined by $\rho(g): = R \oplus R \oplus R \in \mathbb{R}^{9\times9}$, only to the bias vectors. We describe this RPE-version of GTA in more detail in Appendix~\ref{sec:synthexp}.

\figref{fig:syn_results} on the left shows that the GTA-based transformer outperforms both the APE and RPE-based transformers in terms of both training and test performance. In~\figref{fig:syn_results} on the right, the GTA-based transformer reconstructs the image structure better than the other PE schemes.



\begin{figure}[t]
\centering
    \begin{tabular}{lcccc}\multirow{3}{*}{\includegraphics[scale=0.30]{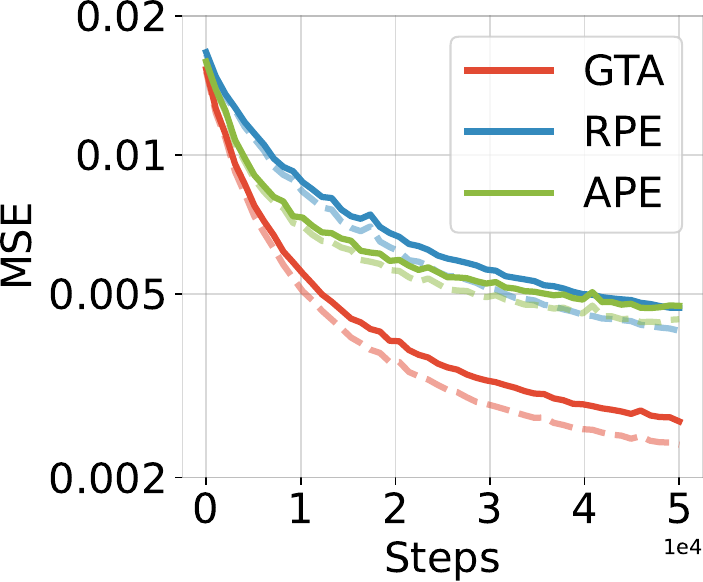}}&
         Ground truth& GTA & RPE & APE \\
         \cmidrule{2-5}
         &
         \includegraphics[scale=0.15]{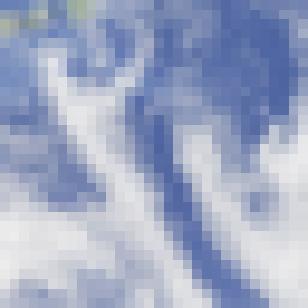} 
         & 
         \includegraphics[scale=0.15]{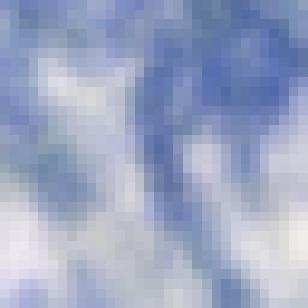} &
         \includegraphics[scale=0.15]{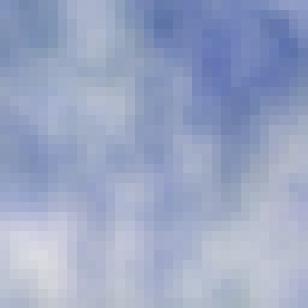} &
         \includegraphics[scale=0.15]{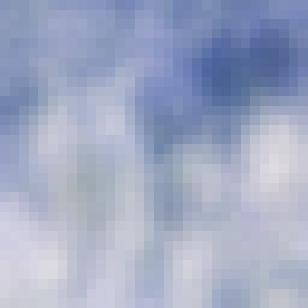} \\
         &
        \includegraphics[scale=0.15]{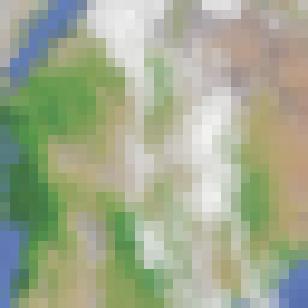} & 
         \includegraphics[scale=0.15]{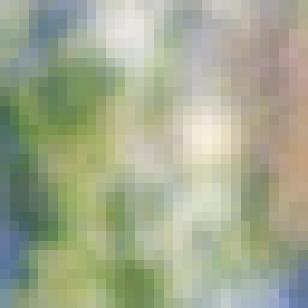} &
         \includegraphics[scale=0.15]{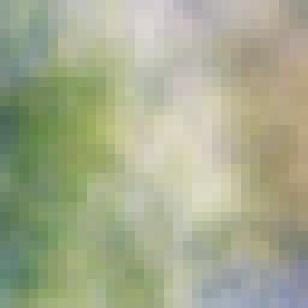} &
         \includegraphics[scale=0.15]{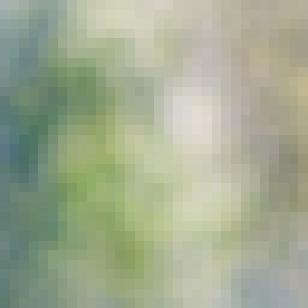} \\
    \end{tabular}%
\caption{\textbf{Results on the synthetic dataset.} Left: The solid and dashed lines indicate test and train errors. Right: Patches predicted with different PE schemes.\label{fig:syn_results}}
\vspace{-3mm}
\end{figure}
\subsection{Token structure and design of representation $\rho$ for NVS} \label{sec:rho_design}

In the previous experiment, tokens were simplified to comprise an entire image feature and an associated camera extrinsic. This differs from typical NVS model token structures where patched image tokens are used, and each of the tokens can be linked not only to a camera transformation but also to a 2D location within an image. 
To adapt GTA to such NVS models, we now describe how we associate each feature with a geometric attribute and outline one specific design choice for $\rho$.

\paragraph{Token structure:}
\begin{wrapfigure}{l}{0.30\textwidth}
    \centering
    \includegraphics[scale=0.054]{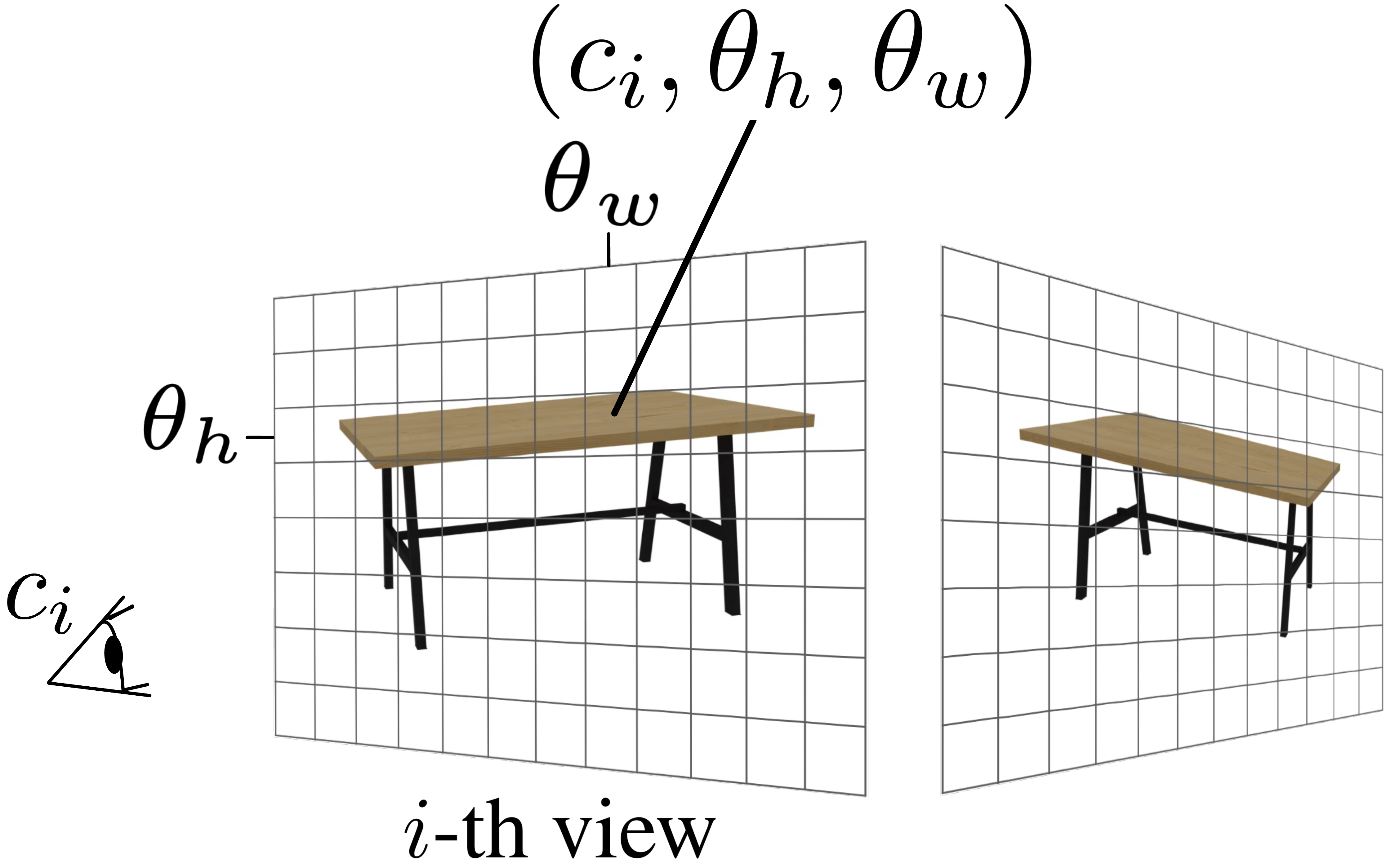}
     \caption{Geometric attributes.
     ~\label{fig:token}}
\end{wrapfigure}
We follow a common way to compose the input tokens for the transformer as in \citep{Sajjadi2022CVPR, Du2023CVPR}. We assume that for each view, we have image patches or pixels of the size of $H \times W$, and each patch or pixel token consists of a pair of a feature value $x\in \mathbb{R}^d$ and geometric attributes that are a camera extrinsic $c\in SE(3)$ and a 2D image position. 
For image PE, it would be natural to encode each position as an element of the 2D translation group $T(2)$. However, we found, similarly to the Fourier feature embeddings used in APE and RPE and rotary PE~\citep{Su2021ARXIV2}, encoding the image positions as elements of the 2D rotation group $SO(2)$ exhibits better performance than using $T(2)$. Thus, we represent each image position as an element of the direct product of the two $SO(2)$ groups: $(\theta_h, \theta_w) \in SO(2) \times SO(2)$ where $\theta_h, \theta_w\in [0, 2\pi)$.
Here, we identify the $SO(2)$ element with the 2D rotation angle. We associate the top left patch (or pixel) with the value $(0, 0)$, while the bottom right patch corresponds to $(2\pi(H-1)/H, 2\pi(W-1)/W)$. For the intermediate patches, we compute their values using linear interpolation of the angle values between the top left and bottom right patches. Overall, we represent the geometric attribute of each token of the $i$-th view by 
\begin{align}
g:= (c_i, \theta_h, \theta_w) \in SE(3)\times SO(2)  \times SO(2) =: G.
\end{align}
\figref{fig:token} illustrates how we represent each geometric attribute of each token.



\boldparagraph{Design of $\rho$} 
\begin{table}[]
    \centering
    \caption{\textbf{Components of $\rho_g$}. 
    \vspace{-2mm}
    \label{tab:rep_list}}
    \begin{tabular}{rcccc}
    &$\sigma_{\rm cam}(c)$ &$\sigma_{\rm rot}(r)$ &$\sigma_{h}({\theta_h})$ &$\sigma_{w}({\theta_w})$ \\
    \midrule
    matrix form &
    $
    \setlength\arraycolsep{2.0pt}
        \begin{bmatrix}
                R & T \\
                0 & 1 
        \end{bmatrix}$ &
    $
    \setlength\arraycolsep{0pt}
    \renewcommand{\arraystretch}{0}
        \begin{bmatrix}
            D_r^{(l_1)} &&\\
            & \ddots &\\
            && D_r^{(l_{N_{\rm rot}})} \\
        \end{bmatrix} $ &
    $
    \setlength\arraycolsep{0pt}
    \renewcommand{\arraystretch}{0}
         \begin{bmatrix}
                M^{(f_1)}_{\theta_h} &&\\
                & \ddots &\\
                && M^{(f_{N_{h}})}_{\theta_h}
        \end{bmatrix}$&
    $
    \setlength\arraycolsep{0pt}
     \renewcommand{\arraystretch}{0}
    \begin{bmatrix}
        M^{(f_1)}_{\theta_w} &&\\
        & \ddots &\\
        && M^{(f_{N_w})}_{\theta_w}
    \end{bmatrix}$\\
    \vspace{3mm}
    multiplicity & $s$ & $t$ & $u$ & $v$ \\
    \end{tabular}%
    \vspace{-3mm}
\end{table} 
What representation to use is a design choice similar to the design choice of the embedding in APE and RPE. As a specific design choice for the representation for NVS tasks, we propose to compose $\rho_g$ by the direct sum of multiple irreducible representation matrices, each responding to a specific component of the group $G$. Specifically, $\rho_g$ is composed of four different types of representations and is expressed in block-diagonal form as follows: 
\begin{align}
    \rho_g := \sigma^{\oplus s}_{\rm cam}(c) \oplus \sigma^{\oplus t}_{\rm rot}(r)  \oplus  \sigma^{\oplus u}_{h}({\theta_h})  \oplus \sigma^{\oplus v}_{w}({\theta_w}),
\end{align}
where ``$\oplus$" denotes block-concatenation $A\oplus B=[\begin{smallmatrix}A & 0 \\ 0 & B\end{smallmatrix}]$ and $A^{\oplus a}$ indicates repeating the block concatenation of $A$ a total of $a$ times. We introduce an additional representation $\sigma_{\rm rot}(r)$ that captures only the rotational information of $c$, with which we find moderate improvements in performance. 
\tabref{tab:rep_list} summarizes the matrix form we use for each representation. 
Specifically, $M_{\theta}^{(f)}$ is a 2D rotation matrix with frequency $f$ that is analogous to the frequency parameter used in Fourier feature embeddings in APE and RPE. $D_r^{(l)}$ can be thought of as the 3D version of $M_{\theta}^{(f)}$. Please refer to Appendix~\ref{sec:rho_components} for more detailed descriptions of these matrices.
\figref{fig:repmat} in the Appendix displays the actual representation matrices used in our experiments. The use of the Kronecker product is also a typical way to compose representations, which we describe in Appendix~\ref{sec:kroneckergta}.

\section{Experimental Evaluation} \label{sec:experiment}
\begin{figure}[t]
    \begin{minipage}{1.0\textwidth}
    \centering
        \captionof{table}{\textbf{Test metrics.} Left: CLEVR-TR, Right: MSN-Hard. $\dagger$Models are trained and evaluated on MultiShapeNet, not MSN-Hard. They are different but generated from the same distribution.\label{tab:comp_nvs_1}}
        \begin{tabular}{lr}
            \toprule
            &PSNR$\uparrow$\\
            \midrule 
          APE &33.66 \\
          RPE & 36.08 \\ 
          \midrule
          SRT & 33.51 \\
          RePAST & 37.27 \\
          GTA~(Ours)& \textbf{39.63} \\
            \bottomrule
        \end{tabular}
        \hspace{1mm}
         \begin{tabular}{lrrr}
            \toprule
                 & PSNR$\uparrow$ & LPIPS $\downarrow$ & SSIM$\uparrow$\\
            \midrule 
            LFN$^\dagger$~\citep{Sitzmann2021NEURIPS} &  14.77 & 0.582 & 0.328 \\
            PixelNeRF$^\dagger$~\citep{Yu2021CVPR} & 21.97 & 0.332 & 0.689 \\
            \midrule 
            SRT~\citep{Sajjadi2022CVPR}& 24.27 & 0.368 &0.741\\
            RePAST~\citep{Safin2023ARXIV} & 24.48 & 0.348 & 0.751 \\
            SRT+GTA~(Ours) & \textbf{25.72} &\textbf{0.289}& \textbf{0.798} \\
        \bottomrule
        \end{tabular}%
    \end{minipage}%
    \vspace{5mm}
     \begin{minipage}{0.74\textwidth}
        \centering
        \begin{tabular}{cccc}
            \multicolumn{4}{c}{Context images}  \\
             \multicolumn{4}{c}{\hspace{-4mm}\includegraphics[scale=0.22]{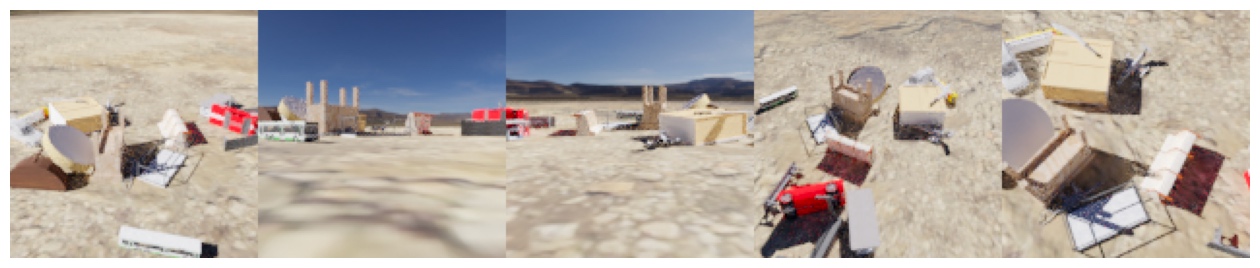}} \\
              \setlength\tabcolsep{0pt}
              \hspace{-4mm}SRT &  \hspace{-4mm}RePAST &  \hspace{-4mm}GTA~(Ours) &  \hspace{-4mm}Ground truth \\
             \hspace{-4mm}\includegraphics[scale=0.29]{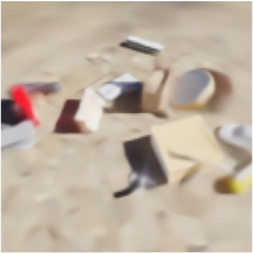} & 
             \hspace{-4mm}\includegraphics[scale=0.29]{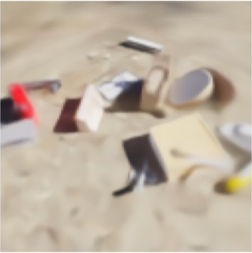} & 
             \hspace{-4mm}\includegraphics[scale=0.29]{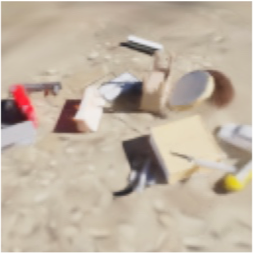} & 
             \hspace{-4mm}\includegraphics[scale=0.29]{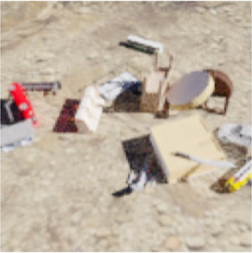}
        \end{tabular}
        \captionof{figure}{\textbf{Qualitative results on MSN-Hard}.\label{fig:rendered_msn} }
    \end{minipage}%
 \begin{minipage}{0.26\textwidth}
    \centering
    \includegraphics[scale=0.35]{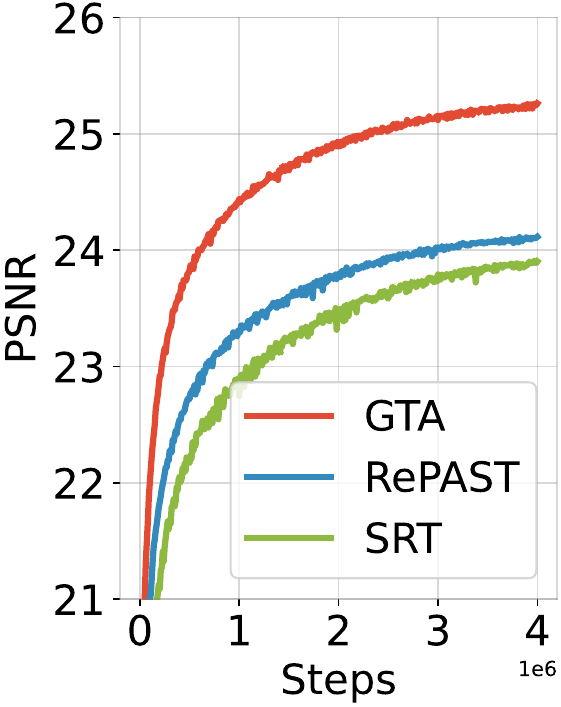}\\
    \captionof{figure}{\textbf{Validation PSNR curves on MSN-Hard.}}
    \label{fig:psnr_msn}
    \end{minipage}%
\end{figure}

We conducted experiments on several sparse NVS tasks to evaluate GTA and compare the reconstruction quality with different PE schemes as well as existing NVS methods. 

\boldparagraph{Datasets} We evaluate our method on two synthetic 360\textdegree\ datasets with sparse and wide baseline views (\textit{CLEVR-TR} and \textit{MSN-Hard}) and on two datasets of real scenes with distant views (\textit{RealEstate10k} and \textit{ACID}). We train a separate model for each dataset and describe the properties of each dataset below.
CLEVR with translation and rotation (CLEVR-TR) is a multi-view version of CLEVR~\citep{Johnson2017CVPR} that we propose. It features scenes with randomly arranged basic objects captured by cameras with azimuth, elevation, and translation transformations. 
We use this dataset to measure the ability of models to understand the underlying geometry of scenes. We set the number of context views to 2 for this dataset. Generating 360\textdegree\ images from 2 context views is challenging because parts of the scene will be unobserved. The task is solvable because all rendered objects have simple shapes and textures. This allows models to infer unobserved regions if they have a good understanding of the scene geometry.
MultiShapeNet-Hard (MSN-Hard) is a challenging dataset introduced in~\citet{Sajjadi2022NEURIPS, Sajjadi2022CVPR}. 
Up to 32 objects appear in each scene and are drawn from 51K ShapeNet objects~\citep{Chang2015}, each of which can have intricate textures and shapes. Each view is captured from a camera pose randomly sampled from 360\textdegree\ viewpoints.
Objects in test scenes are withheld during training. MSN-Hard assesses both the understanding of complex scene geometry and the capability to learn strong 3D object priors.
Each scene has 10 views, and following \citet{Sajjadi2022NEURIPS, Sajjadi2022CVPR}, we use 5 views as context views and the remaining views as target views.
RealEstate10k~\citep{zhou2018stereo} consists of real indoor and outdoor scenes with estimated camera parameters.
ACID~\citep{Liu2021ICCV} is similar to RealEstate10k, but solely includes outdoor scenes. 
Following \citet{Du2023CVPR}, during training, we randomly select two context views and one intermediate target view per scene. At test time, we sample distant context views with 128 time-step intervals and evaluate the reconstruction quality of intermediate views.
\begin{figure}[t]
    \begin{minipage}{\textwidth}
        \centering
        \captionof{table}{\textbf{Results on RealEstate10k and ACID}. Top: NeRF methods. Bottom: transformer methods.}
        \begin{tabular}{lrrrrrr}
            \toprule
                 &\multicolumn{3}{c}{RealEstate10k} &
                 \multicolumn{3}{c}{ACID}  \\  &PSNR$\uparrow$&LPIPS$\downarrow$&SSIM$\uparrow$&PSNR$\uparrow$&LPIPS$\downarrow$&SSIM$\uparrow$\\
            \midrule 
            PixelNeRF~\citep{Yu2021CVPR} & 13.91& 0.591 &0.460 & 16.48 & 0.628 & 0.464 \\
            StereoNeRF~\citep{Chibane2021CVPR} & 15.40 & 0.604  & 0.486 &$-$&$-$&$-$\\
            IBRNet~\citep{Wang2021CVPR} &15.99 & 0.532 & 0.484 & 19.24  & 0.385 & 0.513 \\
            GeoNeRF~\citep{Johari2022CVPR}  & 16.65& 0.541 & 0.511 &$-$&$-$&$-$\\
            MatchNeRF~\citep{Chen2023ARXIV} &\textbf{23.06} & 0.258 & 0.830 &$-$&$-$&$-$\\
            \midrule
            GPNR~\citep{Suhail2022ECCV} &18.55 & 0.459 &0.748& 17.57 & 0.558 & 0.719 \\    
            \citet{Du2023CVPR}  & 21.65 & 0.285 & 0.822 & 23.35  & 0.334 &  0.801  \\ 
            \citet{Du2023CVPR} + GTA~(Ours) & 22.85 & \textbf{0.255} & \textbf{0.850} &  \textbf{24.10} &   \textbf{0.291} & \textbf{0.824} \\
            \bottomrule
        
        \end{tabular}%
    \label{tab:comp_nvs_2}
    \end{minipage}%
    \vspace*{5mm}
    \begin{minipage}{\textwidth}
        \centering
        \begin{tabular}{cccc}
            Context images & \citet{Du2023CVPR} & GTA~(Ours) & Ground truth  \\
             \includegraphics[scale=0.21]{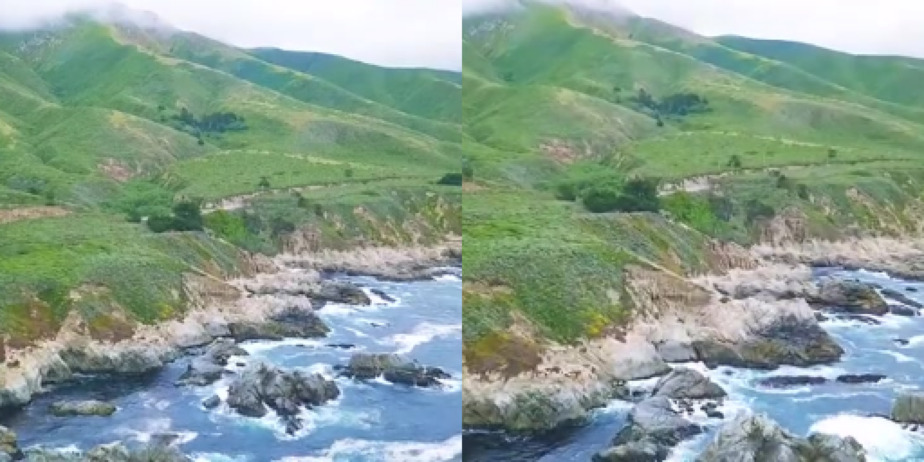} &
             \includegraphics[scale=0.21]{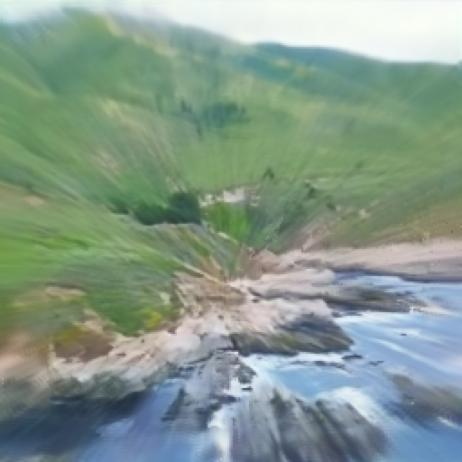} & 
             \includegraphics[scale=0.21]{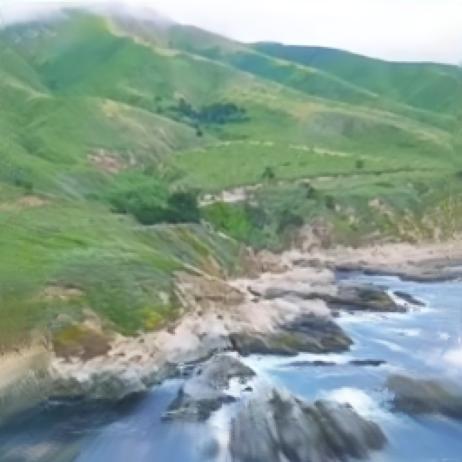} &
             \includegraphics[scale=0.21]{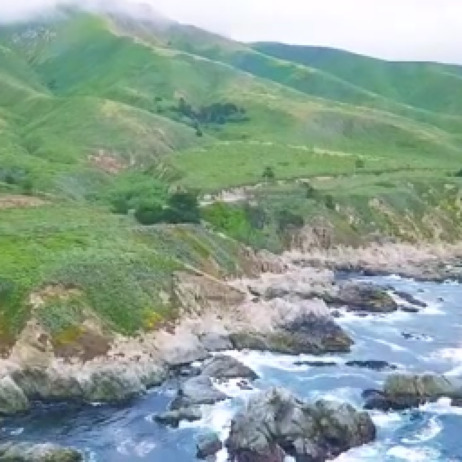} \\
             \includegraphics[scale=0.21]{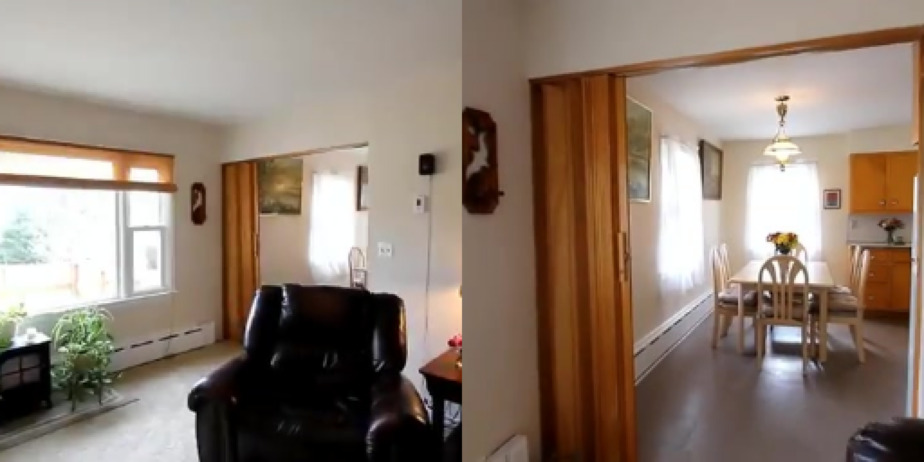} &
             \includegraphics[scale=0.21]{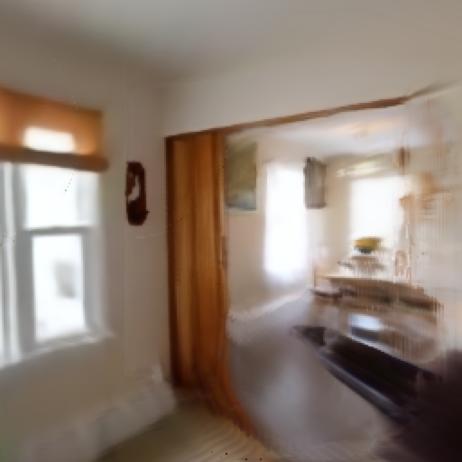} & 
             \includegraphics[scale=0.21]{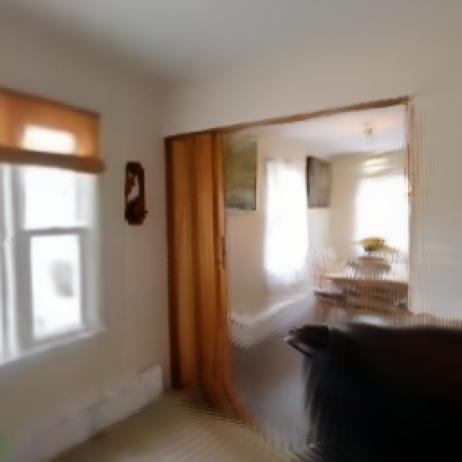} &
             \includegraphics[scale=0.21]{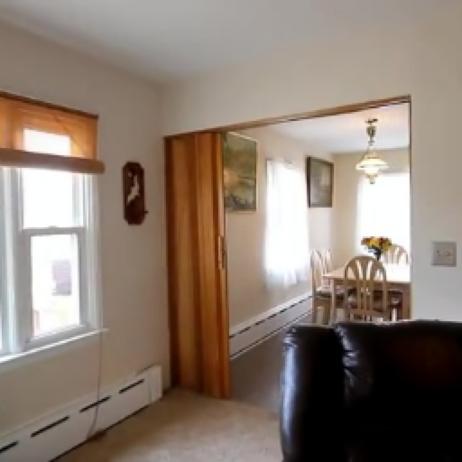}
        \end{tabular}
        \captionof{figure}{\textbf{Qualitative results.} Top: ACID, Bottom: RealEstate10k.\label{fig:rendered_2} }
    \end{minipage}%
    \vspace{-0.5cm}
\end{figure}

\boldparagraph{Baselines} 
Scene representation transformer (SRT)~\citep{Sajjadi2022CVPR}, a transformer-based NVS method, serves as our baseline model on CLEVR-TR and MSN-Hard. SRT is a similar architecture to the one we describe in \figref{fig:synth_exp}, but instead of the extrinsic matrices, SRT encodes the ray information into the architecture by concatenating Fourier feature embeddings of rays to the input pixels of the encoder. SRT is an APE-based model. Details of the SRT rendering process are provided in Appendix~\ref{sec:srt} and \figref{fig:srt_model}.
We also train another more recent transformer-based NVS model called RePAST~\citep{Safin2023ARXIV}. 
This model is a variant of SRT and encodes ray information via an RPE scheme, where, in each attention layer, the ray embeddings are added to the query and key vectors. The rays linked to the queries and keys are transformed with the extrinsic matrix associated with a key-value token pair, before feeding them into the Fourier embedding functions, to represent both rays in the same coordinate system. RePAST is the current state-of-the-art method on MSN-Hard. The key difference between GTA and RePAST is that the relative transformation is applied directly to QKV features in GTA, while it is applied to rays in RePAST.

For RealEstate10k and ACID, we use the model proposed in \citet{Du2023CVPR}, which is the state-of-the-art model on those datasets, as our baseline. Their model is similar to SRT, but has architectural improvements and uses an epipolar-based token sampling strategy. The model encodes extrinsic matrices and 2D image positions to the encoder via APE, and also encodes rays associated with each query and context image patch token in the decoder via APE. 

We implement our model by extending those baselines. Specifically, we replace all attention layers in both the encoder and decoder with GTA and remove any vector embeddings of rays, extrinsic matrices, and image positions from the model. 
We train our models and baselines with the same settings within each dataset. We train each model for 2M and 4M iterations on CLEVR-TR and MSH-Hard and for 300K iterations on both RealEstate10k and ACID, respectively. We report the reproduced numbers of baseline models in the main tables and show comparisons between the reported values and our reproduced results in \tabref{tab:comp_nvs_srt} and~\tabref{tab:comp_nvs_du} in Appendix~\ref{sec:expsettings}.
Please also see Appendix~\ref{sec:expsettings} for more details about our experimental settings.

\boldparagraph{Results}
Tables~\ref{tab:comp_nvs_1} and \ref{tab:comp_nvs_2} show that GTA improves the baselines in all reconstruction metrics on all datasets. 
\figref{fig:rendered_msn} shows that on MSN-Hard, the GTA-based model renders sharper images with more accurate reconstruction of object structures than the baselines.
\figref{fig:rendered_2} shows that our GTA-based transformer further improves the geometric understanding of the scenes over \citet{Du2023CVPR} as evidenced by the sharper results and the better recovered geometric structures. 
Appendix \ref{sec:app_rendered_images} provides additional qualitative results.
Videos are provided in the supplemental material.
We also train models, encoding 2D positions and camera extrinsics via APE and RPE for comparison. See Appendix~\ref{sec:exp_ape_rpe} for details. 
\figref{fig:psnr_msn} shows that GTA-based models improve learning efficiency over SRT and RePAST by a significant margin and reach the same performance as RePAST using only 1/6 of the training steps on MSN-Hard. GTA also outperforms RePAST in terms of wall-clock time as each gradient update step is slightly faster than RePAST, see also \tabref{tab:time} in Appendix~\ref{sec:time}.

\boldparagraph{Comparison to other PE methods}
We compare GTA with other PE methods on CLEVR-TR. All models are trained for 1M iterations. See Appendix~\ref{sec:exp_otherenc} for the implementation details. \tabref{tab:comp_3dvembs} shows that GTA outperforms other PE schemes. GTA is better than RoPE+FTL, which uses RoPE~\citep{Su2021ARXIV2} for the encoder-decoder transformer and transforms latent features of the encoder with camera extrinsics \citep{Worrall2017ICCV}. This shows the efficacy of the layer-wise geometry-aware interactions in GTA.

\begin{table}
    \centering
    \captionof{table}{\textbf{PE schemes.} MLN: Modulated layer normalization~\citep{Hong2023ARXIV, Liu2023ARXIV}. ElemMul: Element-wise Multiplication. GBT: geometry-biased transformers~\citep{Venkat2023ARXIV}. FM: Frustum Embedding~\citep{Liu2022ECCV}. RoPE+FTL:~\citep{Su2021ARXIV2, Worrall2017ICCV}.\label{tab:comp_3dvembs}}
    \begin{tabular}{lcccccccc}
        \toprule & MLN & SRT & ElemMul & GBT & FM & RoPE+FTL & GTA\\
        \midrule 
        PSNR$\uparrow$ & 32.48 &  33.21 & 34.74 & 35.63 & 37.23 & 38.18 & \textbf{38.99} \\
        \bottomrule
    \end{tabular}
    \captionof{table}{\textbf{Effect of the transformation on $V$}. Left: Test PSNRs on ClEVR-TR and MSN-Hard. Right: Inception scores (IS) and FIDs of DiT-B/2~\citep{Peebles2023ICCV} on 256x256 ImageNet.\label{tab:trnsfm_v}}
    \vspace{2mm}
    \setlength\tabcolsep{3.0pt}
    \begin{tabular}{lrr}
        \toprule
         &CLEVR-TR & MSN-Hard \\
        \midrule 
        No $\rho_g$ on $V$ &  36.54&23.77\\
        GTA & \textbf{38.99}& \textbf{24.58}\\
        \bottomrule
    \end{tabular}
    \hspace{5mm}
    \setlength\tabcolsep{3.0pt}
    \begin{tabular}{lrr}
        \toprule
         &IS$\uparrow$ & FID-50K$\downarrow$  \\
        \midrule 
        DiT~\citep{Peebles2023ICCV} &145.3 & 7.02 \\
        DiT + 2D-RoPE &  151.8 & 6.26 \\
        DiT + GTA & \textbf{158.2}& \textbf{5.87}\\
        \bottomrule
    \end{tabular}%
\end{table}

\boldparagraph{Effect of the transformation on $V$}
Rotary positional encoding (RoPE) \citep{Su2021ARXIV2, Sun2022ACL} is similar to the $SO(2)$ representations in GTA. An interesting difference to GTA is that the RoPE only applies transformations to query and key vectors, and not to the value vectors. In our setting, this exclusion leads to a discrepancy between the coordinate system of the key and value vectors, both of which interact with the tokens from which the query vectors are derived. \tabref{tab:trnsfm_v} left shows that removing the transformation on the value vectors leads to a significant drop in performance in our NVS tasks. Additionally, \tabref{tab:trnsfm_v} right shows the performance on the ImageNet generative modeling task with diffusion models. Even on this purely 2D task, the GTA mechanism is better compared to RoPE as an image positional encoding method (For more details of the diffusion experiment, please refer to Appendix~\ref{sec:DiTsettings}).

\begin{figure}
    \centering
    \begin{tabular}{ccc|c}
         Query& RePAST & GTA & \\  
         \includegraphics[scale=0.27]{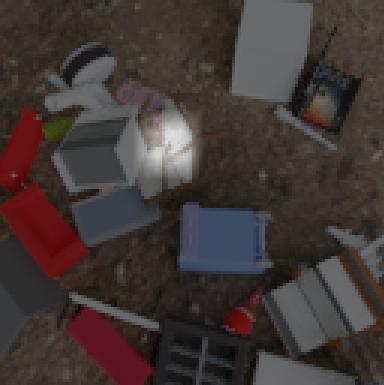} & 
         \includegraphics[scale=0.27]{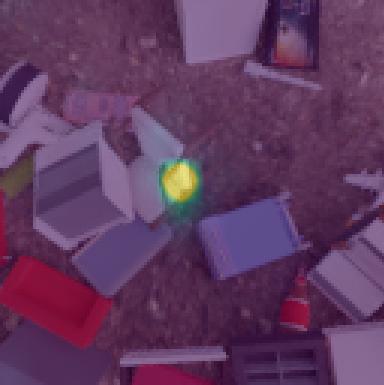} & 
         \includegraphics[scale=0.27]{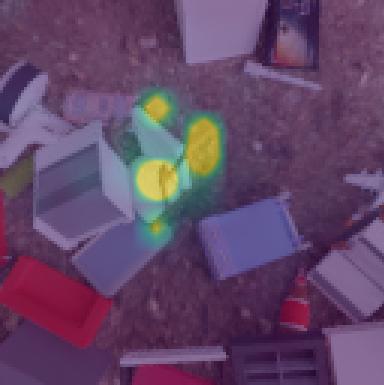} &
         \includegraphics[scale=0.45]{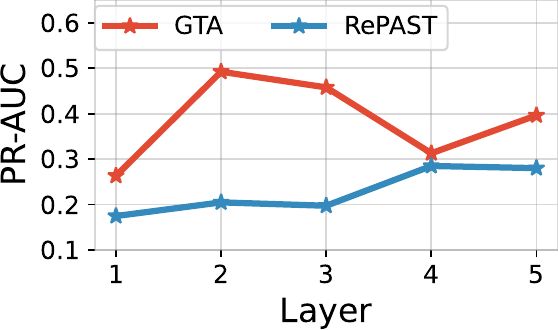}
    \end{tabular}
    \caption{\textbf{Attention analysis.} Given a query token (white region), the attention weights on a context view are visualized. GTA can identify the shape of the object that corresponds to the given query. Right: Quantitative evaluation of alignments between attention matrices and object masks.}
    \label{fig:attn_vis}
\end{figure}

\boldparagraph{Object localization}
As demonstrated in \figref{fig:attn_vis} on MSN-Hard, the GTA-based transformer not only correctly finds patch-to-patch associations but also recovers \textit{patch-to-object} associations already in the second attention layer of the encoder. For quantitative evaluation, we compute precision-recall-AUC (PR-AUC) scores based on object masks provided by MSN-Hard. In short, the score represents, given a query token belonging to a certain object instance, how well the attention matrix aligns with the object masks across all context views.
Details on how we compute PR-AUC are provided in Appendix~\ref{sec:attn_patterns}.
The PR-AUCs for the second attention layer are 0.492 and 0.204 with GTA and RePAST, respectively, which shows that our GTA-based transformer quickly identifies where to focus attention at the object level. 

\begin{table}[t]
    \centering
    \caption{\textbf{Representation design.} Test PSNRs of models trained for 1M iterations.\label{tab:rep_design}}
    \begin{tabular}{cc}
    \subcaptionbox{(\checkmark): representations are not used in the encoder. Camera and image position are needed in the decoder to identify which pixel to render. \label{tab:rep_design_1}}
    [0.48\textwidth]{
    \setlength\tabcolsep{2.0pt}
    \begin{tabular}{cccrr}
        \toprule
          $SE(3)$ & $SO(2)$ & $SO(3)$  & CLEVR-TR & MSN-Hard  \\
        \midrule 
        \checkmark & (\checkmark) &  &37.45 & 20.33 \\
         (\checkmark) & \checkmark &  & 38.26& 23.82\\
          \checkmark & \checkmark & &  38.99 & 24.58 \\
          \checkmark & \checkmark & \checkmark &\textbf{39.00} & \textbf{24.80} \\
        \bottomrule
    \end{tabular}}%
    &
    \subcaptionbox{Image positions encodings. ${}^*$Single frequency. \label{tab:rep_design_impos}}
    [0.48\textwidth]{
    \setlength\tabcolsep{2.0pt}
    \begin{tabular}{lrr}
        \toprule
            & CLEVR-TR & MSN-Hard\\
         \midrule 
         $SE(3)+T(2)$                  & 37.20 & 23.69 \\
         $SE(3)+SO(2)^*$    & 38.82 & 23.98\\
         $SE(3)+SO(2)$ & \textbf{38.99} & \textbf{24.58}\\
        \bottomrule
    \end{tabular}}
    \end{tabular}
\end{table}

\boldparagraph{Representation design}
\tabref{tab:rep_design_1} shows that, without camera encoding ($SE(3)$) or image PE ($SO(2)$) in the encoder, the reconstruction quality degrades, showing that both representations are helpful in aggregating multi-view features. 
Using $SO(3)$ representations causes a moderate improvement on MSN-Hard and no improvement on CLEVR-TR. A reason for this could be that MSN-Hard consists of a wide variety of objects. By using the $SO(3)$ representation, which is invariant to camera translations, the model may be able to encode object-centric features more efficiently. 
\tabref{tab:rep_design_impos} confirms that similar to Fourier feature embeddings used in APE and RPE, multiple frequencies of the $SO(2)$ representations benefit the reconstruction quality. 


\vspace{-2mm}
\section{Conclusion}
\vspace{-2mm}
We have proposed a novel geometry-aware attention mechanism for transformers and demonstrated its efficacy by applying it to sparse wide-baseline novel view synthesis tasks. 
A limitation of GTA is that GTA and general PE schemes rely on known poses or poses estimated by other algorithms, such as COLMAP~\citep{Schoenberger2016CVPR}. An interesting future direction is to simultaneously learn the geometric information together with the forward propagation of features in the transformer.
Developing an algorithm capable of autonomously acquiring such structural information solely from observations, specifically seeking a \textit{universal learner} for diverse forms of structure akin to human capacity, represents a captivating avenue for future research.

\ificlrfinal
\section{Acknowledgement}
Takeru Miyato, Bernhard Jaeger, and Andreas Geiger were supported by the ERC Starting Grant LEGO-3D (850533) and the DFG EXC number 2064/1 - project number 390727645.
The authors thank the International Max Planck Research School for Intelligent Systems (IMPRS-IS) for supporting Bernhard Jaeger. 
We thank Mehdi Sajjadi and Yilun Du for their comments and guidance on how to reproduce the results and thank Karl Stelzner for his open-source contribution of the SRT models.
We thank Haofei Xu and Anpei Chen for conducting the MatchNeRF experiments.
We also thank Haoyu He, Gege Gao, Masanori Koyama, Kashyap Chitta, and Naama Pearl for their feedback and comments.
Takeru Miyato acknowledges his affiliation with the ELLIS (European Laboratory for Learning and Intelligent Systems) PhD program.
\fi

{\small
\bibliography{bibliography_long,bibliography,bibliography_custom}

\begin{thebibliography}{77}
\providecommand{\natexlab}[1]{#1}
\providecommand{\url}[1]{\texttt{#1}}
\expandafter\ifx\csname urlstyle\endcsname\relax
  \providecommand{\doi}[1]{doi: #1}\else
  \providecommand{\doi}{doi: \begingroup \urlstyle{rm}\Url}\fi

\bibitem[Brandstetter et~al.(2022)Brandstetter, Hesselink, van~der Pol,
  Bekkers, and Welling]{Brandstetter2021ICLR}
Johannes Brandstetter, Rob Hesselink, Elise van~der Pol, Erik~J Bekkers, and
  Max Welling.
\newblock Geometric and physical quantities improve {E}(3) equivariant message
  passing.
\newblock In \emph{Proc. of the International Conf. on Learning Representations
  (ICLR)}, 2022.

\bibitem[Brehmer et~al.(2023)Brehmer, De~Haan, Behrends, and
  Cohen]{Brehmer2023NEURIPS}
Johann Brehmer, Pim De~Haan, S{\"o}nke Behrends, and Taco Cohen.
\newblock Geometric algebra transformers.
\newblock In \emph{Advances in Neural Information Processing Systems
  (NeurIPS)}, 2023.

\bibitem[Carion et~al.(2020)Carion, Massa, Synnaeve, Usunier, Kirillov, and
  Zagoruyko]{Carion2020ECCV}
Nicolas Carion, Francisco Massa, Gabriel Synnaeve, Nicolas Usunier, Alexander
  Kirillov, and Sergey Zagoruyko.
\newblock End-to-end object detection with transformers.
\newblock In \emph{Proc. of the European Conf. on Computer Vision (ECCV)},
  2020.

\bibitem[Chang et~al.(2015)Chang, Funkhouser, Guibas, Hanrahan, Huang, Li,
  Savarese, Savva, Song, Su, Xiao, Yi, and Yu]{Chang2015}
Angel~X. Chang, Thomas Funkhouser, Leonidas Guibas, Pat Hanrahan, Qixing Huang,
  Zimo Li, Silvio Savarese, Manolis Savva, Shuran Song, Hao Su, Jianxiong Xiao,
  Li~Yi, and Fisher Yu.
\newblock {ShapeNet: An Information-Rich 3D Model Repository}.
\newblock Technical Report arXiv:1512.03012 [cs.GR], Stanford University ---
  Princeton University --- Toyota Technological Institute at Chicago, 2015.

\bibitem[Chen et~al.(2019)Chen, Song, and Hilliges]{Chen2019CVPRc}
Xu~Chen, Jie Song, and Otmar Hilliges.
\newblock Monocular neural image based rendering with continuous view control.
\newblock In \emph{Proc. IEEE Conf. on Computer Vision and Pattern Recognition
  (CVPR)}, 2019.

\bibitem[Chen et~al.(2023)Chen, Xu, Wu, Zheng, Cham, and Cai]{Chen2023ARXIV}
Yuedong Chen, Haofei Xu, Qianyi Wu, Chuanxia Zheng, Tat{-}Jen Cham, and Jianfei
  Cai.
\newblock Explicit correspondence matching for generalizable neural radiance
  fields.
\newblock \emph{arXiv.org}, 2304.12294, 2023.

\bibitem[Chibane et~al.(2021)Chibane, Bansal, Lazova, and
  Pons{-}Moll]{Chibane2021CVPR}
Julian Chibane, Aayush Bansal, Verica Lazova, and Gerard Pons{-}Moll.
\newblock Stereo radiance fields {(SRF):} learning view synthesis for sparse
  views of novel scenes.
\newblock In \emph{Proc. IEEE Conf. on Computer Vision and Pattern Recognition
  (CVPR)}, 2021.

\bibitem[Chirikjian(2000)]{chirikjian2000engineering}
Gregory~S Chirikjian.
\newblock \emph{Engineering applications of noncommutative harmonic analysis:
  with emphasis on rotation and motion groups}.
\newblock CRC press, 2000.

\bibitem[Chitta et~al.(2022)Chitta, Prakash, Jaeger, Yu, Renz, and
  Geiger]{Chitta2022PAMI}
Kashyap Chitta, Aditya Prakash, Bernhard Jaeger, Zehao Yu, Katrin Renz, and
  Andreas Geiger.
\newblock Transfuser: Imitation with transformer-based sensor fusion for
  autonomous driving.
\newblock \emph{IEEE Trans. on Pattern Analysis and Machine Intelligence
  (PAMI)}, 2022.

\bibitem[Cohen et~al.(2019)Cohen, Weiler, Kicanaoglu, and
  Welling]{Cohen2019ICML}
Taco Cohen, Maurice Weiler, Berkay Kicanaoglu, and Max Welling.
\newblock Gauge equivariant convolutional networks and the icosahedral cnn.
\newblock In \emph{Proc. of the International Conf. on Machine learning
  (ICML)}, 2019.

\bibitem[Cohen \& Welling(2014)Cohen and Welling]{Cohen2014ICLR}
Taco~S Cohen and Max Welling.
\newblock Transformation properties of learned visual representations.
\newblock In \emph{Proc. of the International Conf. on Learning Representations
  (ICLR)}, 2014.

\bibitem[De~Haan et~al.(2021)De~Haan, Weiler, Cohen, and Welling]{De2021ICLR}
Pim De~Haan, Maurice Weiler, Taco Cohen, and Max Welling.
\newblock Gauge equivariant mesh cnns: Anisotropic convolutions on geometric
  graphs.
\newblock In \emph{Proc. of the International Conf. on Learning Representations
  (ICLR)}, 2021.

\bibitem[Dosovitskiy et~al.(2021)Dosovitskiy, Beyer, Kolesnikov, Weissenborn,
  Zhai, Unterthiner, Dehghani, Minderer, Heigold, Gelly, Uszkoreit, and
  Houlsby]{Dosovitskiy2021ICLR}
Alexey Dosovitskiy, Lucas Beyer, Alexander Kolesnikov, Dirk Weissenborn,
  Xiaohua Zhai, Thomas Unterthiner, Mostafa Dehghani, Matthias Minderer, Georg
  Heigold, Sylvain Gelly, Jakob Uszkoreit, and Neil Houlsby.
\newblock An image is worth 16x16 words: Transformers for image recognition at
  scale.
\newblock In \emph{Proc. of the International Conf. on Learning Representations
  (ICLR)}, 2021.

\bibitem[Du et~al.(2023)Du, Smith, Tewari, and Sitzmann]{Du2023CVPR}
Yilun Du, Cameron Smith, Ayush Tewari, and Vincent Sitzmann.
\newblock Learning to render novel views from wide-baseline stereo pairs.
\newblock In \emph{Proc. IEEE Conf. on Computer Vision and Pattern Recognition
  (CVPR)}, 2023.

\bibitem[Dupont et~al.(2020)Dupont, Martin, Colburn, Sankar, Susskind, and
  Shan]{Dupont2020ICML}
Emilien Dupont, Miguel~Bautista Martin, Alex Colburn, Aditya Sankar, Josh
  Susskind, and Qi~Shan.
\newblock Equivariant neural rendering.
\newblock In \emph{Proc. of the International Conf. on Machine learning
  (ICML)}, 2020.

\bibitem[Falorsi et~al.(2018)Falorsi, De~Haan, Davidson, De~Cao, Weiler,
  Forr{\'e}, and Cohen]{Falorsi2018ARXIV}
Luca Falorsi, Pim De~Haan, Tim~R Davidson, Nicola De~Cao, Maurice Weiler,
  Patrick Forr{\'e}, and Taco~S Cohen.
\newblock Explorations in homeomorphic variational auto-encoding.
\newblock \emph{arXiv.org}, 1807.04689, 2018.

\bibitem[Greff et~al.(2022)Greff, Belletti, Beyer, Doersch, Du, Duckworth,
  Fleet, Gnanapragasam, Golemo, Herrmann, Kipf, Kundu, Lagun, Laradji, Liu,
  Meyer, Miao, Nowrouzezahrai, Oztireli, Pot, Radwan, Rebain, Sabour, Sajjadi,
  Sela, Sitzmann, Stone, Sun, Vora, Wang, Wu, Yi, Zhong, and
  Tagliasacchi]{Greff2022CVPR}
Klaus Greff, Francois Belletti, Lucas Beyer, Carl Doersch, Yilun Du, Daniel
  Duckworth, David~J Fleet, Dan Gnanapragasam, Florian Golemo, Charles
  Herrmann, Thomas Kipf, Abhijit Kundu, Dmitry Lagun, Issam Laradji,
  Hsueh-Ti~(Derek) Liu, Henning Meyer, Yishu Miao, Derek Nowrouzezahrai, Cengiz
  Oztireli, Etienne Pot, Noha Radwan, Daniel Rebain, Sara Sabour, Mehdi S.~M.
  Sajjadi, Matan Sela, Vincent Sitzmann, Austin Stone, Deqing Sun, Suhani Vora,
  Ziyu Wang, Tianhao Wu, Kwang~Moo Yi, Fangcheng Zhong, and Andrea
  Tagliasacchi.
\newblock Kubric: a scalable dataset generator.
\newblock In \emph{CVPR}, 2022.

\bibitem[He et~al.(2021)He, Dong, Wang, Tao, and Lin]{He2021NEURIPS}
Lingshen He, Yiming Dong, Yisen Wang, Dacheng Tao, and Zhouchen Lin.
\newblock Gauge equivariant transformer.
\newblock In \emph{Advances in Neural Information Processing Systems
  (NeurIPS)}, 2021.

\bibitem[He et~al.(2020)He, Yan, Fragkiadaki, and Yu]{He2020CVPR}
Yihui He, Rui Yan, Katerina Fragkiadaki, and Shoou{-}I Yu.
\newblock Epipolar transformers.
\newblock In \emph{Proc. IEEE Conf. on Computer Vision and Pattern Recognition
  (CVPR)}, 2020.

\bibitem[Hinton et~al.(2011)Hinton, Krizhevsky, and Wang]{Hinton2011ICANN}
Geoffrey~E Hinton, Alex Krizhevsky, and Sida~D Wang.
\newblock Transforming auto-encoders.
\newblock In \emph{Proc. of the International Conf. on Artificial Neural
  Networks (ICANN)}, 2011.

\bibitem[Hinton et~al.(2018)Hinton, Sabour, and Frosst]{Hinton2018ICLR}
Geoffrey~E Hinton, Sara Sabour, and Nicholas Frosst.
\newblock Matrix capsules with em routing.
\newblock In \emph{Proc. of the International Conf. on Learning Representations
  (ICLR)}, 2018.

\bibitem[Hong et~al.(2023)Hong, Zhang, Gu, Bi, Zhou, Liu, Liu, Sunkavalli, Bui,
  and Tan]{Hong2023ARXIV}
Yicong Hong, Kai Zhang, Jiuxiang Gu, Sai Bi, Yang Zhou, Difan Liu, Feng Liu,
  Kalyan Sunkavalli, Trung Bui, and Hao Tan.
\newblock Lrm: Large reconstruction model for single image to 3d.
\newblock \emph{arXiv.org}, 2311.04400, 2023.

\bibitem[Johari et~al.(2022)Johari, Lepoittevin, and Fleuret]{Johari2022CVPR}
Mohammad~Mahdi Johari, Yann Lepoittevin, and Fran{\c{c}}ois Fleuret.
\newblock Geonerf: Generalizing nerf with geometry priors.
\newblock In \emph{Proc. IEEE Conf. on Computer Vision and Pattern Recognition
  (CVPR)}, 2022.

\bibitem[Johnson et~al.(2017)Johnson, Hariharan, van~der Maaten, Fei-Fei,
  Lawrence~Zitnick, and Girshick]{Johnson2017CVPR}
Justin Johnson, Bharath Hariharan, Laurens van~der Maaten, Li~Fei-Fei,
  C~Lawrence~Zitnick, and Ross Girshick.
\newblock Clevr: A diagnostic dataset for compositional language and elementary
  visual reasoning.
\newblock In \emph{Proc. IEEE Conf. on Computer Vision and Pattern Recognition
  (CVPR)}, 2017.

\bibitem[Koyama et~al.(2023)Koyama, Fukumizu, Hayashi, and
  Miyato]{Koyama2023ARXIV}
Masanori Koyama, Kenji Fukumizu, Kohei Hayashi, and Takeru Miyato.
\newblock Neural fourier transform: A general approach to equivariant
  representation learning.
\newblock \emph{arXiv.org}, 2305.18484, 2023.

\bibitem[Kulh{\'a}nek et~al.(2022)Kulh{\'a}nek, Derner, Sattler, and
  Babu{\v{s}}ka]{Kulhanek2022ECCV}
Jon{\'a}{\v{s}} Kulh{\'a}nek, Erik Derner, Torsten Sattler, and Robert
  Babu{\v{s}}ka.
\newblock Viewformer: Nerf-free neural rendering from few images using
  transformers.
\newblock In \emph{Proc. of the European Conf. on Computer Vision (ECCV)},
  2022.

\bibitem[Li et~al.(2022)Li, Wang, Li, Xie, Sima, Lu, Qiao, and Dai]{Li2022ECCV}
Zhiqi Li, Wenhai Wang, Hongyang Li, Enze Xie, Chonghao Sima, Tong Lu, Yu~Qiao,
  and Jifeng Dai.
\newblock Bevformer: Learning bird's-eye-view representation from multi-camera
  images via spatiotemporal transformers.
\newblock In \emph{Proc. of the European Conf. on Computer Vision (ECCV)},
  2022.

\bibitem[Lin et~al.(2017)Lin, Doll{\'{a}}r, Girshick, He, Hariharan, and
  Belongie]{Lin2017CVPRb}
Tsung{-}Yi Lin, Piotr Doll{\'{a}}r, Ross~B. Girshick, Kaiming He, Bharath
  Hariharan, and Serge~J. Belongie.
\newblock Feature pyramid networks for object detection.
\newblock In \emph{Proc. IEEE Conf. on Computer Vision and Pattern Recognition
  (CVPR)}, 2017.

\bibitem[Liu et~al.(2021)Liu, Tucker, Jampani, Makadia, Snavely, and
  Kanazawa]{Liu2021ICCV}
Andrew Liu, Richard Tucker, Varun Jampani, Ameesh Makadia, Noah Snavely, and
  Angjoo Kanazawa.
\newblock Infinite nature: Perpetual view generation of natural scenes from a
  single image.
\newblock In \emph{Proc. of the IEEE International Conf. on Computer Vision
  (ICCV)}, 2021.

\bibitem[Liu et~al.(2023{\natexlab{a}})Liu, Wu, Van~Hoorick, Tokmakov,
  Zakharov, and Vondrick]{Liu2023ARXIV}
Ruoshi Liu, Rundi Wu, Basile Van~Hoorick, Pavel Tokmakov, Sergey Zakharov, and
  Carl Vondrick.
\newblock Zero-1-to-3: Zero-shot one image to 3d object.
\newblock \emph{arXiv.org}, 2303.11328, 2023{\natexlab{a}}.

\bibitem[Liu et~al.(2022)Liu, Wang, Zhang, and Sun]{Liu2022ECCV}
Yingfei Liu, Tiancai Wang, Xiangyu Zhang, and Jian Sun.
\newblock Petr: Position embedding transformation for multi-view 3d object
  detection.
\newblock In \emph{Proc. of the European Conf. on Computer Vision (ECCV)},
  2022.

\bibitem[Liu et~al.(2023{\natexlab{b}})Liu, Yan, Jia, Li, Gao, Wang, and
  Zhang]{Liu2023ICCV}
Yingfei Liu, Junjie Yan, Fan Jia, Shuailin Li, Aqi Gao, Tiancai Wang, and
  Xiangyu Zhang.
\newblock Petrv2: A unified framework for 3d perception from multi-camera
  images.
\newblock In \emph{Proc. of the IEEE International Conf. on Computer Vision
  (ICCV)}, 2023{\natexlab{b}}.

\bibitem[Loshchilov \& Hutter(2017)Loshchilov and Hutter]{Loshchilov2017ARXIV}
Ilya Loshchilov and Frank Hutter.
\newblock Decoupled weight decay regularization.
\newblock \emph{arXiv.org}, 1711.05101, 2017.

\bibitem[Miyato et~al.(2022)Miyato, Koyama, and Fukumizu]{Miyato2022NEURIPS}
Takeru Miyato, Masanori Koyama, and Kenji Fukumizu.
\newblock Unsupervised learning of equivariant structure from sequences.
\newblock In \emph{Advances in Neural Information Processing Systems
  (NeurIPS)}, 2022.

\bibitem[Park et~al.(2022)Park, Biza, Zhao, van~de Meent, and
  Walters]{Park2022ICML}
Jung~Yeon Park, Ondrej Biza, Linfeng Zhao, Jan~Willem van~de Meent, and Robin
  Walters.
\newblock Learning symmetric embeddings for equivariant world models.
\newblock In \emph{Proc. of the International Conf. on Machine learning
  (ICML)}, 2022.

\bibitem[Peebles \& Xie(2023)Peebles and Xie]{Peebles2023ICCV}
William Peebles and Saining Xie.
\newblock Scalable diffusion models with transformers.
\newblock In \emph{Proc. of the IEEE International Conf. on Computer Vision
  (ICCV)}, 2023.

\bibitem[Peng et~al.(2023)Peng, Chen, Fu, Liang, and Cheng]{Peng2023WACV}
Lang Peng, Zhirong Chen, Zhangjie Fu, Pengpeng Liang, and Erkang Cheng.
\newblock Bevsegformer: Bird's eye view semantic segmentation from arbitrary
  camera rigs.
\newblock In \emph{Proc. of the IEEE Winter Conference on Applications of
  Computer Vision (WACV)}, 2023.

\bibitem[Perez et~al.(2018)Perez, Strub, De~Vries, Dumoulin, and
  Courville]{Perez2018AAAI}
Ethan Perez, Florian Strub, Harm De~Vries, Vincent Dumoulin, and Aaron
  Courville.
\newblock Film: Visual reasoning with a general conditioning layer.
\newblock In \emph{Proc. of the Conf. on Artificial Intelligence (AAAI)}, 2018.

\bibitem[Qin et~al.(2022)Qin, Yu, Wang, Guo, Peng, and Xu]{Qin2022CVPR}
Zheng Qin, Hao Yu, Changjian Wang, Yulan Guo, Yuxing Peng, and Kai Xu.
\newblock Geometric transformer for fast and robust point cloud registration.
\newblock In \emph{Proc. IEEE Conf. on Computer Vision and Pattern Recognition
  (CVPR)}, 2022.

\bibitem[Ramachandran et~al.(2019)Ramachandran, Parmar, Vaswani, Bello,
  Levskaya, and Shlens]{Ramachandran2019NEURIPS}
Prajit Ramachandran, Niki Parmar, Ashish Vaswani, Irwan Bello, Anselm Levskaya,
  and Jon Shlens.
\newblock Stand-alone self-attention in vision models.
\newblock \emph{Advances in Neural Information Processing Systems (NeurIPS)},
  2019.

\bibitem[Ranftl et~al.(2021)Ranftl, Bochkovskiy, and Koltun]{Ranftl2021ICCV}
Ren{\'e} Ranftl, Alexey Bochkovskiy, and Vladlen Koltun.
\newblock Vision transformers for dense prediction.
\newblock In \emph{Proc. of the IEEE International Conf. on Computer Vision
  (ICCV)}, 2021.

\bibitem[Reading et~al.(2021)Reading, Harakeh, Chae, and
  Waslander]{Reading2021CVPR}
Cody Reading, Ali Harakeh, Julia Chae, and Steven~L. Waslander.
\newblock Categorical depth distribution network for monocular 3d object
  detection.
\newblock In \emph{Proc. IEEE Conf. on Computer Vision and Pattern Recognition
  (CVPR)}, 2021.

\bibitem[Rhodin et~al.(2018)Rhodin, Salzmann, and Fua]{Rhodin2018ECCV}
Helge Rhodin, Mathieu Salzmann, and Pascal Fua.
\newblock Unsupervised geometry-aware representation for 3d human pose
  estimation.
\newblock In \emph{Proc. of the European Conf. on Computer Vision (ECCV)},
  2018.

\bibitem[Romero et~al.(2020)Romero, Bekkers, Tomczak, and
  Hoogendoorn]{Romero2020ICML}
David Romero, Erik Bekkers, Jakub Tomczak, and Mark Hoogendoorn.
\newblock Attentive group equivariant convolutional networks.
\newblock In \emph{Proc. of the International Conf. on Machine learning
  (ICML)}, 2020.

\bibitem[Russakovsky et~al.(2015)Russakovsky, Deng, Su, Krause, Satheesh, Ma,
  Huang, Karpathy, Khosla, Bernstein, Berg, and Fei-Fei]{Russakovsky2015IJCV}
Olga Russakovsky, Jia Deng, Hao Su, Jonathan Krause, Sanjeev Satheesh, Sean Ma,
  Zhiheng Huang, Andrej Karpathy, Aditya Khosla, Michael Bernstein,
  Alexander~C. Berg, and Li~Fei-Fei.
\newblock {ImageNet Large Scale Visual Recognition Challenge}.
\newblock \emph{International Journal of Computer Vision (IJCV)}, 115\penalty0
  (3):\penalty0 211--252, 2015.

\bibitem[Sabour et~al.(2017)Sabour, Frosst, and Hinton]{Sabour2017NEURIPS}
Sara Sabour, Nicholas Frosst, and Geoffrey~E Hinton.
\newblock Dynamic routing between capsules.
\newblock In \emph{Advances in Neural Information Processing Systems
  (NeurIPS)}, 2017.

\bibitem[Safin et~al.(2023)Safin, Durckworth, and Sajjadi]{Safin2023ARXIV}
Aleksandr Safin, Daniel Durckworth, and Mehdi~SM Sajjadi.
\newblock {R}e{PAST}: Relative pose attention scene representation transformer.
\newblock \emph{arXiv.org}, 2304.00947, 2023.

\bibitem[{Saha} et~al.(2022){Saha}, {Mendez Maldonado}, {Russell}, and
  {Bowden}]{Saha2022ICRA}
Avishkar {Saha}, Oscar {Mendez Maldonado}, Chris {Russell}, and Richard
  {Bowden}.
\newblock {Translating Images into Maps}.
\newblock In \emph{Proc. IEEE International Conf. on Robotics and Automation
  (ICRA)}, 2022.

\bibitem[Sajjadi et~al.(2022{\natexlab{a}})Sajjadi, Duckworth, Mahendran, van
  Steenkiste, Paveti{\'c}, Lu{\v{c}}i{\'c}, Guibas, Greff, and
  Kipf]{Sajjadi2022NEURIPS}
Mehdi~SM Sajjadi, Daniel Duckworth, Aravindh Mahendran, Sjoerd van Steenkiste,
  Filip Paveti{\'c}, Mario Lu{\v{c}}i{\'c}, Leonidas~J Guibas, Klaus Greff, and
  Thomas Kipf.
\newblock Object scene representation transformer.
\newblock In \emph{Advances in Neural Information Processing Systems
  (NeurIPS)}, 2022{\natexlab{a}}.

\bibitem[Sajjadi et~al.(2022{\natexlab{b}})Sajjadi, Meyer, Pot, Bergmann,
  Greff, Radwan, Vora, Lu{\v{c}}i{\'c}, Duckworth, Dosovitskiy,
  et~al.]{Sajjadi2022CVPR}
Mehdi~SM Sajjadi, Henning Meyer, Etienne Pot, Urs Bergmann, Klaus Greff, Noha
  Radwan, Suhani Vora, Mario Lu{\v{c}}i{\'c}, Daniel Duckworth, Alexey
  Dosovitskiy, et~al.
\newblock Scene representation transformer: Geometry-free novel view synthesis
  through set-latent scene representations.
\newblock In \emph{Proc. IEEE Conf. on Computer Vision and Pattern Recognition
  (CVPR)}, 2022{\natexlab{b}}.

\bibitem[Schönberger \& Frahm(2016)Schönberger and
  Frahm]{Schoenberger2016CVPR}
Johannes~Lutz Schönberger and Jan-Michael Frahm.
\newblock Structure-from-motion revisited.
\newblock In \emph{Proc. IEEE Conf. on Computer Vision and Pattern Recognition
  (CVPR)}, 2016.

\bibitem[Shao et~al.(2023)Shao, Wang, Chen, Li, and Liu]{Shao2023CORL}
Hao Shao, Letian Wang, Ruobing Chen, Hongsheng Li, and Yu~Liu.
\newblock Safety-enhanced autonomous driving using interpretable sensor fusion
  transformer.
\newblock In \emph{Proc. Conf. on Robot Learning (CoRL)}, 2023.

\bibitem[Shaw et~al.(2018)Shaw, Uszkoreit, and Vaswani]{Shaw2018NAACL}
Peter Shaw, Jakob Uszkoreit, and Ashish Vaswani.
\newblock Self-attention with relative position representations.
\newblock In \emph{Annual Conference of the North American Chapter of the
  Association for Computational Linguistics (NAACL-HLT)}, 2018.

\bibitem[Shu et~al.(2023)Shu, Deng, Yu, and Liu]{Shu20233ICCV}
Changyong Shu, Jiajun Deng, Fisher Yu, and Yifan Liu.
\newblock 3dppe: 3d point positional encoding for transformer-based
  multi-camera 3d object detection.
\newblock In \emph{Proc. of the IEEE International Conf. on Computer Vision
  (ICCV)}, 2023.

\bibitem[Sitzmann et~al.(2021)Sitzmann, Rezchikov, Freeman, Tenenbaum, and
  Durand]{Sitzmann2021NEURIPS}
Vincent Sitzmann, Semon Rezchikov, Bill Freeman, Josh Tenenbaum, and Fredo
  Durand.
\newblock Light field networks: Neural scene representations with
  single-evaluation rendering.
\newblock In \emph{Advances in Neural Information Processing Systems
  (NeurIPS)}, 2021.

\bibitem[Su et~al.(2021)Su, Lu, Pan, Murtadha, Wen, and Liu]{Su2021ARXIV2}
Jianlin Su, Yu~Lu, Shengfeng Pan, Ahmed Murtadha, Bo~Wen, and Yunfeng Liu.
\newblock Roformer: Enhanced transformer with rotary position embedding.
\newblock \emph{arXiv preprint arXiv:2104.09864}, 2021.

\bibitem[Suhail et~al.(2022)Suhail, Esteves, Sigal, and
  Makadia]{Suhail2022ECCV}
Mohammed Suhail, Carlos Esteves, Leonid Sigal, and Ameesh Makadia.
\newblock Generalizable patch-based neural rendering.
\newblock In \emph{Proc. of the European Conf. on Computer Vision (ECCV)},
  2022.

\bibitem[Sun et~al.(2022)Sun, Dong, Patra, Ma, Huang, Benhaim, Chaudhary, Song,
  and Wei]{Sun2022ACL}
Yutao Sun, Li~Dong, Barun Patra, Shuming Ma, Shaohan Huang, Alon Benhaim,
  Vishrav Chaudhary, Xia Song, and Furu Wei.
\newblock A length-extrapolatable transformer.
\newblock In \emph{ACL}, 2022.

\bibitem[Varma et~al.(2023)Varma, Wang, Chen, Chen, Venugopalan, and
  Wang]{Varma2023ICLR}
Mukund Varma, Peihao Wang, Xuxi Chen, Tianlong Chen, Subhashini Venugopalan,
  and Zhangyang Wang.
\newblock Is attention all that nerf needs?
\newblock In \emph{Proc. of the International Conf. on Learning Representations
  (ICLR)}, 2023.

\bibitem[Vaswani et~al.(2017)Vaswani, Shazeer, Parmar, Uszkoreit, Jones, Gomez,
  Kaiser, and Polosukhin]{Vaswani2017NEURIPS}
Ashish Vaswani, Noam Shazeer, Niki Parmar, Jakob Uszkoreit, Llion Jones,
  Aidan~N Gomez, {\L}ukasz Kaiser, and Illia Polosukhin.
\newblock Attention is all you need.
\newblock In \emph{NeurIPS}, 2017.

\bibitem[Venkat et~al.(2023)Venkat, Agarwal, Singh, and
  Tulsiani]{Venkat2023ARXIV}
Naveen Venkat, Mayank Agarwal, Maneesh Singh, and Shubham Tulsiani.
\newblock Geometry-biased transformers for novel view synthesis.
\newblock \emph{arXiv.org}, 2301.04650, 2023.

\bibitem[Wang et~al.(2021{\natexlab{a}})Wang, Geiger, and Tang]{Wang2021CVPR}
Shaofei Wang, Andreas Geiger, and Siyu Tang.
\newblock Locally aware piecewise transformation fields for 3d human mesh
  registration.
\newblock In \emph{Proc. IEEE Conf. on Computer Vision and Pattern Recognition
  (CVPR)}, 2021{\natexlab{a}}.

\bibitem[Wang et~al.(2023)Wang, Liu, Wang, Li, and Zhang]{Wang2023ARXIV}
Shihao Wang, Yingfei Liu, Tiancai Wang, Ying Li, and Xiangyu Zhang.
\newblock Exploring object-centric temporal modeling for efficient multi-view
  3d object detection.
\newblock \emph{arXiv.org}, 2303.11926, 2023.

\bibitem[Wang et~al.(2018)Wang, Girshick, Gupta, and He]{Wang2018CVPRd}
Xiaolong Wang, Ross~B. Girshick, Abhinav Gupta, and Kaiming He.
\newblock Non-local neural networks.
\newblock In \emph{Proc. IEEE Conf. on Computer Vision and Pattern Recognition
  (CVPR)}, pp.\  7794--7803, 2018.

\bibitem[Wang et~al.(2021{\natexlab{b}})Wang, Guizilini, Zhang, Wang, Zhao, and
  Solomon]{Wang2021CORL}
Yue Wang, Vitor Guizilini, Tianyuan Zhang, Yilun Wang, Hang Zhao, and Justin
  Solomon.
\newblock {DETR3D:} 3d object detection from multi-view images via 3d-to-2d
  queries.
\newblock In \emph{Proc. Conf. on Robot Learning (CoRL)}, 2021{\natexlab{b}}.

\bibitem[Watson et~al.(2023)Watson, Chan, Martin-Brualla, Ho, Tagliasacchi, and
  Norouzi]{Watson2023ICLR}
Daniel Watson, William Chan, Ricardo Martin-Brualla, Jonathan Ho, Andrea
  Tagliasacchi, and Mohammad Norouzi.
\newblock Novel view synthesis with diffusion models.
\newblock In \emph{Proc. of the International Conf. on Learning Representations
  (ICLR)}, 2023.

\bibitem[Worrall et~al.(2017)Worrall, Garbin, Turmukhambetov, and
  Brostow]{Worrall2017ICCV}
Daniel~E Worrall, Stephan~J Garbin, Daniyar Turmukhambetov, and Gabriel~J
  Brostow.
\newblock Interpretable transformations with encoder-decoder networks.
\newblock In \emph{Proc. of the IEEE International Conf. on Computer Vision
  (ICCV)}, 2017.

\bibitem[Wu et~al.(2021)Wu, Peng, Chen, Fu, and Chao]{Wu2021CVPRb}
Kan Wu, Houwen Peng, Minghao Chen, Jianlong Fu, and Hongyang Chao.
\newblock Rethinking and improving relative position encoding for vision
  transformer.
\newblock In \emph{Proc. IEEE Conf. on Computer Vision and Pattern Recognition
  (CVPR)}, 2021.

\bibitem[Xu et~al.(2023)Xu, Zhang, Cai, Rezatofighi, Yu, Tao, and
  Geiger]{Xu2023PAMI}
Haofei Xu, Jing Zhang, Jianfei Cai, Hamid Rezatofighi, Fisher Yu, Dacheng Tao,
  and Andreas Geiger.
\newblock Unifying flow, stereo and depth estimation.
\newblock \emph{IEEE Trans. on Pattern Analysis and Machine Intelligence
  (PAMI)}, 2023.

\bibitem[Yu et~al.(2021{\natexlab{a}})Yu, Li, Tancik, Li, Ng, and
  Kanazawa]{Yu2021ICCV}
Alex Yu, Ruilong Li, Matthew Tancik, Hao Li, Ren Ng, and Angjoo Kanazawa.
\newblock {PlenOctrees} for real-time rendering of neural radiance fields.
\newblock In \emph{Proc. of the IEEE International Conf. on Computer Vision
  (ICCV)}, 2021{\natexlab{a}}.

\bibitem[Yu et~al.(2021{\natexlab{b}})Yu, Ye, Tancik, and Kanazawa]{Yu2021CVPR}
Alex Yu, Vickie Ye, Matthew Tancik, and Angjoo Kanazawa.
\newblock pixelnerf: Neural radiance fields from one or few images.
\newblock In \emph{Proc. IEEE Conf. on Computer Vision and Pattern Recognition
  (CVPR)}, 2021{\natexlab{b}}.

\bibitem[Zhang \& Sennrich(2019)Zhang and Sennrich]{Zhang2019NEURIPS}
Biao Zhang and Rico Sennrich.
\newblock Root mean square layer normalization.
\newblock In \emph{Advances in Neural Information Processing Systems
  (NeurIPS)}, volume~32, 2019.

\bibitem[Zhang et~al.(2018)Zhang, Isola, Efros, Shechtman, and
  Wang]{Zhang2018CVPRa}
Richard Zhang, Phillip Isola, Alexei~A. Efros, Eli Shechtman, and Oliver Wang.
\newblock The unreasonable effectiveness of deep features as a perceptual
  metric.
\newblock In \emph{Proc. IEEE Conf. on Computer Vision and Pattern Recognition
  (CVPR)}, pp.\  586--595, 2018.

\bibitem[{Zhou} \& {Kr{\"a}henb{\"u}hl}(2022){Zhou} and
  {Kr{\"a}henb{\"u}hl}]{Zhou2022CVPR}
Brady {Zhou} and Philipp {Kr{\"a}henb{\"u}hl}.
\newblock {Cross-view Transformers for real-time Map-view Semantic
  Segmentation}.
\newblock In \emph{Proc. IEEE Conf. on Computer Vision and Pattern Recognition
  (CVPR)}, 2022.

\bibitem[Zhou et~al.(2018)Zhou, Tucker, Flynn, Fyffe, and
  Snavely]{zhou2018stereo}
Tinghui Zhou, Richard Tucker, John Flynn, Graham Fyffe, and Noah Snavely.
\newblock Stereo magnification.
\newblock \emph{ACM Transactions on Graphics}, 2018.

\bibitem[Zhu et~al.(2021)Zhu, Su, Lu, Li, Wang, and Dai]{Zhu2021ICLR}
Xizhou Zhu, Weijie Su, Lewei Lu, Bin Li, Xiaogang Wang, and Jifeng Dai.
\newblock Deformable {DETR:} deformable transformers for end-to-end object
  detection.
\newblock In \emph{Proc. of the International Conf. on Learning Representations
  (ICLR)}, 2021.

\bibitem[Zou et~al.(2023)Zou, Yu, Guo, Li, Liang, Cao, and Zhang]{Zou2023ARXIV}
Zi-Xin Zou, Zhipeng Yu, Yuan-Chen Guo, Yangguang Li, Ding Liang, Yan-Pei Cao,
  and Song-Hai Zhang.
\newblock Triplane meets gaussian splatting: Fast and generalizable single-view
  3d reconstruction with transformers.
\newblock \emph{arXiv.org}, 2312.09147, 2023.

\end{thebibliography}
}
\bibliographystyle{iclr2024_conference}

\onecolumn
\appendix

\part{Appendix} 
{
  \hypersetup{linkcolor=black}
  \parttoc
}
\section{Additional details of GTA}\label{sec:algo}
\algref{alg:overall} provides an algorithmic description based on \eqnref{eq:rta_impl} for single-head self-attention. For multi-head attention, we simply apply the group representations to all $QKV$ vectors of each head.

\begin{algorithm}
\caption{GTA for single head self-attention. \label{alg:overall}}
Input: Input tokens: $X\in \mathbb{R}^{N\times d}$, group representations: $\brhog=[\rho_{g_1}, \rho_{g_2}, ..., \rho_{g_N}]$, and weights: $W^Q, W^K, W^V \in \mathbb{R}^{d \times d}$.
\begin{enumerate}
\item Compute query, key, and value from $X$:
$$Q=XW^Q, K=XW^K, V=XW^V.$$
\item
Transform each variable with the group representations:
$$
Q \leftarrow \brhog^{\rm T}\circledcirc Q, K \leftarrow \brhog^{-1}\circledcirc K, V \leftarrow \brhog^{-1}\circledcirc V
$$

\item Compute $QKV$ attention in the same way as in the vanilla attention: 
$$
O = {\rm softmax}\left(\frac{Q K^{\rm T}}{\sqrt{d}} \right) V
$$
\item 
Apply group representations to $O$: 
$$
    O \leftarrow \brhog \circledcirc O 
$$

\item Return $O$
\end{enumerate}
\end{algorithm}

\subsection{Computational complexity}
Since the ${\bf P}_{\bf g}\circledcirc \cdot$ operation is an $n$-times  multiplication of a $d\times d$ matrix with a $d$-dimensional vector, the computational complexity of additional computation for our attention over the vanilla attention is $O(nd^2)$. This can be reduced by constructing the representation matrix with a block diagonal, with each block being small. If we keep the largest block size of the representation constant against $n$ and $d$, then the order of the ${\bf P}_{\bf g}\circledcirc \cdot$  operation becomes $O(nd)$. Thus, if $d_{\rm max}$ is relatively small, or if we increase $n$ or $d$, the computation overhead of the $\circledcirc$ operation becomes negligible compared to the computation times of the other components of a transformer, which are $O(n^2d)$ for attention and $O(nd^2)$ for feedforward layers.
In our experiments, we use a block-structured representation with a maximum block size of 5 (see ~\secref{sec:rho_design} and ~\figref{fig:repmat}).

\subsection{Details of the representation matrices}\label{sec:rho_components}
$\rho_g$ is composed of four different types of representations $\rho_c, \rho_r, \rho_{\theta_h}, \rho_{\theta_w}$ with the multiplicities of $s,t,u,v$.
Below, we describe the details of each representation.

\boldparagraph{$\sigma_{\rm cam}(c)$}
We use a homogenous rigid transformation as the representation of $c \in SE(3)$:
\begin{align}
    \sigma_{\rm cam}(c) := \begin{bmatrix}R & T \\ 0 & 1\end{bmatrix} \in \mathbb{R}^{4\times 4}\label{eq:se3}.
\end{align}
\boldparagraph{$\sigma_{\rm rot}(r)$}
 We compose $\sigma_{\rm rot}(r)$ via block concatenation of Wigner-D-matrices~\citep{chirikjian2000engineering}.  
\begin{align}
    \sigma_{\rm rot}(r) := {\textstyle \bigoplus}_k \sigma_{\rm rot}^{(l_k)}(r),~ \sigma_{\rm rot}^{(l)}(r) :=D_r^{(l)}  \in \mathbb{R}^{(2l+1)\times (2l+1)}  \label{eq:so3}
\end{align}
 where $D^{(l)}_r$ is $l$-th Wigner-D-matrix given $r$. Here, $\bigoplus_{a\in S} A^{(a)}:= A^{(a_{1})}\oplus \cdots \oplus A^{(a_{|S|})}$ and we omit the index set symbol from the above equation.
 We use these matrices because Wigner-D-matrices are the \textit{only irreducible representations} of $SO(3)$. Any linear representation~$\sigma_r, r\in SO(3)$ is equivalent to a direct sum of the matrices under a similarity transformation~\citep{chirikjian2000engineering}. 

\boldparagraph{$\sigma_h({\theta_h})$ and $\sigma_w({\theta_w})$}
Similar to $\sigma_{\rm rot}(r)$, we use 2D rotation matrices with different frequencies for $\sigma_h({\theta_h})$ and $\sigma_w({\theta_w})$. Specifically, for $\sigma_h({\theta_h})$ given a set of frequencies $\{f_k\}_{k=1}^{N_{h}}$, we define the representation as follows:
\begin{align}
    \sigma_h({\theta_h}) := {\textstyle \bigoplus}_k \sigma^{(f_k)}_h({\theta_h}),~ \sigma^{(f)}_h({\theta_h}) := M^{(f)}_{\theta_h}  = \begin{bmatrix} \cos(f\theta_h)  &  -\sin(f\theta_h) \\  \sin(f\theta_h)  &  \cos(f\theta_h) \end{bmatrix} \in \mathbb{R}^{2\times 2}. \label{eq:so2}
\end{align}
$\sigma_w({\theta_w})$ is defined analogously.

\begin{figure}[H]
    \centering
    \includegraphics[scale=0.4]{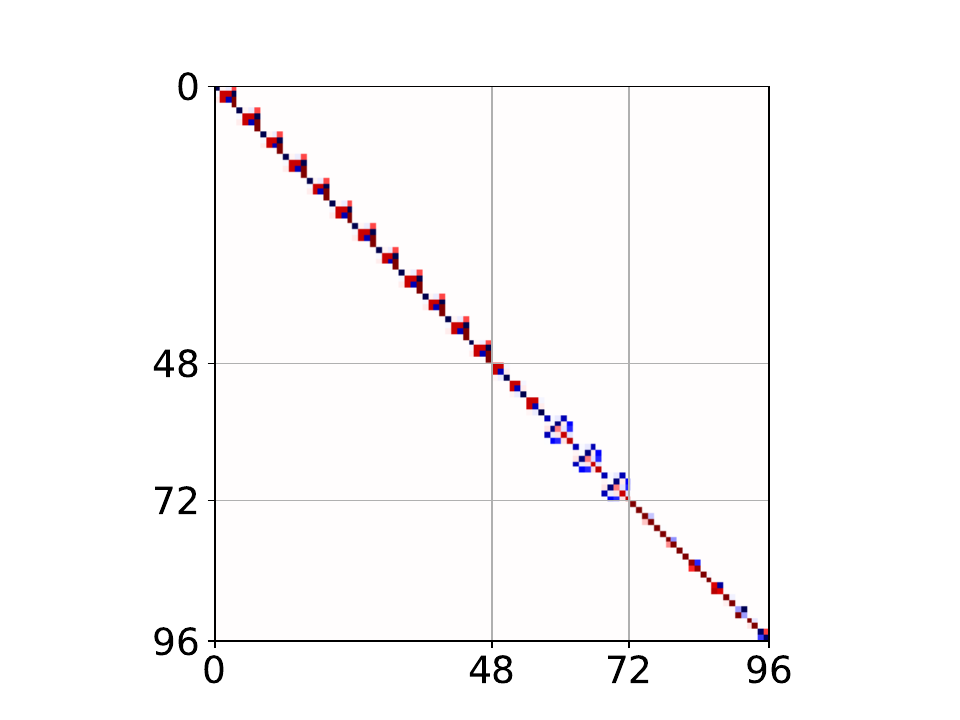}
    \includegraphics[scale=0.4]{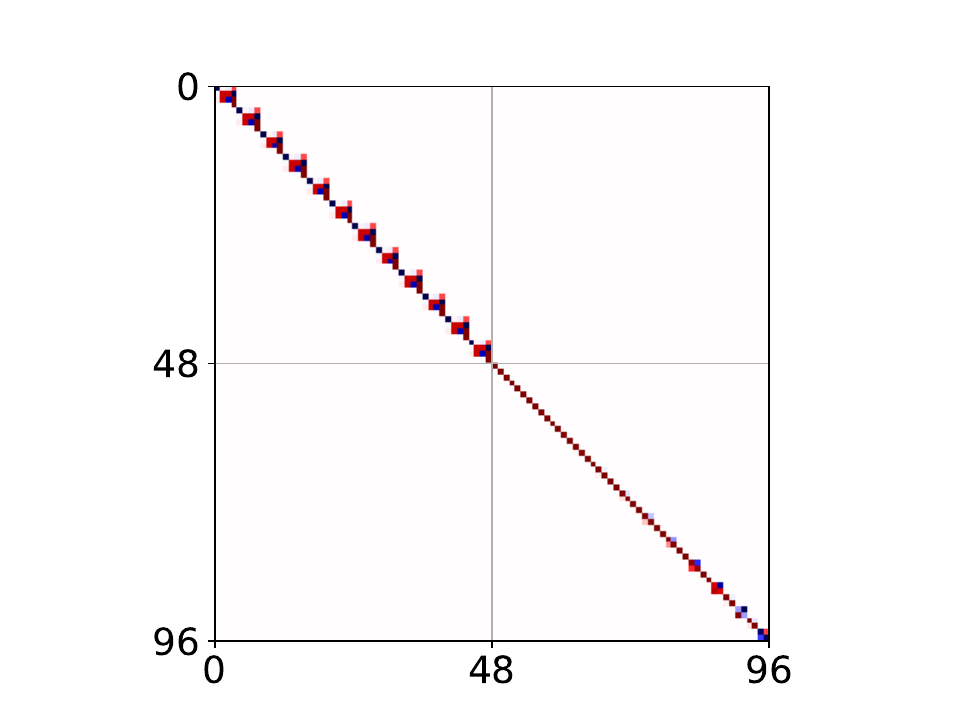}
    \caption{\textbf{Representation matrices on MSN-Hard.} Left: with $SO(3)$, Right: without $SO(3)$. Left: Dimensions 1-48 correspond to $\sigma_{\rm cam}^{\oplus 12}(c)$, dimensions 49-72 correspond to $\sigma_{\rm rot}^{\oplus 3}(r)$, and dimensions 73-96 correspond to $\sigma_{h}(\theta_h)$ and $\sigma_{w}(\theta_w)$. Right: Dimensions 1-48 correspond to $\sigma_{\rm cam}^{\oplus 12}(c)$ and dimensions 49-96 correspond to $\sigma_{h}(\theta_h)$ and $\sigma_{w}(\theta_w)$. }
    \label{fig:repmat}
\end{figure}

We use the following strategy to choose the multiplicities $s,t,u,v$ and frequencies \( \{l\}, \{f_h\}, \{f_w\} \):
\begin{tight_enumerate}
    
\item Given the feature dimension \( d \) of the attention layer, we split the dimensions into three components based on the ratio of \( 2:1:1 \).

\item
    \begin{tight_itemize}
    \item We apply \( \sigma_{\rm cam}^{\oplus s} \) to the first half of the dimensions. As \( \sigma_{\rm cam} \) does not possess multiple frequencies, its multiplicity is set to \( d/8 \).
    \item \( \sigma_{\rm rot}^{\oplus r} \) is applied to a quarter of the dimensions. For the frequency parameters \( \{l\} \) of \( \sigma_{\rm rot} \), we consistently use the 1st and 2nd degrees of the Winger-D matrices. Considering the combined sizes of these matrices is 8, the multiplicity for \( \sigma_{\rm rot} \) becomes \( d/32 \).
    \item  For the remaining \( 1/4 \) of the dimensions of each QKV vector, we apply both \( \sigma_{h}^{\oplus t} \) and \( \sigma_{w}^{\oplus u} \). Regarding the frequency parameters \( \{f_h\}, \{f_w\} \), we utilize \( d/16 \) octaves with the maximum frequency set at 1 for both \( \sigma_{h} \) and \( \sigma_{w} \). The multiplicities for these are both 1.
\end{tight_itemize}
\end{tight_enumerate}
Based on this strategy, we use the multiplicities and frequencies shown in \tabref{tab:rep_hp}.
Also \figref{fig:repmat} displays the actual representation matrices used on the MSN-Hard dataset.

\begin{table}[H]
    \centering
    \caption{\textbf{Multiplicity and frequency parameters.} Here, $d$ is the dimensions of each attention head. Since the baseline model on RealEsate10k and ACID uses different feature sizes for query-key vectors and value vectors, we also use different sizes of representation matrices for each feature.}
    \setlength\tabcolsep{1.0pt}
    \begin{tabular}{ccccc}
    \toprule
        &$d$ &  $\{s, t, u, v\}$ & $\{l_1, ..., l_{N_{\rm rot}}\}$ & $\{f_1, ..., l_{N_{\{\rm h, w\}}}\}$ \\
        \midrule
        CLEVR-TR & \multirow{2}{*}{64} &  $\{8, 3, 1, 1\}$ & $\{1, 2\}$ & $\{1,... ,1/2^3\}$\\ 
        CLEVR-TR wo/ $SO(3)$ &  &$\{8, 0, 1, 1\}$ & $-$  & $\{1,1/2,1/4,...,1/2^{7}\}$ \\
        MSH-Hard  &\multirow{2}{*}{96}& $\{12, 3, 1, 1\}$ & $\{1, 2\}$ & $\{1,1/2,1/4,...,1/2^5\}$  \\
        MSH-Hard wo/ $SO(3)$ & &$\{12, 0, 1, 1\}$ & $-$ & $\{1,1/2,1/4,...,1/2^{11}\}$ \\
        Realestate10k and ACID (Encoder)   & 64 &  $\{8, 3, 1, 1\}$ & $\{1, 2\}$ & $\{1,...,1/2^3\}$  \\
        Realestate10k and ACID (Decoder, key)  & 128 &  $\{16, 6, 1, 1\}$ & $\{1, 2\}$ & $\{1,...,1/2^7\}$  \\
        Realestate10k and ACID (Decoder, value)   & 256 &  $\{32, 12, 1, 1\}$ & $\{1, 2\}$ & $\{1,...,1/2^{15}\}$  \\
        \bottomrule
    \end{tabular}
    \label{tab:rep_hp}
\end{table}

\subsection{Variants of GTA}
Here we would like to introduce two variants of GTA. The one is the Euclidean version of GTA where we use the Euclidean distance for the attention similarity. The other one is GTA with a group representation composed of the Kronekcer product of smaller representations.
We see in \tabref{tab:kroneckergta} that the performances of those variants of GTA are a little degraded but relatively close to the original GTA. We will detail each variant in the following sections. 
\subsubsection{Euclidean GTA}\label{sec:euclidgta}
The unconventional aspect of~\eqnref{eq:rta_impl} is the presence of the transpose in the transformation of the query vectors. The transpose is necessary for having the reference coordinate invariance, and the need arises from the fact that the dot-product similarity is not invariant under $SE(3)$ transformations when the translation component is non-zero. To ensure both reference coordinate invariance and consistent transformations across the $Q, K, V$ vectors, one can utilize the Euclidean similarity for computing the attention matrix instead of the dot-product similarity. The formula for the self-attention layer with squared Euclidean distance is given by:
\begin{align}
    O  &:= {\rm Attn}_{\rm Euclid}(Q, K, V) = {\rm softmax}(\mathcal{E}(Q, K))V, \\
    &\text{where}~\mathcal{E}(Q,K) \in \mathbb{R}^{N\times N}, \mathcal{E}_{ij}(Q,K) = - \|Q_i-K_j\|_2^2
    \label{eq:vatt_euclid}.
\end{align}
Then the Euclidean version of GTA (GTA-Euclid) is written in the following form:
\begin{align}
O = \brhog \circledcirc {\rm Attn}_{\rm Euclid}\left( \brhog^{-1} \circledcirc Q, 
\brhog^{-1} \circledcirc K, \brhog^{-1} \circledcirc V\right).
\label{eq:gta_impl_euclid}
\end{align}
\eqnref{eq:gta_impl_euclid} possesses the reference coordinate invariance property since the square distance is preserved under rigid transformations. The numbers of \tabref{tab:kroneckergta} are produced under the same setting as the original GTA, except that we replaced the dot-product attention with the Euclidean attention that we describe above.

\subsubsection{Kronecker GTA}\label{sec:kroneckergta}
Another typical way to compose a representation matrix is using the Kronecker product. 
The Kronecker product of two square matrices $A, B \in \mathbb{R}^{m\times m}, \mathbb{R}^{n\times n}$ is defined as: 
\begin{align} 
A\otimes B = \begin{bmatrix}
    a_{11} B & \cdots & a_{1m} B \\
    \vdots  & \ddots  &\vdots \\
    a_{m1}B & \cdots  & a_{mm} B
\end{bmatrix} \in \mathbb{R}^{mn\times mn}.
\end{align}
The important property of the Kronecker product is that the Kronecker product of two representations is also a representation:  $(\rho_1 \otimes \rho_2)(gg') = (\rho_1 \otimes \rho_2)(g) (\rho_1 \otimes \rho_2)(g') $ where $(\rho_1\otimes \rho_2)(g) := \rho_1(g) \otimes \rho_2(g)$. We implement the Kronecker version of GTA, which we denote GTA-Kronecker, where we use the Kronecker product of the $SE(3)$ representation and $SO(2)$ representations as a representation $\rho_g$:
\begin{align}
    \rho_g = \rho_{\rm cam} (c) \otimes (\rho_{h} (\theta_h) \oplus \rho_{w} (\theta_w)), \text{where $g = (c, \theta_h, \theta_w)$}.
\end{align}
In the results presented in \tabref{tab:kroneckergta}, the multiplicity of $\rho_{\rm cam}$, $\rho_{h}$, $\rho_{g}$ are set to 1, and the number of frequencies for both $\rho_{h}$ and $\rho_{w}$ is set to 4 on CLEVR-TR and 6 on MSN-Hard.

\begin{table}[H]
        \setlength\tabcolsep{3.0pt}
        \centering
        \captionof{table}{\textbf{Results with different representation forms.}\label{tab:diffrep}}
        \begin{tabular}{lrr}
            \toprule & CLEVR-TR & MSN-Hard\\
            \midrule 
            GTA-Kronecker & 38.32 & 24.52\\
            GTA-Euclid & 38.59 & 24.75 \\
            GTA & \textbf{38.99} & \textbf{24.80}\\
            \bottomrule
        \end{tabular}%
    \label{tab:kroneckergta}
\end{table}

\subsection{Relation to equivariant and gauge equivariant networks}\label{sec:gauge}
The gauge transform used in gauge equivariant networks~\citep{Cohen2019ICML, De2021ICLR, He2021NEURIPS, Brandstetter2021ICLR} is related to the relative transform $\rho(g_i g_j^{-1})$ used in our attention mechanism.
However, the equivariant models and ours differ because they are built on different motivations. In short, the gauge equivariant layers are built to preserve the feature field structure determined by a gauge transformation. In contrast, since image features themselves are not 3D-structured, our model applies the relative transform only on the query and key-value pair in the attention mechanism but does not impose equivariance on the weight matrices of the attention and the feedforward layers. The relative transformation in GTA can be thought of as a form of guidance that helps the model learn structured features within the attention mechanism from the initially unstructured raw multi-view images.

\citet{Brehmer2023NEURIPS} introduce geometric algebra (GA) to construct equivariant transformer networks. Elements of GA are themselves operators that can act on GA, which may enable us to construct expressive equivariant models by forming bilinear layers that allow interactions between different multi-vector subspaces. In NVS tasks, where the input consists of raw images lacking geometric structure, directly employing such equivariant models may not be straightforward. However, integrating GA into the GTA mechanism could potentially enhance the network's expressivity, warranting further investigation.

\section{Additional experimental results}

\subsection{Training curves on CLEVR-TR and MSN-Hard.}
\begin{figure}[H]
    \centering
    \setlength\tabcolsep{3.0pt}
    \renewcommand{\arraystretch}{0}
    \begin{tabular}{cccc}
        \includegraphics[scale=0.3]{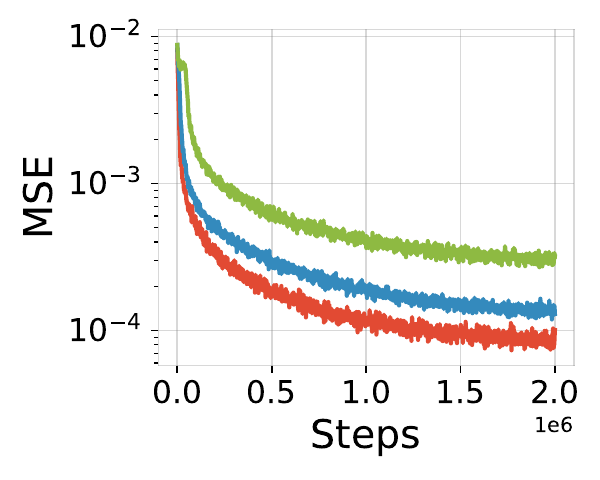}&
        \includegraphics[scale=0.3]{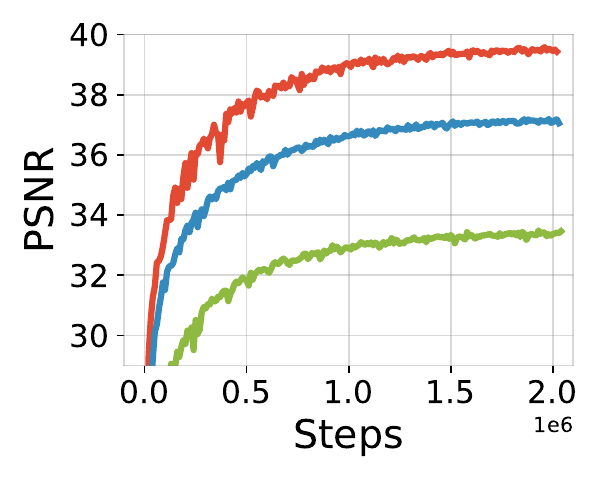}&
        \includegraphics[scale=0.3]{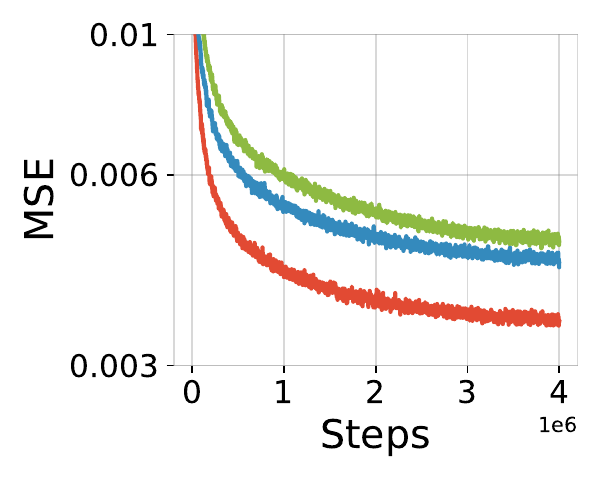} &
        \includegraphics[scale=0.3]{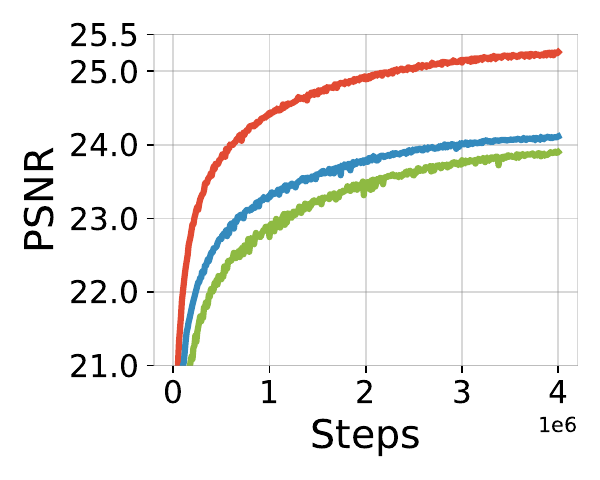}\\
        \multicolumn{2}{c}{\includegraphics[scale=0.26]{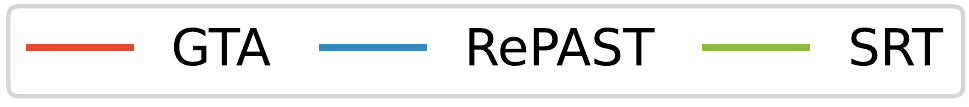}} &
        \multicolumn{2}{c}{\includegraphics[scale=0.26]{gfx/legend_processed.pdf}}\\
    \end{tabular}
    \caption{\textbf{Training and validation curves.} Left: CLEVR-TR, Right: MSN-Hard.}
    \label{fig:psnr}
    \vspace{0cm}
\end{figure}

\subsection{Results with higher resolution}
\tabref{tab:realestate384} shows results on RealEstate10k with 384x384 resolutions (1.5 times larger height and width than in \tabref{tab:comp_nvs_2}). We see that GTA also improves over the baseline model at higher resolution. 
\begin{table}[H]
    \setlength\tabcolsep{3.0pt}
    \centering
    \caption{\textbf{384$\times$384 resolution on RealEstate10K}\label{tab:realestate384}}
    \begin{tabular}{lrrr}
        \toprule
            & PSNR$\uparrow$ & LPIPS $\downarrow$ & SSIM$\uparrow$ \\
        \midrule 
        \citet{Du2023CVPR} & 21.77  & 0.316 & 0.848 \\
         \citet{Du2023CVPR} + GTA (Ours)& \textbf{22.77} & \textbf{0.290} & \textbf{0.864} \\
        \bottomrule
    \end{tabular}
\end{table}

\subsection{Results with 3 context views}
We train models with 3-context views and show the results in \tabref{tab:realestate_nv3}. We see that GTA is also better with context views more than 2.
\begin{table}[H]
    \centering
    \caption{\textbf{Results with different numbers of context views on RealEstate10K}\label{tab:realestate_nv3}}
    \begin{tabular}{lrr}
        \toprule
            & 2-view PSNR$\uparrow$ & 3-view PSNR$\uparrow$ \\
        \midrule 
        \citet{Du2023CVPR} & 21.65  & 21.88 \\
          \citet{Du2023CVPR} + GTA (Ours)& \textbf{22.85} & \textbf{23.22} \\
        \bottomrule
    \end{tabular}%
\end{table}

\subsection{Robustness to camera noise}
\tabref{tab:cnoises} shows results on CLEVR-TR with a presence of camera noise. We train RePAST and GTA with camera noise added to each camera extrinsic of the second view. We perturb camera extrinsics by adding  Gaussian noise to the coefficients of the $SE(3)$-Lie algebra basis. The mean and variance of the noise is set to $(m, \sigma) = (0, 0.1)$ during training. GTA shows better performance than RePAST regardless of the noise level. 
\begin{table}[H]
    \centering
    \caption{\textbf{Test PSNRs with camera noise on CLEVR-TR and MSN-Hard}. $\sigma_{\rm test}$ indicates the noise strength at test time.\label{tab:cnoises} ${}^\dagger$No noise injection during training. }
    \begin{tabular}{lrrrrr}
        \toprule &\multicolumn{2}{c}{CLEVR-TR} &\multicolumn{2}{c}{MSN-Hard}\\
           $\sigma_{\rm test}$ & 0.01 &  0.1 & 0.01 &  0.1  \\
        \midrule 
         RePAST~\citep{Safin2023ARXIV} & 35.26  & 35.14 & 22.76 & 22.60  \\
         SRT+GTA (Ours)& \textbf{36.66} & \textbf{36.57} & \textbf{24.06} & \textbf{24.16}  \\
        \bottomrule
    \end{tabular}
\end{table}

\begin{table}
    \centering
    \caption{\textbf{Test metrics on CLEVR-TR.}\label{tab:clevrtr}}
    \begin{tabular}{lrrr}
        \toprule
            & PSNR$\uparrow$ & LPIPS $\downarrow$ & SSIM$\uparrow$ \\
        \midrule 
        APE & 33.66 & 0.161 & 0.960 \\
        RPE & 36.06 & 0.159 &  0.971\\
        SRT & 33.51 & 0.158 & 0.960 \\
        RePAST & 37.27 &0.119 &0.977 \\
         GTA (Ours)& \textbf{39.63} & \textbf{0.108} & \textbf{0.984} \\
        \bottomrule
    \end{tabular}%
\end{table}

\subsection{Performance with different random seeds}
\begin{figure}
    \centering
    \includegraphics[scale=0.6]{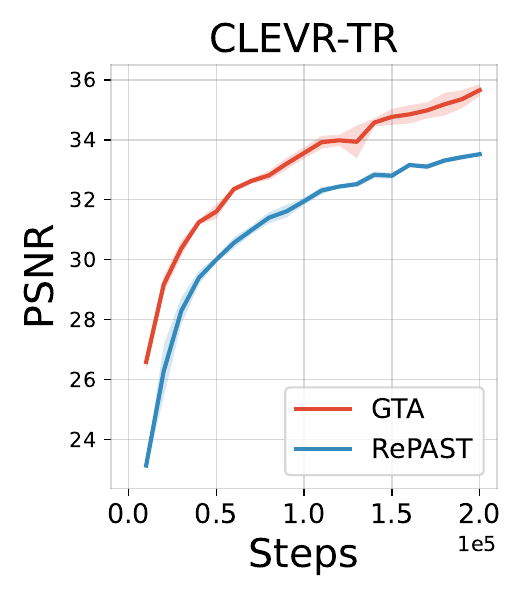}
    \includegraphics[scale=0.6]{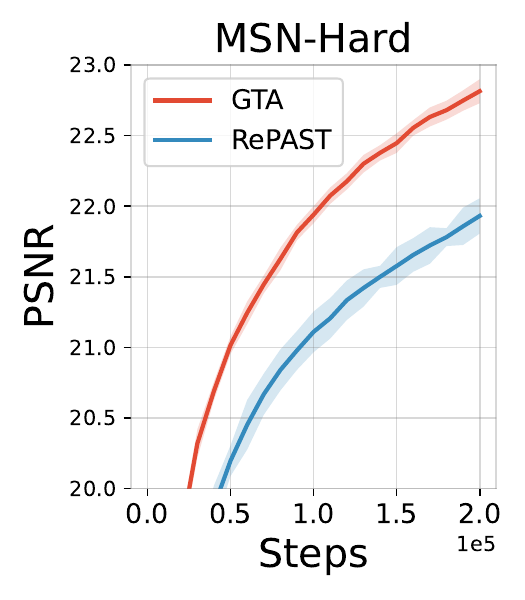}
    \caption{\textbf{Mean and standard deviation plots of validation PSNRs on CLEVR-TR and MSN-Hard}. Due to the heavy computation requirements for training, we only trained models with 200,000 iterations and measured the validation PSNRs during the course of the training.}
    \label{fig:seed_variance}
\end{figure}

We observe that the performance variance of different random weight initializations is quite small, as shown in \figref{fig:seed_variance}, which displays the mean and standard deviation across 4 different seeds. We see that the variance is relatively insignificant compared to the performance difference between the compared methods. Consequently, the results reported above are statistically meaningful.

\subsection{Performance dependence on the reference coordinates} \tabref{tab:comp_coord} highlights the importance of coordinate-choice invariance. ``SRT (global coord)" is trained with camera poses that have their origin set to always be in the center of all objects. This setting enables the model to know how far the ray origin is from the center of the scene, therefore enabling the model to easily find the position of the surface of objects that intersect with the ray. We see that SRT's performance heavily depends on the choice of reference coordinate system. Our model is, by construction, invariant to the choice of reference coordinate system of cameras and outperforms even the privileged version of SRT.
\begin{table}[H]
    \centering
    \caption{\label{tab:comp_coord} \textbf{Test PSNRs in a setting where global coordinates are shared across scenes.} All numbers show test PSNRs and are produced with models trained for 1M iterations. Note that GTA is invariant to the reference coordinates of the extrinsics, and the performance is not affected by the choice of the reference coordinate system. }
    \centering
    \begin{tabular}{lrr}
        \toprule
        Method&  CLEVR-TR &  MSH-Hard    \\
        \midrule 
        SRT & 32.97 & 23.15\\   
        SRT (global coord) & 37.93 & 24.20\\
        GTA wo $SO(3)$ & \textbf{38.99} & \textbf{24.58}\\
        \bottomrule
    \end{tabular}  
\end{table}

\begin{table}[t]
    \centering
    \caption{\textbf{Computational time to perform one gradient step, encode a single scene, and render a single entire image on MSN-Hard (top) and RealEstate10K (bottom)}. All time values are expressed in milliseconds (ms).  As for one gradient step time, we only measure time for forward-backward props and weight updates and exclude data loading time. We measure each time on a single A100 with bfloat16 precision for MSN-Hard and float32 precision for RealEstate10K. 
    }
    \begin{tabular}{lrrr}
        \toprule 
        Method & One gradient step&  Encoding & Rendering\\
        \midrule
         SRT & 296& 5.88 & 16.4\\
         RePAST & 394 & 7.24 & 21.4 \\
         GTA & 379 & 17.7 & 20.9 \\
         \bottomrule
    \end{tabular}%
    \vspace{5mm}
    \begin{tabular}{lrrr}
        \toprule 
        Method & One gradient step &  Encoding & Rendering\\
        \midrule
         \citet{Du2023CVPR}& 619 & 49.8 & 1.42$\times 10^{3}$ \\
         GTA & 806 & 74.3 &  2.05$\times 10^{3}$ \\
         \bottomrule
    \end{tabular}

    \label{tab:time}
\end{table}

\subsection{Analysis of attention patterns}\label{sec:attn_patterns}
\begin{figure}[t]
    \centering
    \begin{tabular}{cc}
        GTA & RePAST  \\
    \midrule
    \includegraphics[scale=0.075]{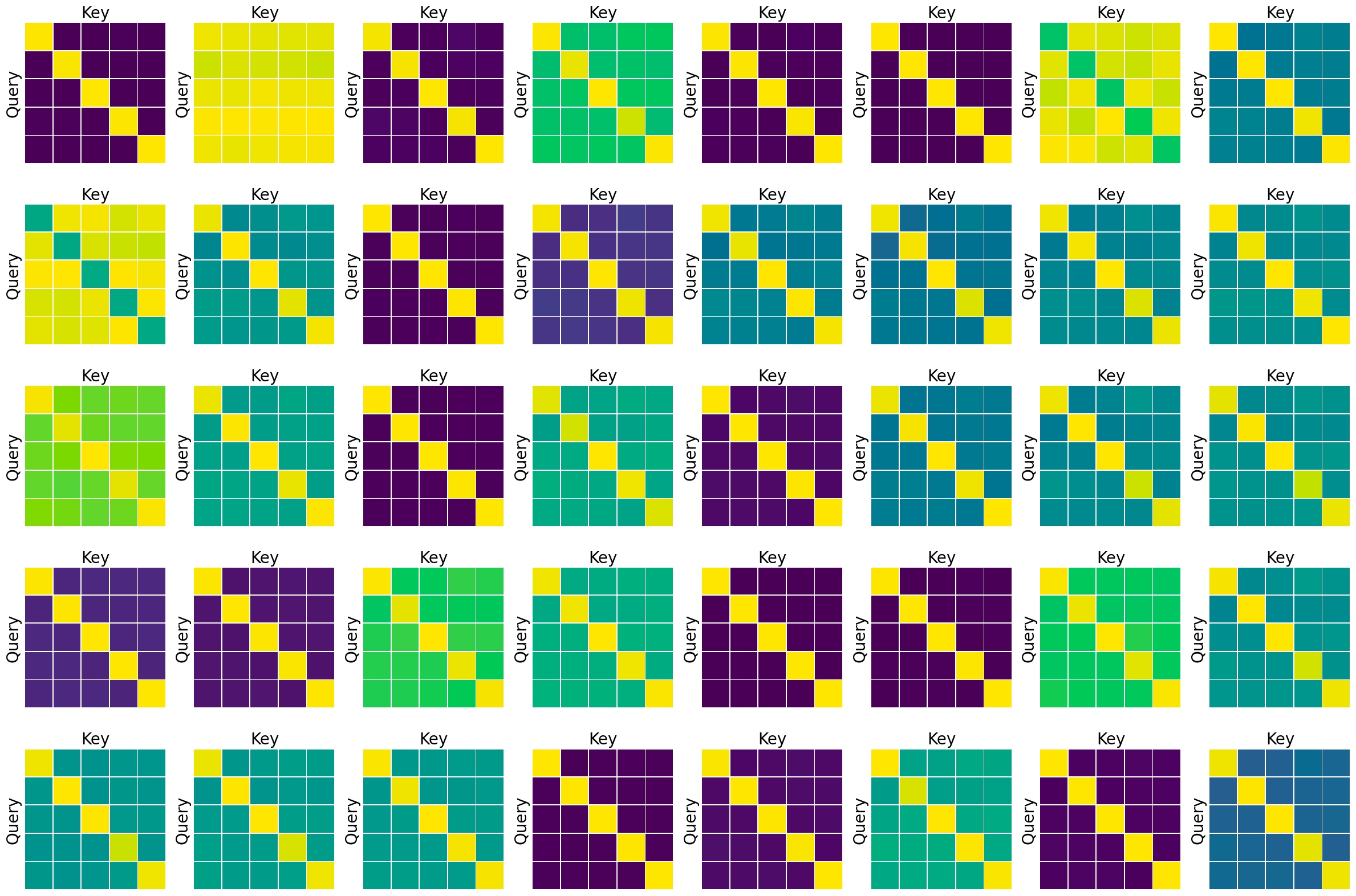} &\includegraphics[scale=0.075]{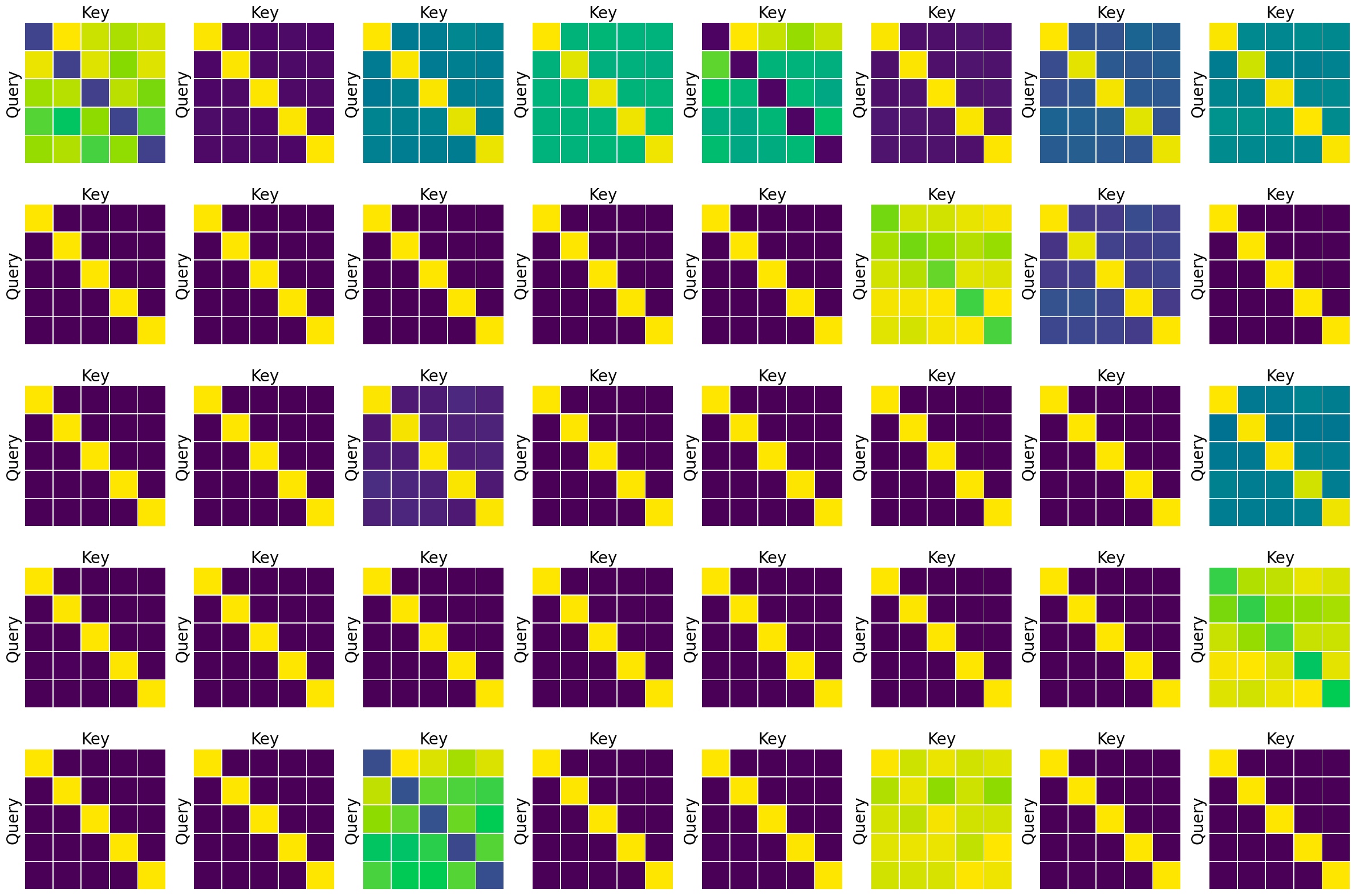}
    \end{tabular}
    \caption{\textbf{Visualization of view-to-view attention maps}. \label{fig:attn_pattern}
    The $(i, j)$-th element of each 5x5 matrix represents the average of attention weights between all pairs of each query token of the $i$-th context view and each key token of the $j$-th context view. The $(l, m)$-th panel shows the weight of the $m$-th head at the $l$-th layer. Yellow and dark purple cells indicate high and low attention weight, respectively. 
    A matrix with high diagonal values means that the corresponding attention head attends \textit{within each view} while with high non-diagonal values means the corresponding attention head attends \textit{across views}.}
\end{figure}

\begin{figure}
    \centering
    \begin{tabular}{ccccc}
        \toprule
        GTA \\
        \includegraphics[scale=0.25]{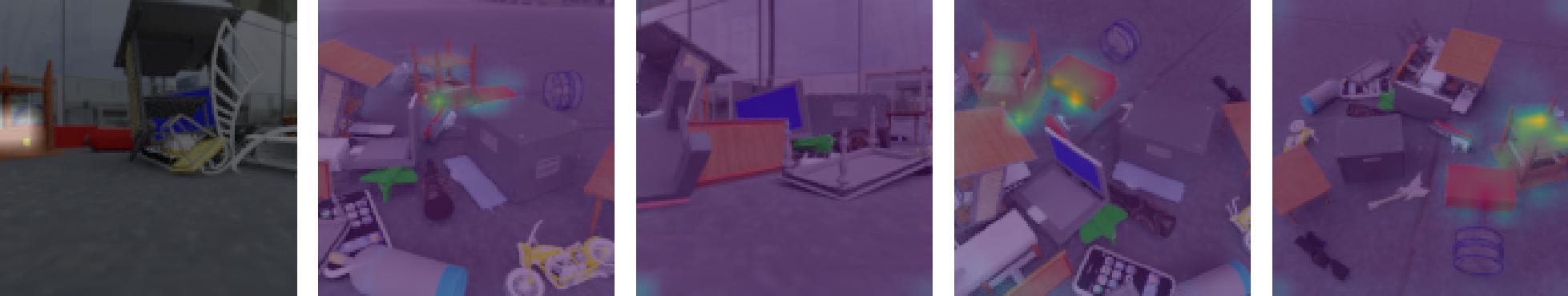}\\
        RePAST\\
        \includegraphics[scale=0.25]{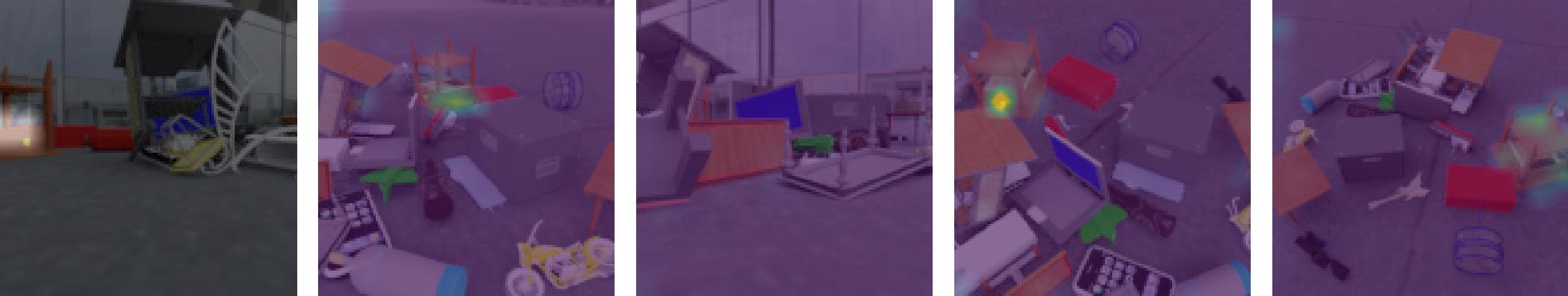}\\
        \midrule
        GTA \\
        \includegraphics[scale=0.25]{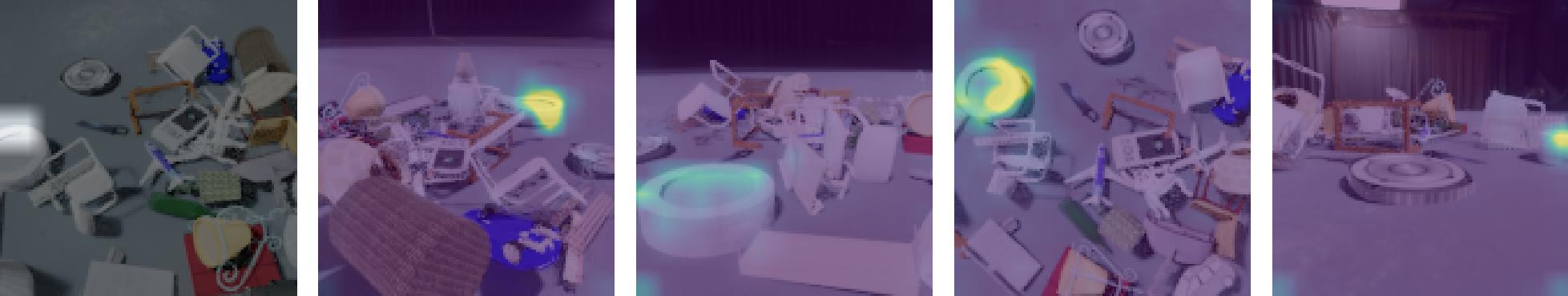}\\
        RePAST\\
        \includegraphics[scale=0.25]{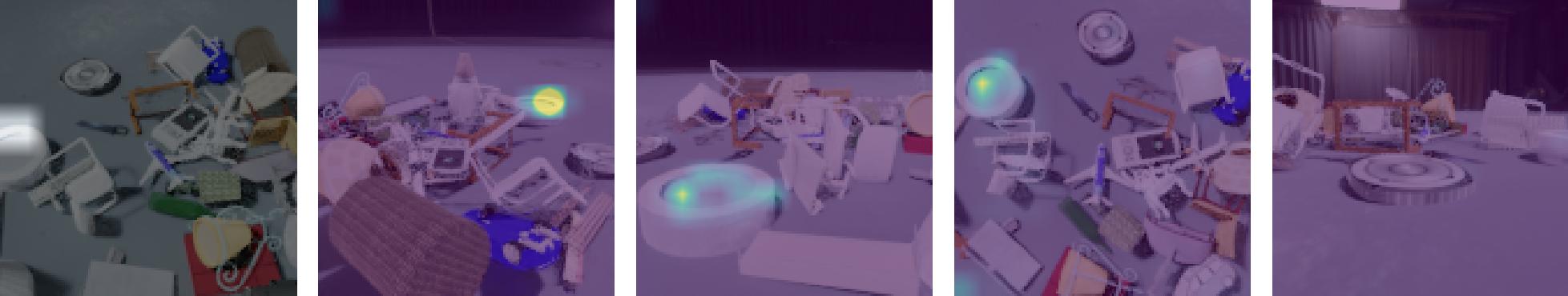}\\
        \midrule
        GTA\\
        \includegraphics[scale=0.25]{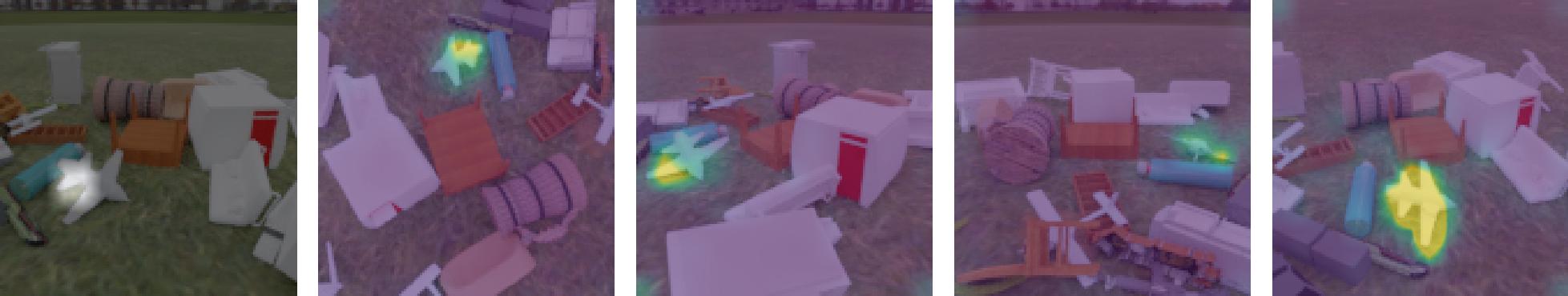}\\
        RePAST\\
        \includegraphics[scale=0.25]{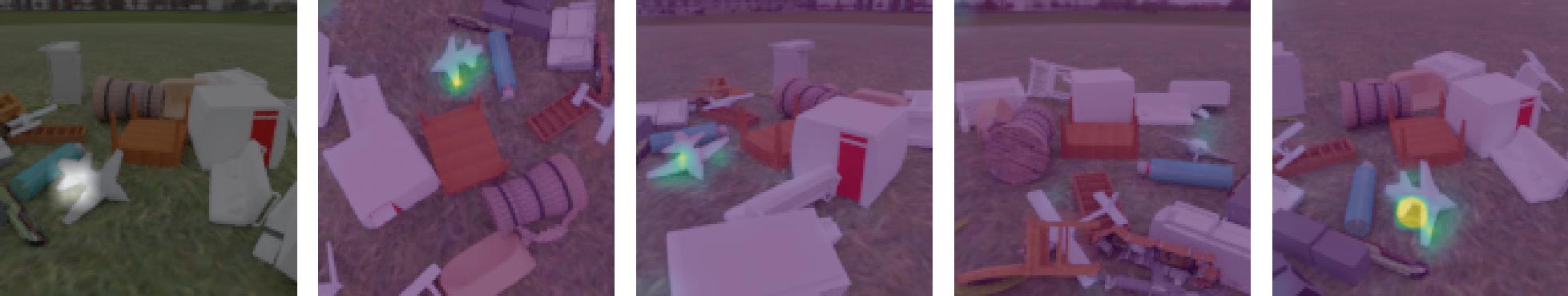}\\
        \bottomrule
    \end{tabular}
    \caption{\textbf{Additional attention map visualizations.}}
    \label{fig:attn_vis_app}
\end{figure}

\begin{figure}
    \centering
    \begin{tabular}{ccccc}
        \toprule
         Layer$=1$ & 2 & 3 & 4 & 5  \\
         \midrule
        \includegraphics[scale=0.23]{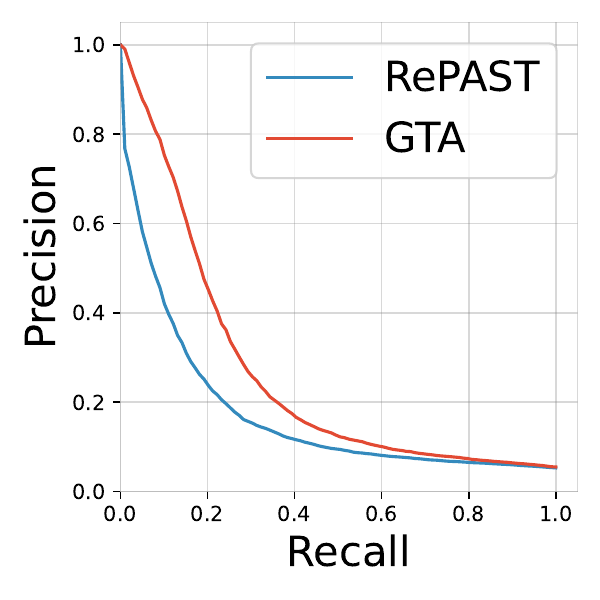} &
        \includegraphics[scale=0.23]{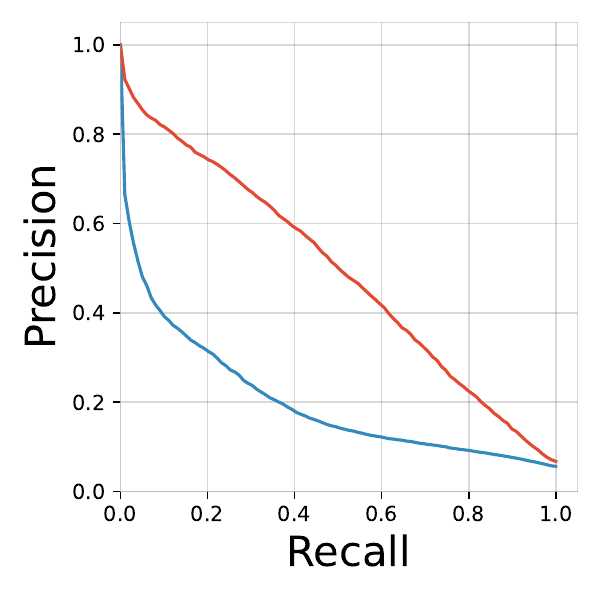} &
        \includegraphics[scale=0.23]{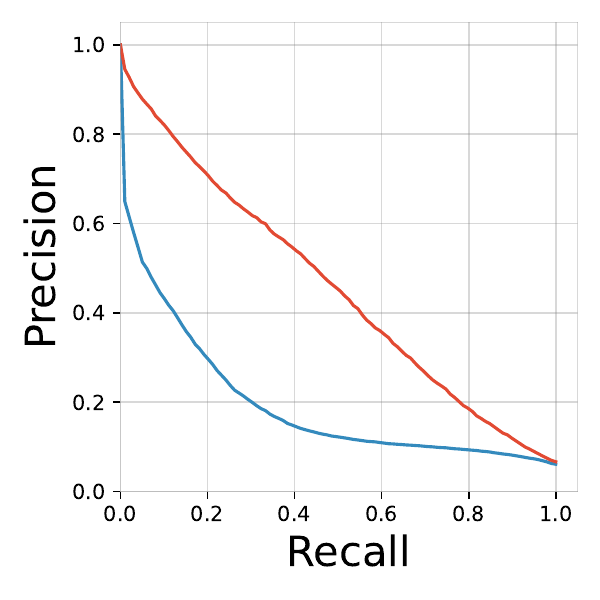} &
        \includegraphics[scale=0.23]{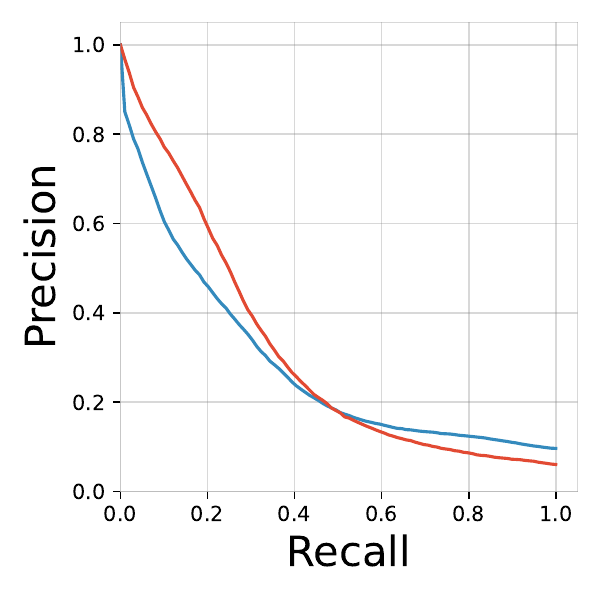} &
        \includegraphics[scale=0.23]{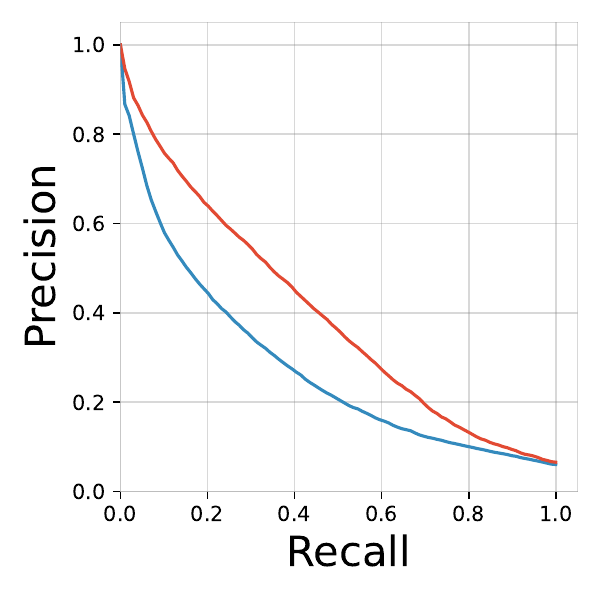} \\
        \bottomrule
    \end{tabular}
    \caption{\textbf{Precision-recall curves of the attention matrices of each encoder layer}. }
    \label{fig:prcurves}
\end{figure}

We conducted an analysis on the attention matrices of the encoders trained on MSN-Hard.
We found that the GTA-based model tends to attend to features of different views more than RePAST, which we show in \figref{fig:attn_pattern}. 
Furthermore, we see that GTA not only correctly attends to the respective patches of different views, but also can attend to object level regions (\figref{fig:attn_vis} and \ref{fig:attn_vis_app}). Surprisingly, these attention patterns are seen at the very beginning of the course of the encoding process: the visualized attention maps are obtained in the 2nd attention layer. To evaluate how well the attention maps $\alpha$ weigh respective object features across views,  we compute a retrieval-based metric with instance segmentation masks of objects provided by MSN-Hard. Specifically, given a certain layer's attention maps $\alpha$:
\begin{tight_enumerate}
        \item We randomly sample the $i$-th query patch token with 2D position $p \in\{1,..., 16\} \times \{1,..., 16\}$. 
        \item We compute the attention map $\bar{\alpha}_i \in [0, 1]^{5^*16^*16}$ averaged over all heads. 
        \item We then identify which object belongs to that token's position by looking at the corresponding $8\times 8$ region of the instance masks. Note that multiple objects can belong to the region. 
        \item For each belonging object, we compute precision and recall values with $\mathbbm{1}[\bar{\alpha}_i > t]$ as prediction and 0–1 masks of the corresponding object as ground truth on all context views, by changing the threshold value $t\in[0, 1]$. \
        \item In the final step, we calculate a weighted average of the precision and recall values for each object. To determine the weight of each object, we consider the number of pixels assigned to that object's mask within the 8x8 region. We then normalize these weights so that their sum equals to be one. 
\end{tight_enumerate}
We collect multiple precision and recall values by randomly sampling scenes and patch positions 2000 times and then compute the average of the collected precision-recall curves. In \figref{fig:prcurves}, we show averaged precision-recall curves. \tabref{fig:attn_vis} shows the area under the precision-recall curves (PR-AUCs) of each layer. We see that the GTA-based model learns well-aligned attention maps with the ground truth object masks for every layer.

\subsection{Computational time }\label{sec:time}
We measure the time to perform one-step gradient descent, as well as encoding and decoding for each method. \tabref{tab:time} shows that the computational overhead added by the use of GTA is comparable to RePAST on MSN-Hard. In contrast to GTA and RePAST-based models which encode positional information into every layer, SRT and ~\citet{Du2023CVPR} add positional embeddings only to each encoder and decoder input. As a result, the computational time of SRT for one-step gradient descent is around 1.3x faster than RePAST and GTA, and that of ~\citet{Du2023CVPR} is 1.3x faster than GTA.

\section{Experimental settings}

\subsection{Details of the synthetic experiments in \secref{sec:gta}}\label{sec:synthexp}
We use $10,000$ training and test scenes. For the intrinsics, both the vertical and horizontal sensor width are set to 1.0, and the focal length is set to 4.0, leading to an angle of view of 28\textdegree.   

For optimization, we use AdamW~\citep{Loshchilov2017ARXIV} with weight decay $0.001$. For each PE method, we trained multiple models with different learning rates of \{0.0001, 0.0002, 0.0005\} and found 0.0002 to work best for all models, and hence show results with this learning rate.
We use three attention layers for both the encoder and the decoder. The image feature dimension is $32 \times 32 \times 3$. This feature is flattened and fed into a 2 layer-MLP to be transformed into the same dimensions as the token dimension $d$. We also apply a 2 layer-MLP to the output of the decoder to obtain the $3,072$ dimensional predicted image feature.
The token dimensions $d$ are set to 512 for APE and RPE.
As we mention in the descriptions of the synthetic experiment, $\rho_g$ is composed of block concatenation of $3\times 3$ rotation matrices, and we set $d$ to 510 for GTA, which is divisible by 3.
Note that there is no difficulty with the case where $d$ is not divisible by 3. In that case, we can apply $\rho_g$ only to certain components of vectors whose dimensions are divisible by 3 and apply no transformation to the other dimensions. This corresponds to applying a trivial representation, \ie, the identity matrix, to the remaining vectors.

The RPE-based model we designed is a sensible model. For example, if $b^{Q}=b^{K}$ and the set of three-dimensional vector blocks of $b^{Q}$ forms an orthonormal basis, then the inner product of the transformed query and key bias vectors becomes the \textit{trace} of the product of the rotation matrices: $\langle \rho(r)b^{Q}, \rho(r')b^{K} \rangle = {\rm tr}(R^{\rm T}R')$. ${\rm tr}(A^{\rm T}B)$ is a natural inner product for matrices, by which we can bias the attention weight based on the inner-product-based similarity of matrices. Hence, we initialize each of the biases with vectorized identity matrices.

\subsection{Experimental settings in \secref{sec:experiment}}\label{sec:expsettings}


\begin{table}[t]
    \centering
    \caption{\textbf{Dataset properties and architecture hyperparameters}. \# target pixels indicate how many query pixels are sampled for each scene during training.
    $^*$We use 12 heads for the attention layers in SRT and 8 heads in RePAST and GTA because 12 head models do not fit into our GPU memory with those methods.  ${}^\dagger$The decoder's attention layers only have 
    a single head. Also, the token dimensions in the decoder are set to 128 for query-key vectors and 256 for value vectors.}
    \setlength\tabcolsep{3.0pt}
    \begin{tabular}{ccccc}
    \toprule
     dataset & CLEVR-TR & MSN-Hard & RealEstate10k & ACID \\
     \midrule
    \# Training scenes & 20,000 & 1,000,000 & 66,837 & 10,974  \\
     \# Test scenes & 1,000 & 10,000 &7,192 & 1,910 \\
      Batch size   & 32 & 64 & 48 & 48 \\
      Training steps & 2,000,000 & 4,000,000 & 300,000 & 200,000 \\ 
      Learning rate &1e-4 & 1e-4 & \multicolumn{2}{c}{5e-4} \\
       \# Context views & 2 & 5 & \multicolumn{2}{c}{2} \\ 
      \# Target pixels & 512 & 2,048 & \multicolumn{2}{c}{192} \\
      \# Self-attention layers in the encoder & 5 & 5 & \multicolumn{2}{c}{12} \\
      \# Cross-attention layers in the decoder & 2 & 2 &  \multicolumn{2}{c}{2} \\
      \# Heads in attention layers & 6 & 12/8$^{*}$ & \multicolumn{2}{c}{12$^{\dagger}$} \\
     Token dimensions & 384 & 768 & \multicolumn{2}{c}{768$^\dagger$} \\
      MLP dimensions & 768 & 1,536 & \multicolumn{2}{c}{3,072} \\
      \bottomrule
    \end{tabular}
    
    \label{tab:hp}
\end{table}
\tabref{tab:hp} shows dataset properties and hyperparameters that we use in our experiments. 
We train with 4 RTX 2080 Ti GPUs on CLEVR-TR and with 4 Nvidia A100 GPUs on the other datasets.

\paragraph{CLEVR-TR and MSN-Hard} CLEVR-TR is synthesized by using {\rm Kubric}~\citep{Greff2022CVPR}. The resolution of each image is $240\times 320$. The camera poses of the dataset include translation, azimuth, and elevation transformations. The camera does not always look at the center of the scene. 

MSN-Hard is also a synthetically generated dataset. Up to 32 objects sampled from ShapeNet~\citep{Chang2015} appear in each scene. All 51K ShapeNet objects are used for this dataset, and the training and test sets do not share the same objects with each other. MSN-Hard includes instance masks for each object in a scene, which we use to compute the attention matrix alignment score described in \secref{sec:experiment} and Appendix~\ref{sec:attn_patterns}.
The resolution of each image is $128\times 128$.

We basically follow the same architecture and hyperparameters of the improved version of SRT described in the appendix of \citet{Sajjadi2022NEURIPS}, except that we use AdamW~\citep{Loshchilov2017ARXIV} with the weight decay set to the default parameter and dropout with a ratio of 0.01 at every attention output and hidden layers of feedforward MLPs. 

Since there is no official code or released models available for SRT and RePAST, we train both baselines ourselves and obtain almost comparable but slightly worse results (\tabref{tab:comp_nvs_srt}). 
This is because we train the models with a smaller batch size and target ray samples than in the original setting due to our limited computational resources (4 A100s). Note that our model, which is also trained with a smaller batch size, still outperforms the original SRT and RePAST models' scores.

\begin{table}[t]
    \centering
        \caption{\textbf{Performance comparison between numbers reported in \citet{Safin2023ARXIV} and our reproduced numbers}. Note that \citet{Safin2023ARXIV} uses 4x larger batch size than available in our experimental setting (4 A100s). The number of iterations for which we train each model is the same as \citet{Safin2023ARXIV}.\label{tab:comp_nvs_srt}}
    \setlength\tabcolsep{3.0pt}
     \begin{tabular}{lrrr}
        \toprule
             & PSNR$\uparrow$ & LPIPS$_{\rm VGG/Alex}$ $\downarrow$ & SSIM$\uparrow$\\
        \midrule 
        SRT~\citep{Sajjadi2022CVPR} &  24.56 & NA/0.223 & 0.784 \\
        RePAST~\citep{Safin2023ARXIV} & 24.89  & NA/0.202 & 0.794\\
        \midrule 
        SRT & 24.27 & 0.368/0.279 &0.741\\
        RePAST & 24.48 & 0.348/0.243 & 0.751 \\
        SRT+GTA (Ours) & \textbf{25.72} &\textbf{0.289}/\textbf{0.185}& \textbf{0.798} \\
    \bottomrule
    \end{tabular}%
\end{table}

\paragraph{RealEstate10k and ACID}
Both datasets are sampled from videos available on YouTube. At the time we conducted our experiments, some of the scenes used in \citet{Du2023CVPR} were no longer available on YouTube. We used scenes $66,837$ and $10,974$ training scenes and $7,192$ and $1,910$ test scenes for RealEstate10k and ACID, respectively.
The resolution of each image in the original sequences is $360\times 640$. For training, we apply downsampling followed by a random crop and random horizontal flipping to each image, and the resulting resolution is $256 \times 256$.
For test time, we apply downsampling followed by a center crop to each image.
The resolution of each processed image is also $256\times 256$.
We follow the same architecture and optimizer hyperparameters of ~\citet{Du2023CVPR}.
Although the authors of \citet{Du2023CVPR} released the training code and their model on RealEstate10k, we observed that the model produces worse results than those reported in their work. The results were still subpar even when we trained models with their code. As a result, we decided to train each model with more iterations (300K) compared to the 100K iterations mentioned in their paper and achieved comparable scores on both datasets. Consequently, we also trained GTA-based models for 300K iterations as well.

\begin{table}[]
    \centering
    \caption{\textbf{Comparison between results reported in \citet{Du2023CVPR} (Top) and our reproduced results (Bottom).} \label{tab:comp_nvs_du} }
    \begin{tabular}{lrrrrrr}
        \toprule
             &\multicolumn{3}{c}{RealEstate10k} &
             \multicolumn{3}{c}{ACID}  \\  &PSNR$\uparrow$&LPIPS$\downarrow$&SSIM$\uparrow$&PSNR$\uparrow$&LPIPS$\downarrow$&SSIM$\uparrow$\\
        \midrule 
        \citet{Du2023CVPR} & 21.38 & 0.262 & 0.839 & 23.63  & 0.364 & 0.781 \\
        \midrule 

        \citet{Du2023CVPR}   & 21.65 & 0.284 & 0.822 & 23.35  & 0.334 &  0.801  \\ 
        \citet{Du2023CVPR} + GTA~(Ours) & \textbf{22.85} & \textbf{0.255} & \textbf{0.850} &  \textbf{24.10} &   \textbf{0.291} & \textbf{0.824} \\
        \bottomrule
        
        \end{tabular}%
   
    \label{tab:my_label}
\end{table}

\subsubsection{Scene representation transformer (SRT)}\label{sec:srt}
\begin{figure*}
    \centering
    \includegraphics[scale=0.18]{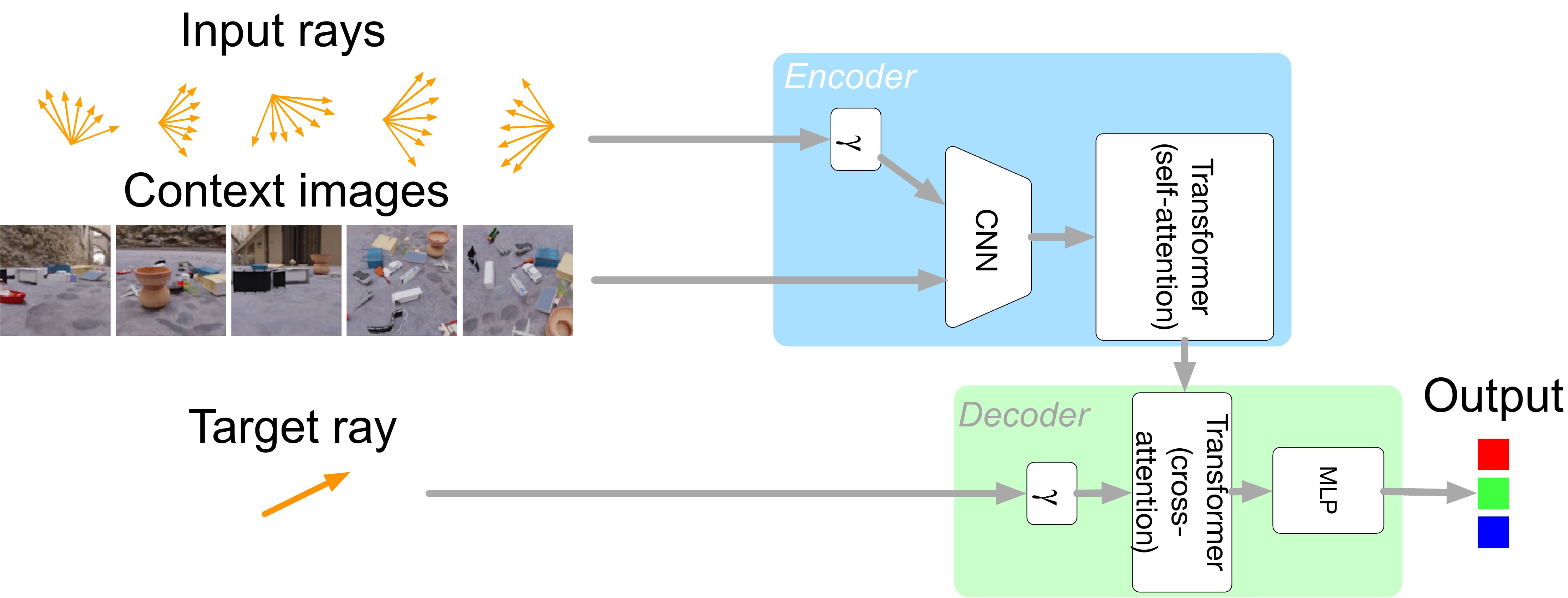}
    \caption{\textbf{Scene representation transformer (SRT) rendering process}. The encoder $E$ consisting of a stack of convolution layers followed by a transformer encoder translates context images into a set representation $S$. The decoder $D$ predicts an RGB pixel value given a target ray and $S$. 
    In our model, every attention layer in both the encoder and decoder is replaced with GTA. We also remove the input and target ray embeddings from the input of the encoder and decoder, respectively. We input a learned constant vector to the decoder instead of the target ray embeddings.\label{fig:srt_model}}
\end{figure*}

\boldparagraph{Encoding views}
Let us denote $N_{\rm context}$-triplets of input view images and their associated camera information by ${\bf I}:=\{(I_i, c_i, M_i)\}_{i=1}^{N_{\rm context}}$, where $N_{\rm context}$ is the number of context views, $I_i\in R^{H\times W \times 3}$ is the $i$-th input RGB image, and $c_i \in \mathbb{R}^{4\times 4}, M_i \in \mathbb{R}^{3\times 3}$ are a camera extrinsic and a camera intrinsic matrix associated of the $i$-th view. 
The SRT encoder $E$ encodes the context of views into scene representation $S$ and is composed of a CNN and a transformer $E_{\rm transformer}$. First, a 6-layer CNN $E_{\rm CNN}$ is applied to a ray-concatenated image $I'$ of each view to obtain $(H/D)\times (W/D)$-resolution features:
\begin{align}
    F_i= E_{\rm CNN}(I_i') \in \mathbb{R}^{(H/D)\times (W/D) \times d}, ~I_{ihw}' = I_{ihw} \oplus \gamma(r_{ihw})\label{eq:cnn} 
\end{align}
where $d$ is the output channel size of the CNN, and $D$ is the downsampling factor, which is set to $8$. $\gamma$ is a Fourier embedding function that transforms ray $r = (o, d)\in \mathbb{R}^3\times \mathcal{S}$ into a concatenation of the Fourier features with multiple frequencies. Each ray $r_{ihw}$ is computed from given camera's extrinsic and intrinsic parameters ($c_i$, $M_i$). Here, ``$\oplus$" denotes vector concatenation. 

Next, a transformer-based encoder $E_{\rm transformer}$ processes the flattened CNN features of all views together to output the scene representation:
\begin{align}
    S := \{s_i\}_{i=1}^{N_{\rm context}{}^*(H/D)^*(W/D)} = E_{\rm transfomer}\left(\{f_i\}_{i=1}^{N_{\rm context}{}^*(H/D)^*(W/D)}\right) \label{eq:transformer}
\end{align}
where $\{f_i\}$ is the set of flattened CNN features. 

\boldparagraph{Rendering a view}
Given the scene representation $S$ and a target ray $r^*$, the decoder $D$ outputs an RGB pixel:
\begin{align}
    \hat{a}_{r^*} = D(\gamma(r^*), S) \in \mathbb{R}^{3}.
\end{align}
where $\gamma$ is the same function used in \eqnref{eq:cnn}.
The architecture of $D$ comprises two stacks of a cross-attention block followed by a feedforward MLP.
The cross-attention layers determine which token in the set $S$ to attend to, to render a pixel corresponding to the given target ray. The output of the cross-attention layers is then processed by a 4-layer MLP, to get the final RGB prediction.
The number of hidden dimensions of this MLP is set to $1536$.

\boldparagraph{Optimization}
The encoder and the decoder are optimized by minimizing the mean squared error between given target pixels $a_r$ and the predictions:
\begin{align}
    \mathcal{L}(E, D) = \sum_{r^{*}}  \| a_{r^{*}} - \hat{a}_{r^{*}} \|^2_2.
\end{align}

\subsubsection{Details of the architecture and loss of ~\citet{Du2023CVPR}}
\citet{Du2023CVPR} proposes an SRT-based transformer NVS model with a sophisticated architecture. 
The major differences between their model and SRT are that they use a dense vision transformer~\citep{Ranftl2021ICCV} for their encoder. They also use an epipolar-based sampling technique to select context view tokens, a process that helps render pixels efficiently in the decoding process. 

We use the same optimization losses for training models based on this architecture as ~\citet{Du2023CVPR}.
Specifically, we use the $L_1$ loss between target and predicted pixels on RealEstate10k and ACID.
We also use the following combined loss after the 30K-th iterations on ACID.
\begin{align}
    L_1(P, \hat{P}) + \lambda_{\rm LPIPS} L_{\rm LPIPS}(P, \hat{P}) + \lambda_{\rm depth}L_{\rm depth}(P, \hat{P})
\end{align}
where $P, P'\in \mathbb{R}^{32\times 32\times 3}$ are target and predicted patches. $L_{\rm LPIPS}$ is the perceptual similarity metric proposed by~\citet{Zhang2018CVPRa}. $L_{\rm depth}$ is a regularization loss that promotes the smoothness of estimated depths in the model. Please refer to~\citet{Du2023CVPR} for more details.
On RealEstate10k, we found that using the combined loss above deteriorates reconstruction metrics. Therefore, we train models on RealEsatate10k solely with the $L_1$ loss for 300K iterations.

\subsubsection{APE- and RPE-based transformers on CLEVR-TR}
\label{sec:exp_ape_rpe}
For the APE-based model, we replace the ray embeddings in SRT with a linear projection of the combined 2D positional embedding and flattened $SE(3)$ matrix. To build an RPE-based model, we follow the same procedure as in \secref{sec:gta} and apply the representations to the bias vectors appended to the QKV vectors. Each bias dimension is set to 16 for the $\sigma_{\rm cam}$ and $16$ for $\sigma_{h}$ and $\sigma_{w}$. The multiplicities and frequency parameters are determined as described in \secref{sec:rho_design}. $\{s, u, v\}$ is set to $\{4, 1, 1\}$ and $\{f\}$ is set to $\{1,...,1/2^3\}$ for both $\sigma_{h}$ and $\sigma_{w}$.
\tabref{tab:clevrtr} shows an extended version of \tabref{tab:comp_nvs_1}, which includes LPIPS~\citep{Zhang2018CVPRa} and SSIM performance.

\subsubsection{Implementation of other PE methods}
\label{sec:exp_otherenc}

\begin{figure}
    \centering
    \includegraphics[scale=0.3]{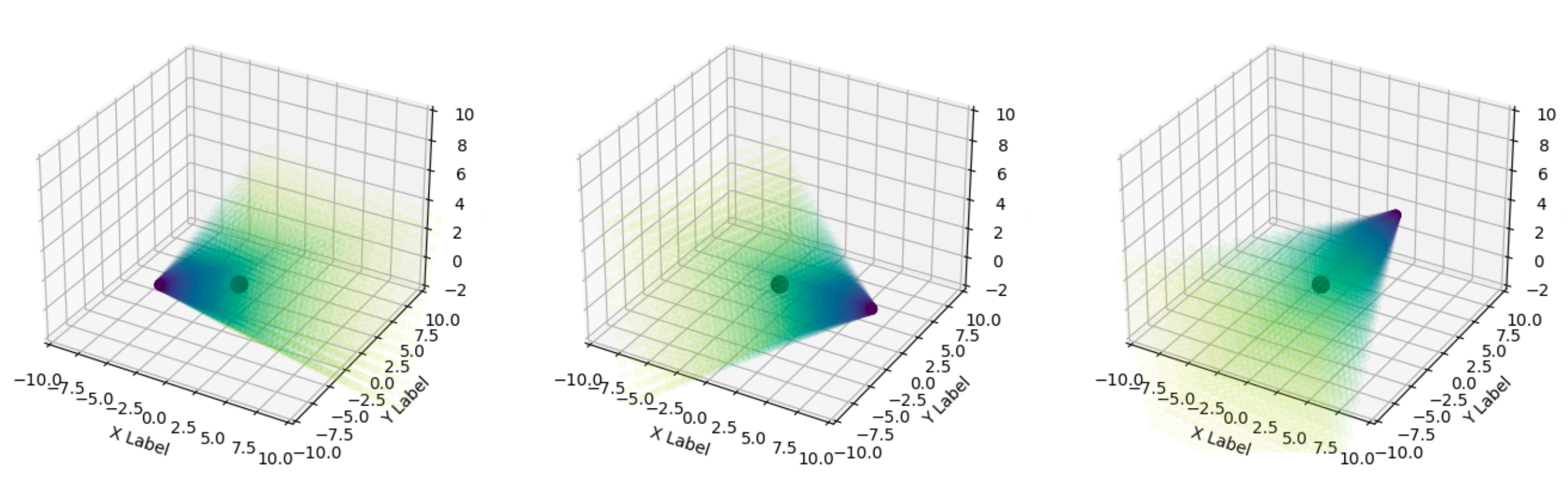}
    \caption{\textbf{Frustum points on CLEVR-TR}. The black point indicates the origin $(0,0,0)$. Each object is sampled with its center in the range of $\in [-4, 4]\times [-4, 4] \times\{t/2\}$ where $t$ is the height of the object.}
    \label{fig:frustum_points}
\end{figure}

\boldparagraph{Frustum positional embeddings~\citep{Liu2022ECCV}}
Given an intrinsic $K\in \mathbb{R}^{3\times 3}$, we transform the 2D image position of each token by 
$K^{-1} (x, y, 1)^{\rm T}$. We follow~\citet{Liu2022ECCV} and generate points at multiple depths with the linear-increasing discretization~\citep{Reading2021CVPR}, where each depth value at index $i=1,...,D$ is computed by
\begin{align}
    d_{min} + \frac{d_{max} - d_{min}}{D(D+1)} i(i+1)
\end{align}
where $[d_{min}, d_{max}]$ is the full depth range and $D$ is the number of depth bins. Examples of the generated 3D points are visualized in \figref{fig:frustum_points}.
The concatenation of 3D points of multiple depth at each pixel is further processed by a learned 1-layer MLP, and added to input.

\boldparagraph{Modulated layer normalization ~\citep{Hong2023ARXIV, Liu2023ARXIV}}
Modulated layer normalization (MLN) modulates and biases each token feature $x$ by using vector features $\gamma, \beta$ each of which encodes geometric information. In \citet{Liu2023ARXIV}, each token's geometric information is a triplet of a camera transformation, velocity, and time difference of consecutive frames. However, in our NVS tasks, the last two information does not exist. Thus, the vectors simply encode the camera transformation. Each $\gamma, \beta$ is computed by:
$\gamma = \xi_{\gamma}(vec(E^{-1})), \beta = \xi_{\beta}(vec(E^{-1}))$ where $vec$ flattens the input matrix and $\xi_{\gamma, \beta}$ are learned linear transformations.
Each token $x$ is transformed with $\gamma$ and $\beta$ as follows:
\begin{align}
    x' = \gamma \odot LN(x) + \beta
\end{align}
where $\odot$ denotes element-wise multiplication.

\boldparagraph{Geometry-biased transformers~(GBT)~\citep{Venkat2023ARXIV}}
GBT biases the attention matrix of each layer by using the ray distance. Specifically, suppose each token associates with a ray $r=(o, d)\in \mathbb{R}^3\times \mathcal{S}^2$. GBT first converts $r$ into plücker coordinate $r'=(d, m)$ where $m=o \times d$. Then the ray distance between two rays $r'^{q}=(d^q, m^q)$ and $r'^{k}=(d^k, m^k)$ linked to each query vector $q$ and key vector $k$ is computed by:
\begin{align}
    dist(r'^{q}, r'^{k}) =  \begin{cases}
      \frac{|d^q\cdot m^k + d^k\cdot m^q|}{||d_q \times d_k||_2} & \text{if $d^q \times d^k \neq 0$}\\
      \frac{\|d^q(m^q-m^k/s) + d^k\|_2}{\|d^q\|_2^2} & \text{if $d^q=s d^k$, $s \neq 0$}.
    \end{cases}  
\end{align}
The GBT's attention matrix is computed by:
\begin{align}
    {\rm softmax}(QK^{\rm T} - \gamma^2 D(Q, K)),
\end{align}
where $D(Q,K) \in \mathbb{R}^{N\times N}, D_{ij}(Q,K)= dist(r'^{Q_i}, r'^{K_j})$. $\gamma \in \mathbb{R}$ is a learned scaler parameter that controls the magnitude of the distance bias.
Following~\citet{Venkat2023ARXIV}, in addition to this bias term, we also add a Fourier positional embedding computed with the plücker coordinate representation of the ray at each patch in the encoder and at each pixel in the decoder.

\boldparagraph{Element-wise multiplication}
In this approach, for each token with a geometric attribute $g$, we first concatenate the flattened $SE(3)$ homogeneous matrix and flattened $SO(2)$ image positional representations with multiple frequencies. The number of frequencies is set to the same number as in GTA on CLEVR-TR. The concatenated flattened matrices are then linearly transformed to the same dimensional vectors as each $Q,K,V$. Then these vectors are element-wise multiplied to $Q,K,V$ and the output of $Attn$ in \eqnref{eq:rta_impl} in a similar way to GTA.

\boldparagraph{RoPE+FTL~\citep{Su2021ARXIV2, Worrall2017ICCV}}
RoPE~\citep{Su2021ARXIV2} is similar to GTA but does not use the SE(3) part (extrinsic matrices) as well as transformations on value vectors. In this approach, we remove SE(3) component from the representations. Also, we remove the transformations on the value vectors from each attention layer. As an implementation of FTL~\citep{Worrall2017ICCV}, we apply SE(3) matrices to the encoder output to get transformed features to render novel views with the decoder. 

\subsection{2D image generation with DiT~\citep{Peebles2023ICCV}}\label{sec:DiTsettings}
RoPE~\citep{Su2021ARXIV2} is a method commonly used to encode positional information in transformer models. GTA and RoPE are similar but differ in that, in GTA, group transformations are applied to the value vectors in addition to the query and key vectors, leading to improvements in our NVS tasks compared to models without this transformation. To further investigate the effectiveness of the value transformation, we conduct a 2D image generation experiment. We will describe the experimental setting in the following. We also opensource the code for this experiments in the same repository as our NVS experiments, and please refer to it for further details.

Following the experimental setup of DiT~\citep{Peebles2023ICCV}, we use a transformer-based denoising network for image generation on ImageNet~\citep{Russakovsky2015IJCV}. The image resolution is set to 256x256, and we choose the DiT-B/2 model as our baseline.
Since the original DiT model does not adopt RoPE encoding, we trained models with both RoPE and GTA positional encodings. We use the same representation matrix $\rho_g$ for both RoPE and GTA, which is written as follows:
\begin{align}
    \rho_g := \sigma_{h}({\theta_h})  \oplus \sigma_{w}({\theta_w}).
\end{align}
Here, the notation of each symbol is the same as in the main section.
The representation design of each $\sigma_{h}$ and $\sigma_{w}$ follows the original work of RoPE~\citep{Su2021ARXIV2}. 
Training of each model is conducted for 2.5M iterations  (approximately 500 epochs) with batch size of 256. We experiment with mixed-precision training (BFloat16), but observed instability when using RoPE and GTA. To address this, we adopt RMSNorm~\citep{Zhang2019NEURIPS} applied to each $Q$ and $K$ vector, with which we find that no instability is made throughout the training. We report in \tabref{tab:trnsfm_v} (Right) inception scores and FIDs with classifier-free guidance and its scale set to 1.5. We show comparisons of generated images in Section~\ref{sec:dit_samples}.

\newpage
\section{Rendered images}
\label{sec:app_rendered_images}
\begin{figure}[h]
\newcommand\size{0.49}
    \centering
    \begin{tabular}{ccc}
        \multicolumn{3}{c}{Context images} \\
        \multicolumn{3}{c}{
         \includegraphics[scale=\size]{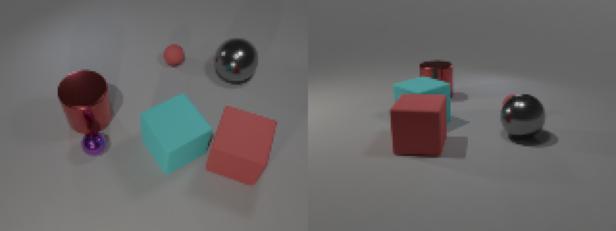}
         } \\
         APE & RPE & SRT \\
         \includegraphics[scale=\size]{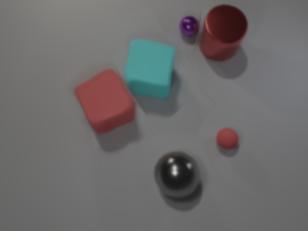} & 
         \includegraphics[scale=\size]{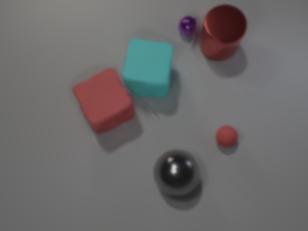} &
         \includegraphics[scale=\size]{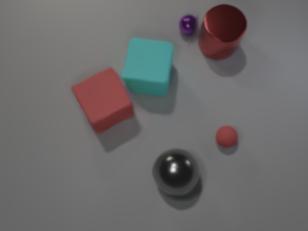} \\
         RePAST & GTA  & Ground truth \\
         \includegraphics[scale=\size]{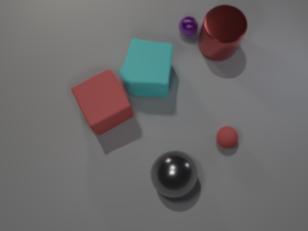} &
         \includegraphics[scale=\size]{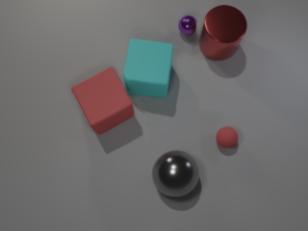} & 
         \includegraphics[scale=\size]{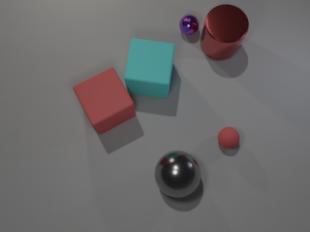} \\
        \midrule
        \multicolumn{3}{c}{Context images} \\
        \multicolumn{3}{c}{
         \includegraphics[scale=\size]{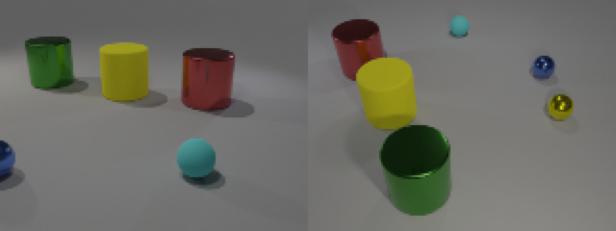}
         } \\
         APE & RPE & SRT \\
         \includegraphics[scale=\size]{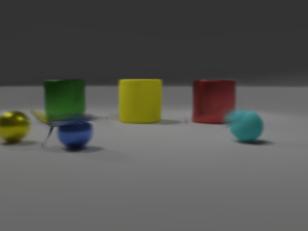} & 
         \includegraphics[scale=\size]{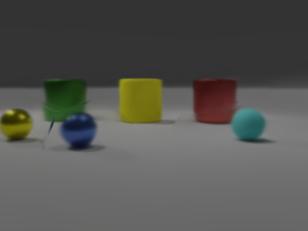} &
         \includegraphics[scale=\size]{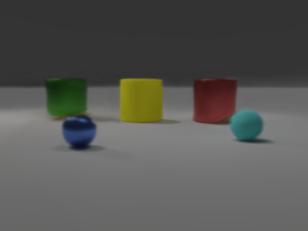} \\
         RePAST & GTA  & Ground truth \\
         \includegraphics[scale=\size]{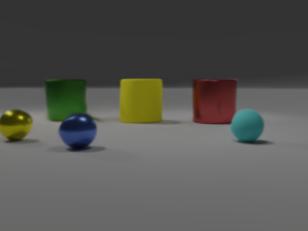} &
         \includegraphics[scale=\size]{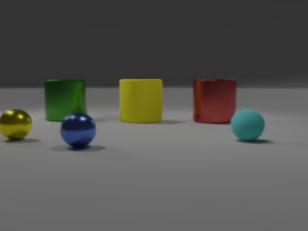} & 
         \includegraphics[scale=\size]{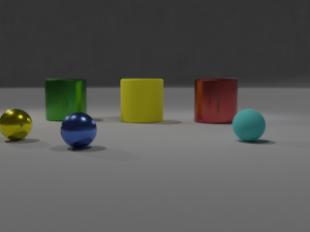} \\
    \end{tabular}
    \caption{\textbf{Qualitative results on CLEVR-TR.}\label{fig:rendered_clevrtr} }
\end{figure}

\begin{figure}[H]
    \begin{tabular}{cccc}
        
        \multicolumn{4}{c}{Context images} \\
        \multicolumn{4}{c}{\includegraphics[scale=0.44]{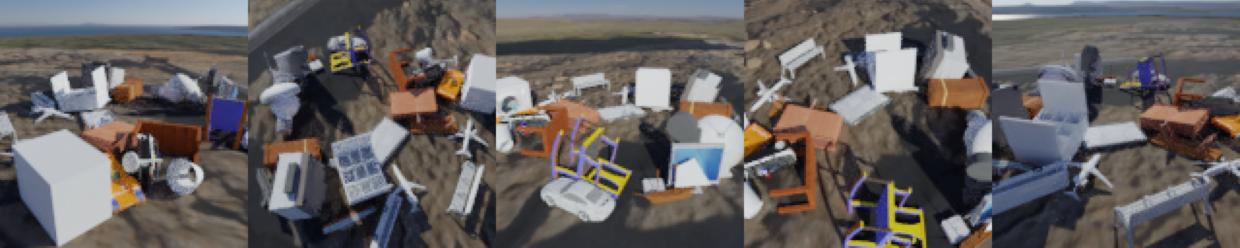}} \\
        SRT & RePAST & GTA & Ground truth  \\
         \includegraphics[scale=0.40]{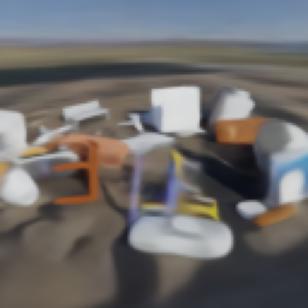} & 
         \includegraphics[scale=0.40]{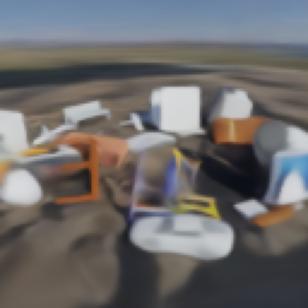} &
         \includegraphics[scale=0.40]{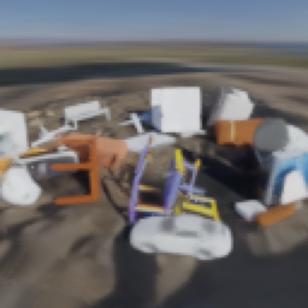} &
         \includegraphics[scale=0.40]{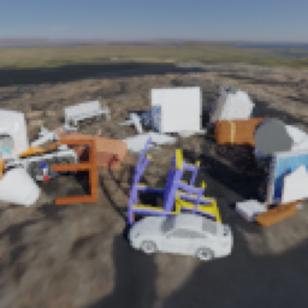} \\
         \midrule
          \multicolumn{4}{c}{Context images} \\
        \multicolumn{4}{c}{\includegraphics[scale=0.44]{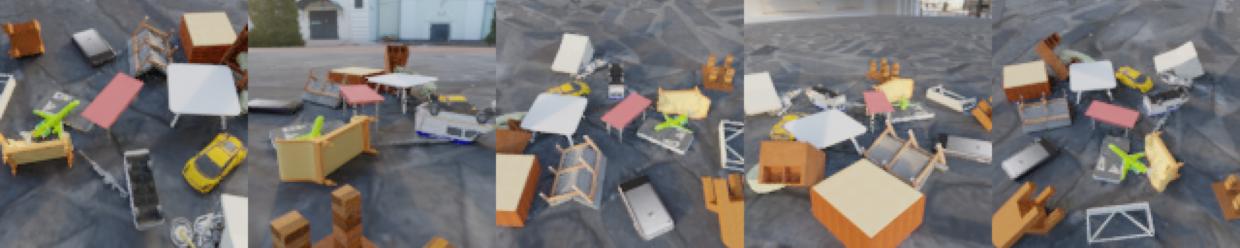}} \\
        SRT & RePAST & GTA & Ground truth  \\
         \includegraphics[scale=0.40]{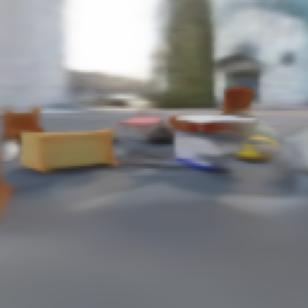} & 
         \includegraphics[scale=0.40]{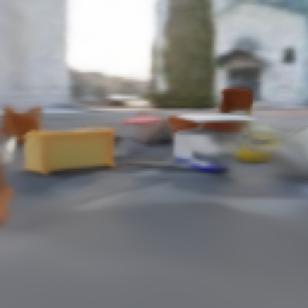} &
         \includegraphics[scale=0.40]{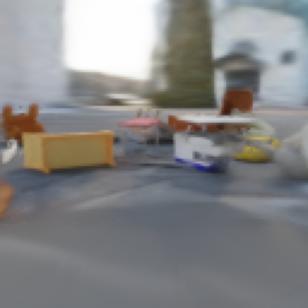} &
         \includegraphics[scale=0.40]{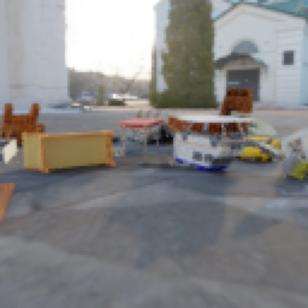} \\
        \midrule
          \multicolumn{4}{c}{Context images} \\
        \multicolumn{4}{c}{\includegraphics[scale=0.44]{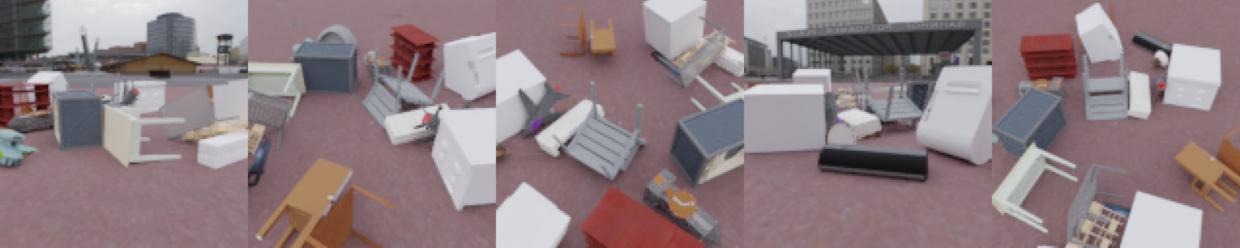}} \\
        SRT & RePAST & GTA & Ground truth  \\
         \includegraphics[scale=0.40]{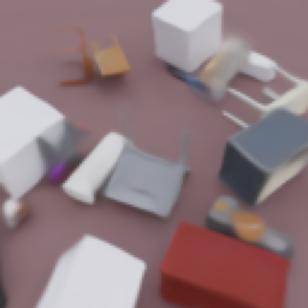} & 
         \includegraphics[scale=0.40]{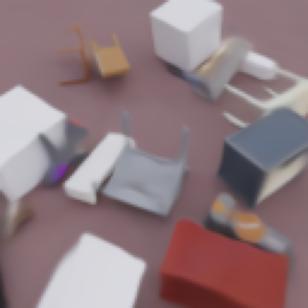} &
         \includegraphics[scale=0.40]{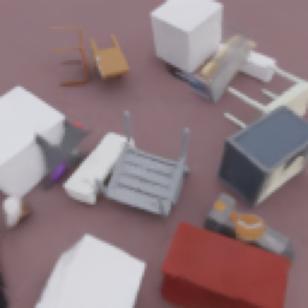} &
         \includegraphics[scale=0.40]{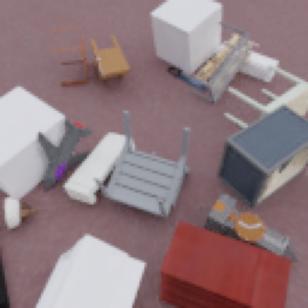} \\
       
    \end{tabular}
    \caption{\textbf{Qualitative results on MSN-Hard.}\label{fig:rendered_msn_app} }
\end{figure}

\begin{figure}[H]
    \begin{tabular}{cccc}
        Context images & \citet{Du2023CVPR} & GTA & Ground truth  \\
        \midrule
         \includegraphics[scale=0.21]{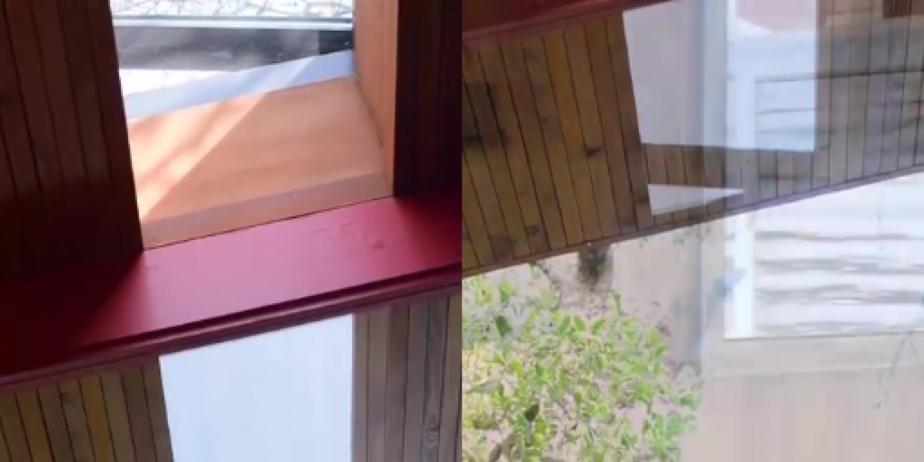} &
         \includegraphics[scale=0.21]{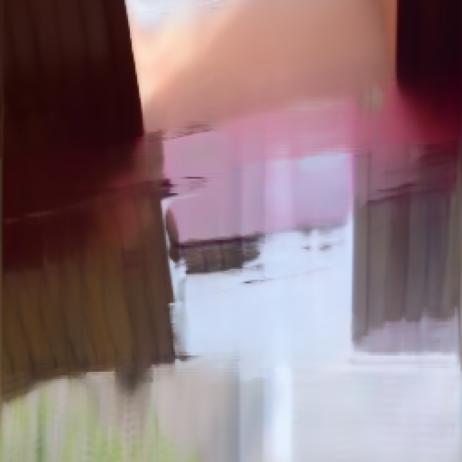} & 
         \includegraphics[scale=0.21]{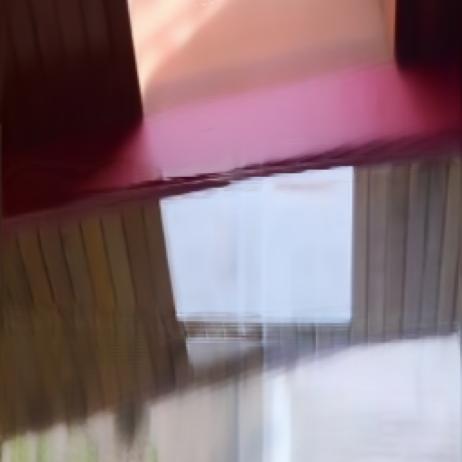} &
         \includegraphics[scale=0.21]{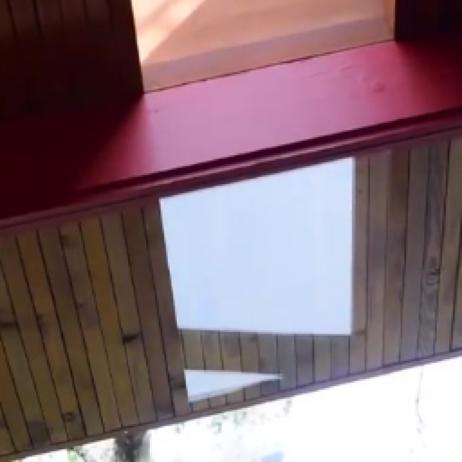} \\
         \includegraphics[scale=0.21]{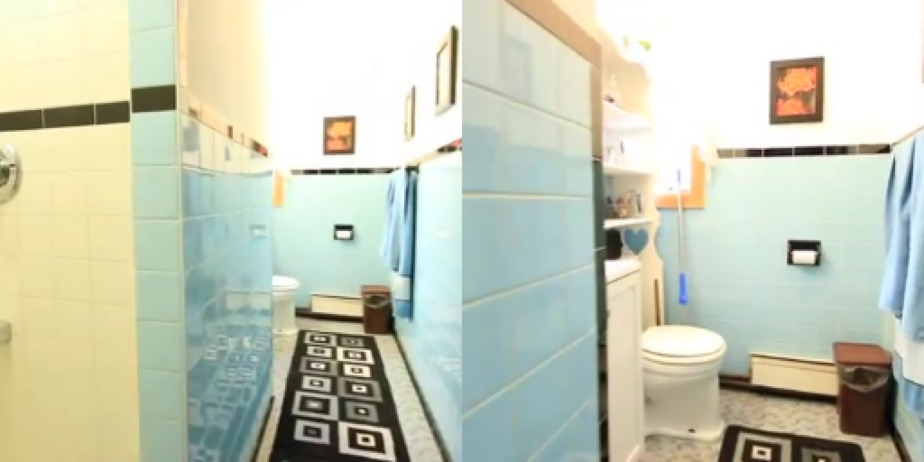} &
         \includegraphics[scale=0.21]{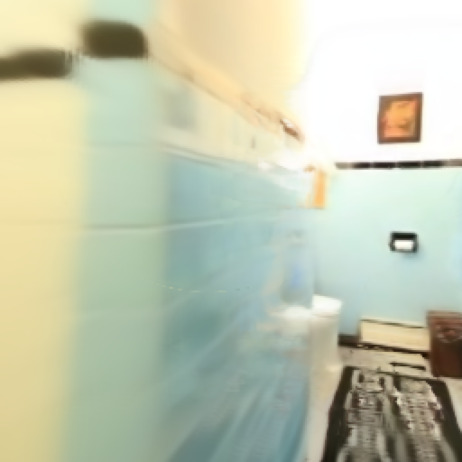} & 
         \includegraphics[scale=0.21]{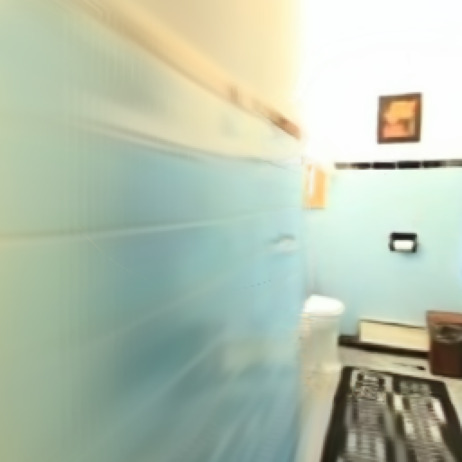} &
         \includegraphics[scale=0.21]{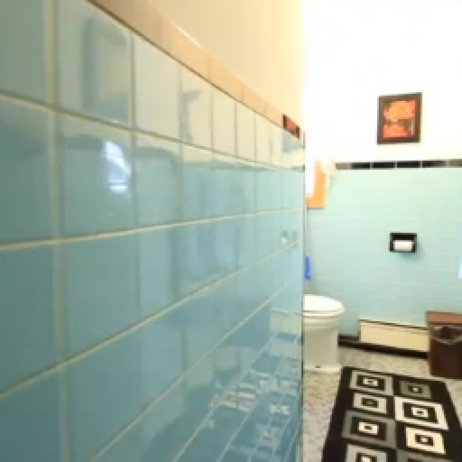} \\
        \includegraphics[scale=0.21]{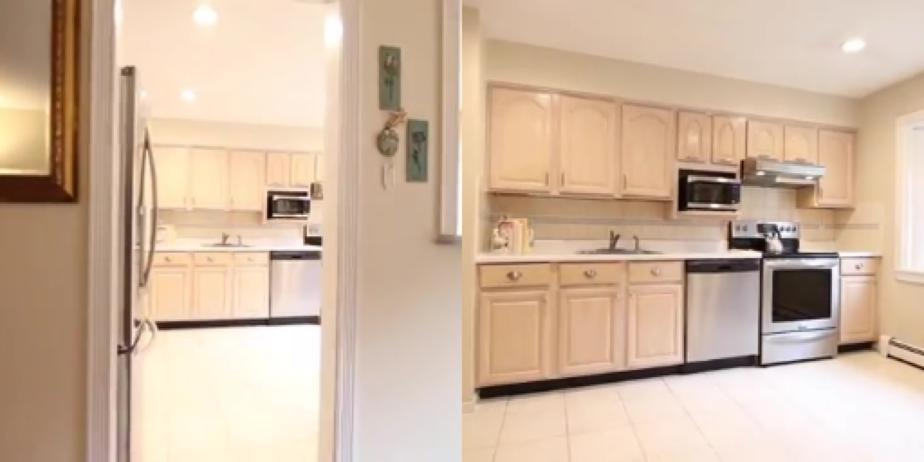} &
         \includegraphics[scale=0.21]{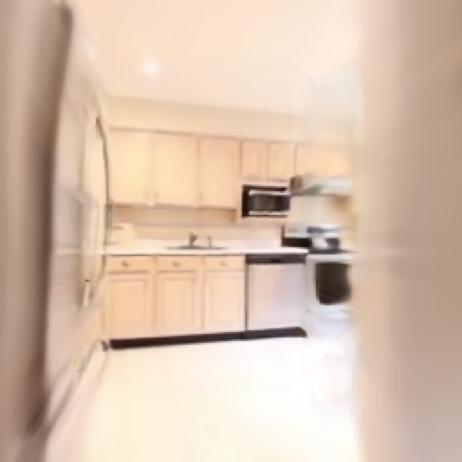} & 
         \includegraphics[scale=0.21]{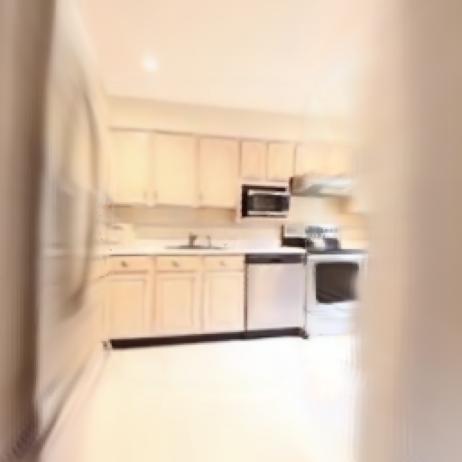} &
         \includegraphics[scale=0.21]{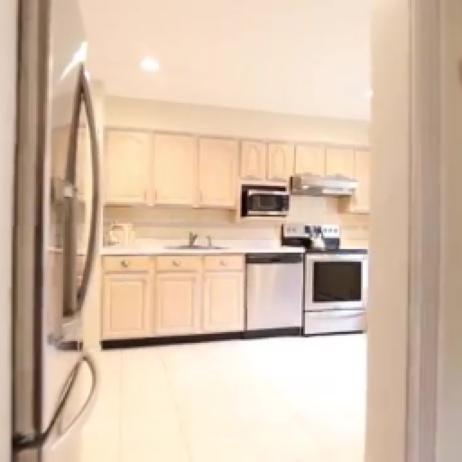} \\
        \includegraphics[scale=0.21]{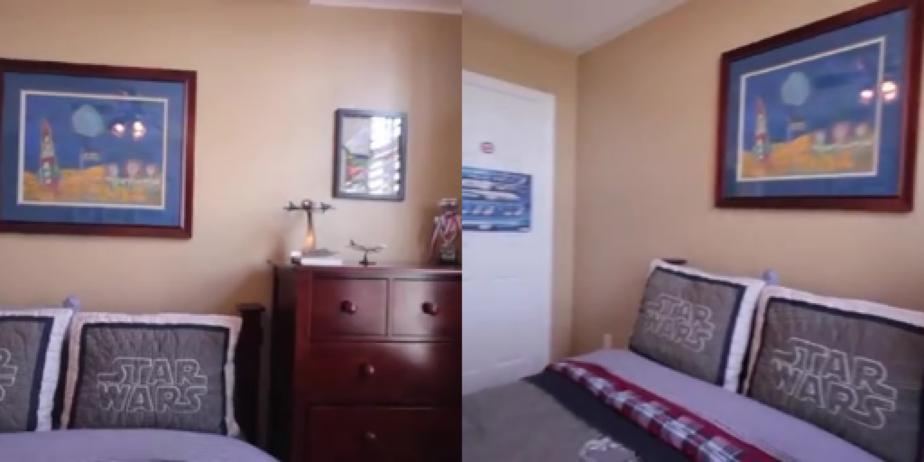} &
         \includegraphics[scale=0.21]{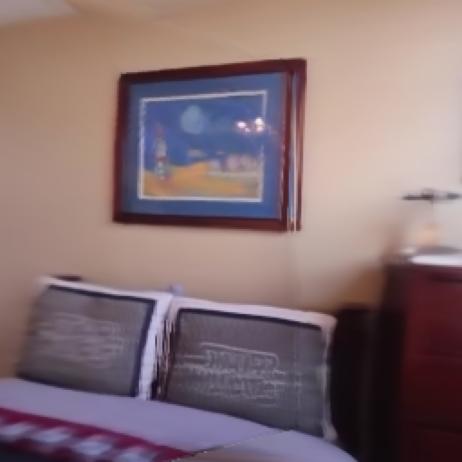} & 
         \includegraphics[scale=0.21]{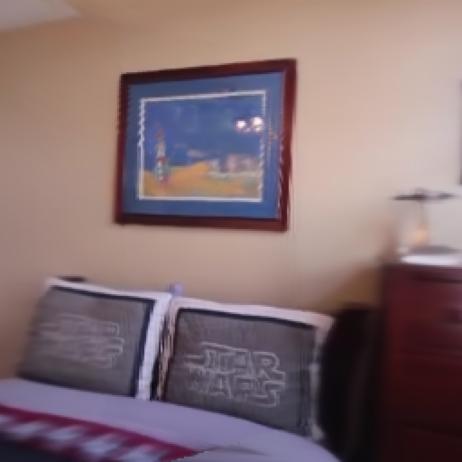} &
         \includegraphics[scale=0.21]{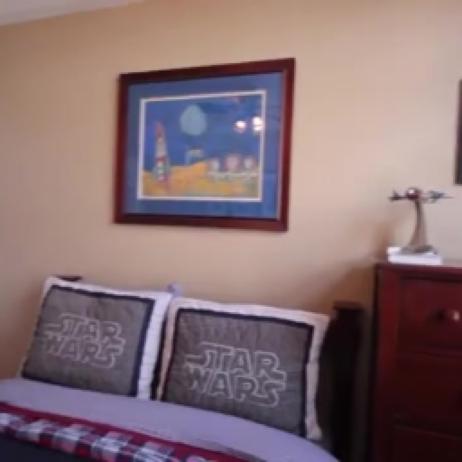} \\
        \includegraphics[scale=0.21]{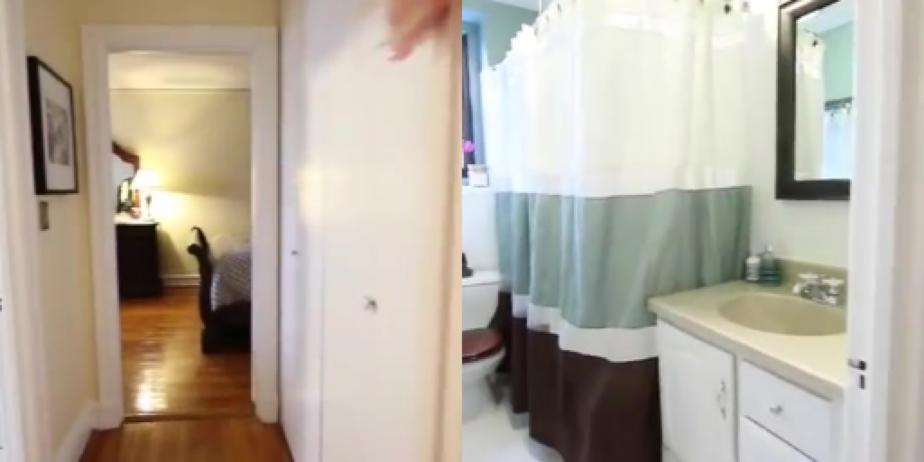} &
         \includegraphics[scale=0.21]{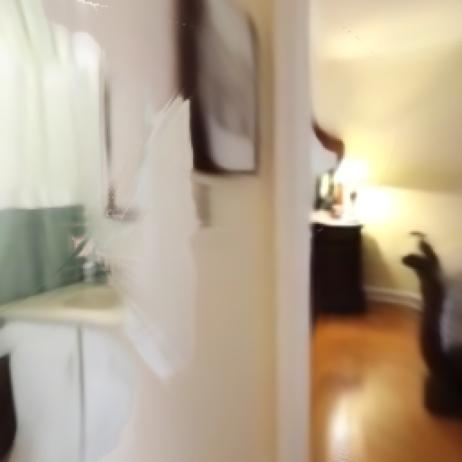} & 
         \includegraphics[scale=0.21]{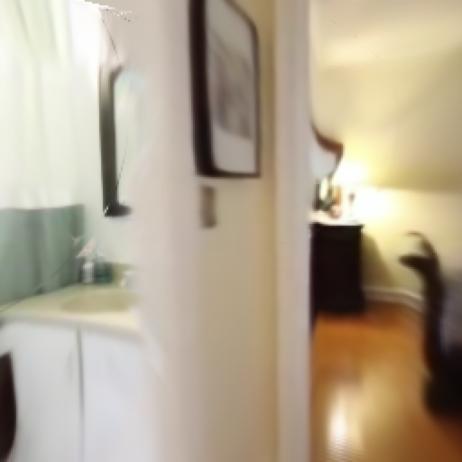} &
         \includegraphics[scale=0.21]{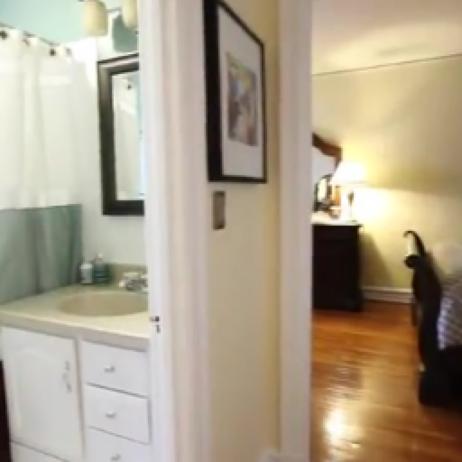} \\
    \end{tabular}
    \caption{\textbf{Qualitative results on RealEstate10k.}\label{fig:rendered_realestate} }
\end{figure}

\begin{figure}[H]
    \begin{tabular}{cccc}
        Context images & \citet{Du2023CVPR} & GTA & Ground truth  \\
        \midrule
         \includegraphics[scale=0.21]{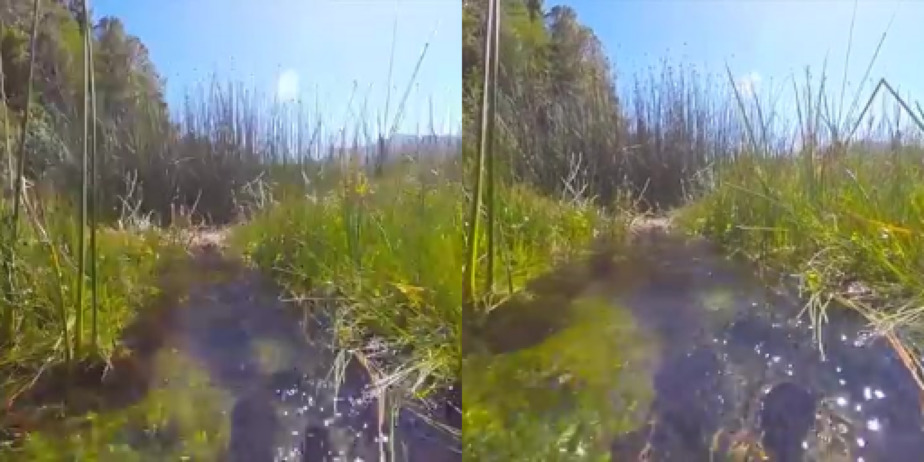} &
         \includegraphics[scale=0.21]{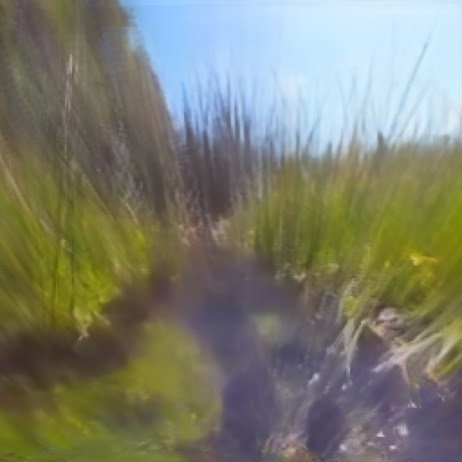} & 
         \includegraphics[scale=0.21]{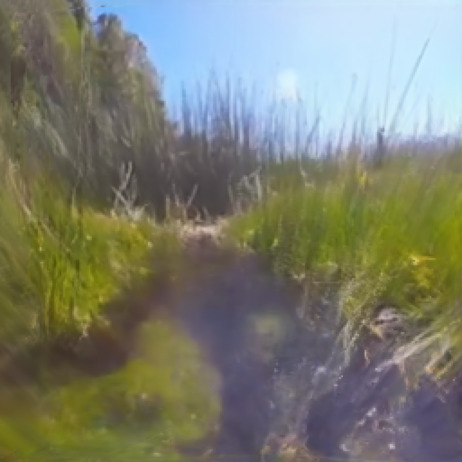} &
         \includegraphics[scale=0.21]{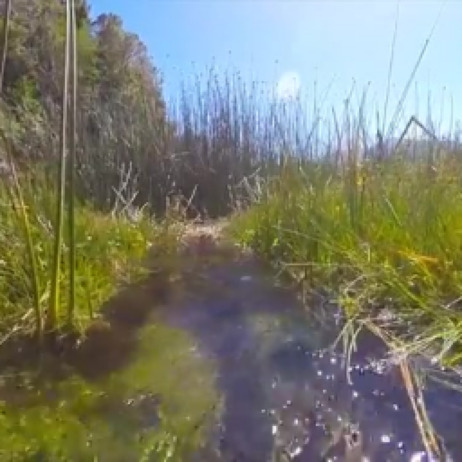} \\
         \includegraphics[scale=0.21]{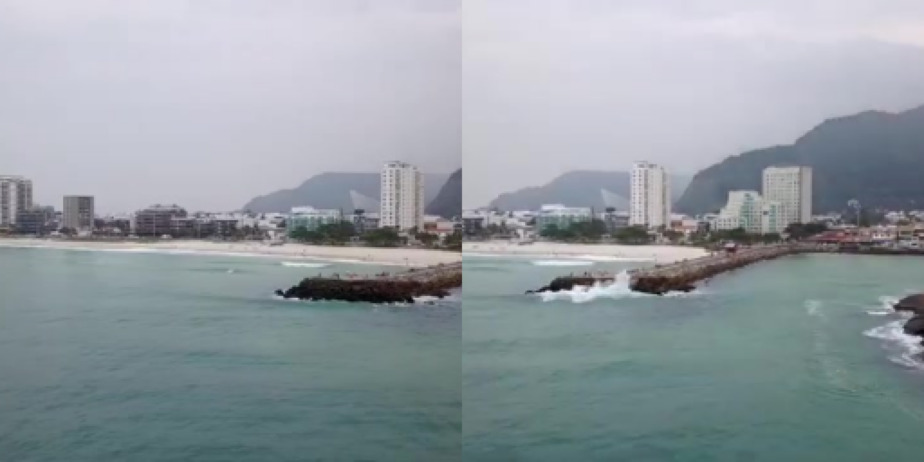} &
         \includegraphics[scale=0.21]{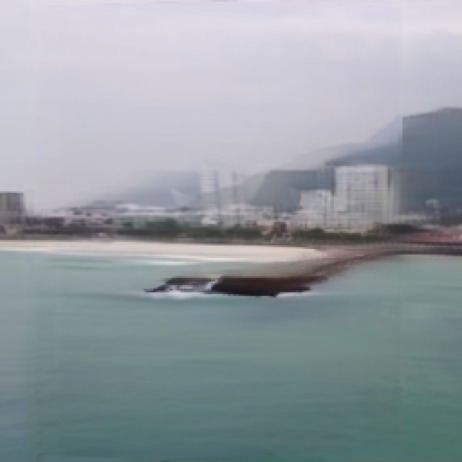} & 
         \includegraphics[scale=0.21]{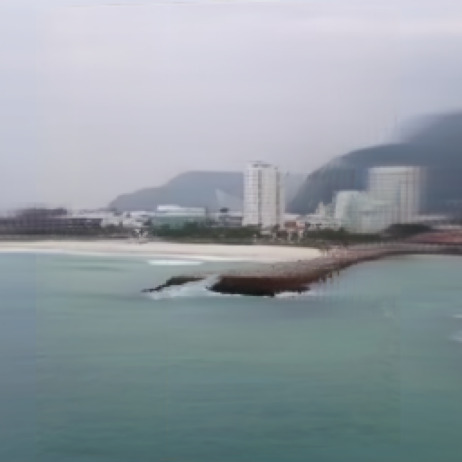} &
         \includegraphics[scale=0.21]{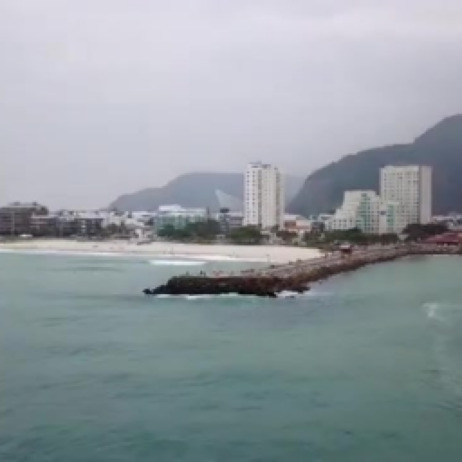} \\
        \includegraphics[scale=0.21]{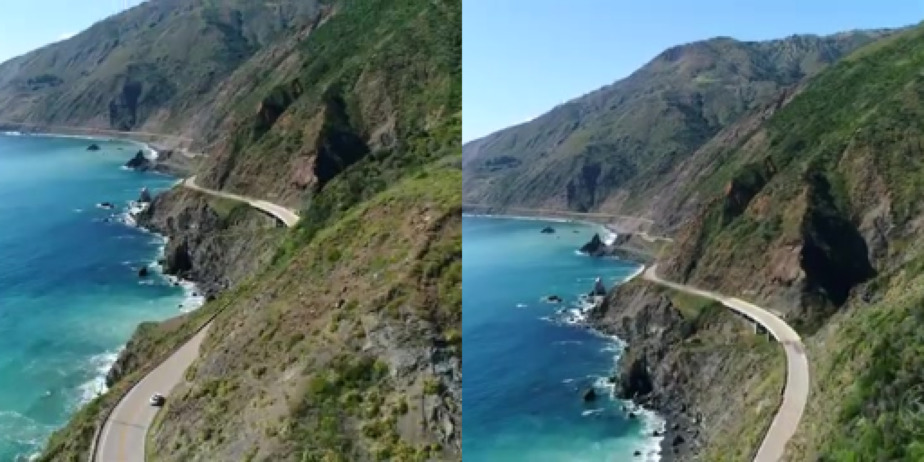} &
         \includegraphics[scale=0.21]{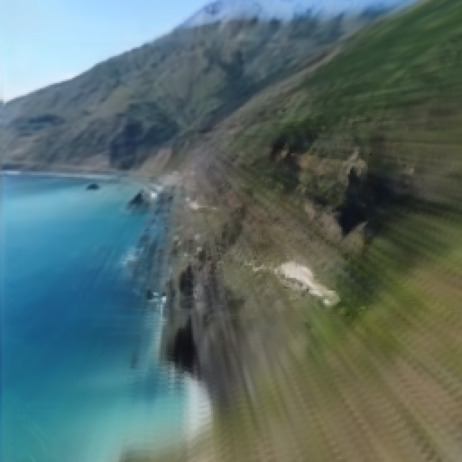} & 
         \includegraphics[scale=0.21]{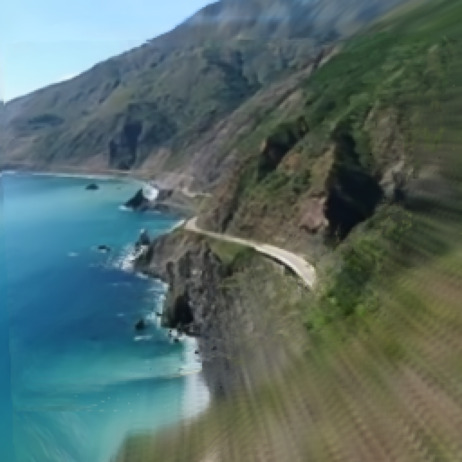} &
         \includegraphics[scale=0.21]{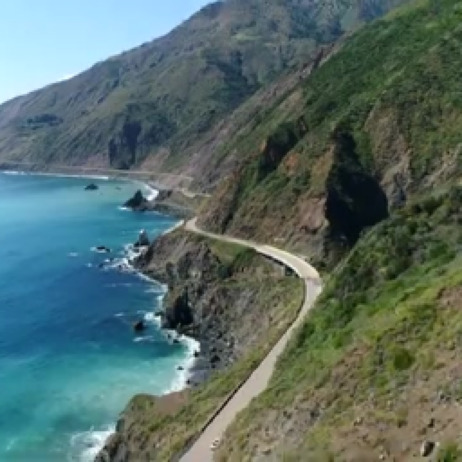} \\
        \includegraphics[scale=0.21]{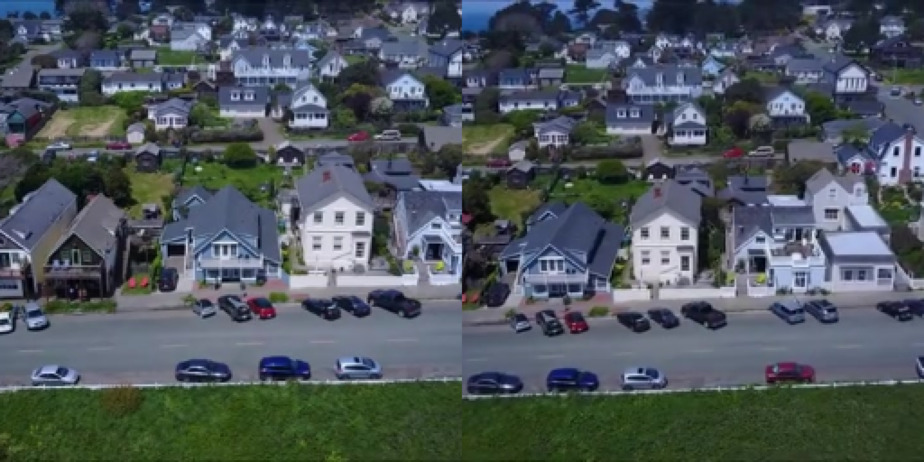} &
         \includegraphics[scale=0.21]{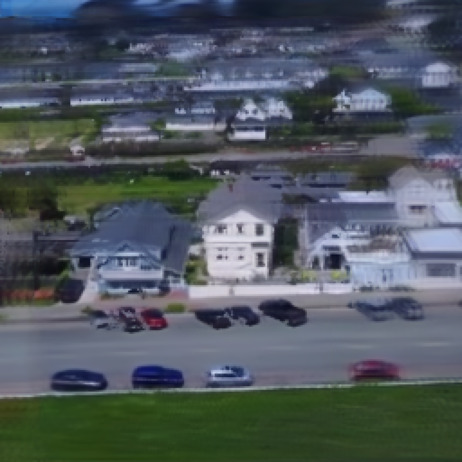} & 
         \includegraphics[scale=0.21]{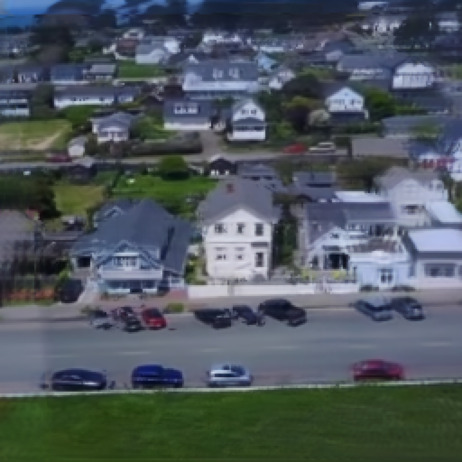} &
         \includegraphics[scale=0.21]{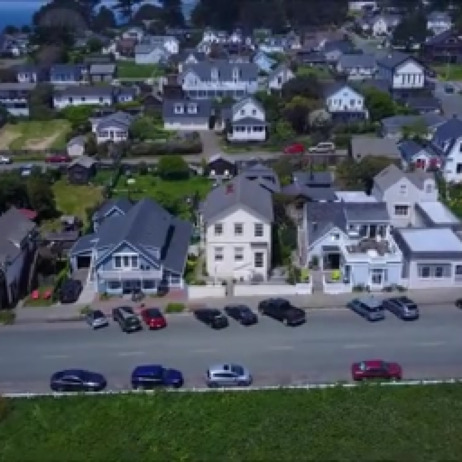} \\
        \includegraphics[scale=0.21]{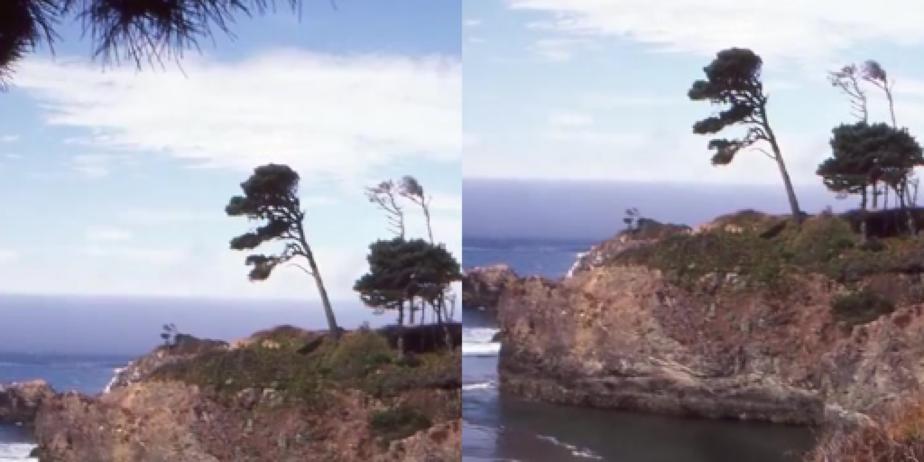} &
         \includegraphics[scale=0.21]{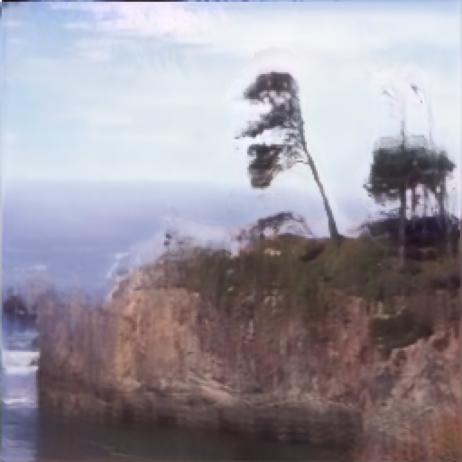} & 
         \includegraphics[scale=0.21]{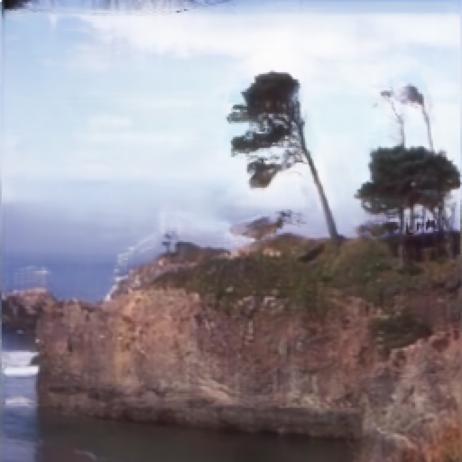} &
         \includegraphics[scale=0.21]{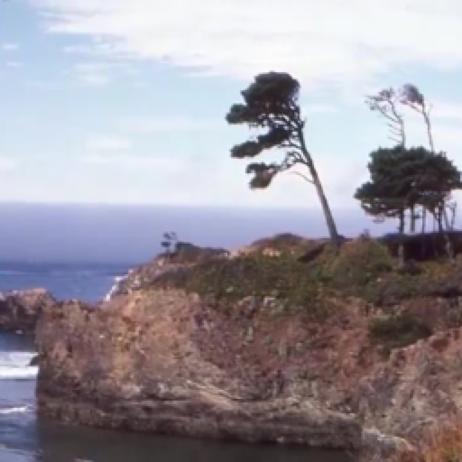} \\
    \end{tabular}
    \caption{\textbf{Qualitative results on ACID.}\label{fig:rendered_acid} }
\end{figure}

\section{Generated images of DiTs}\label{sec:dit_samples}

\begin{figure}[H]
    \centering
    \begin{tabular}{ccc}
        DiT + GTA & DiT + RoPE & DiT~\citep{Peebles2023ICCV} \\
        \midrule
         \includegraphics[scale=0.18]{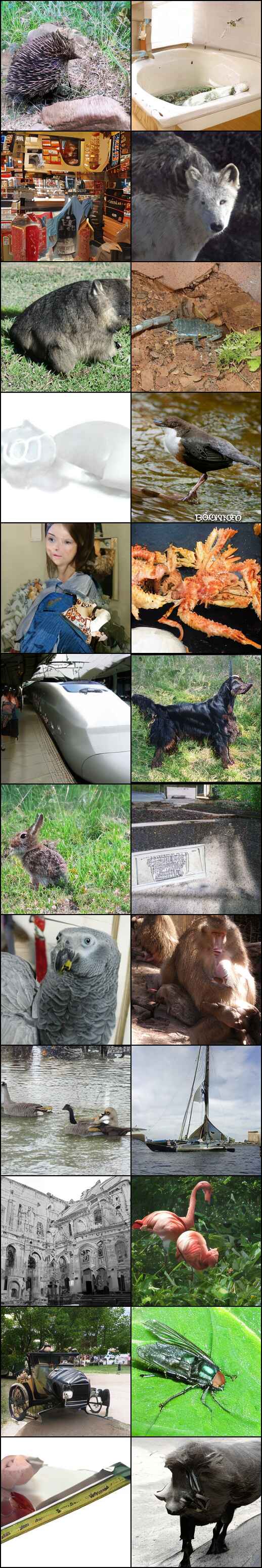} &
         \includegraphics[scale=0.18]{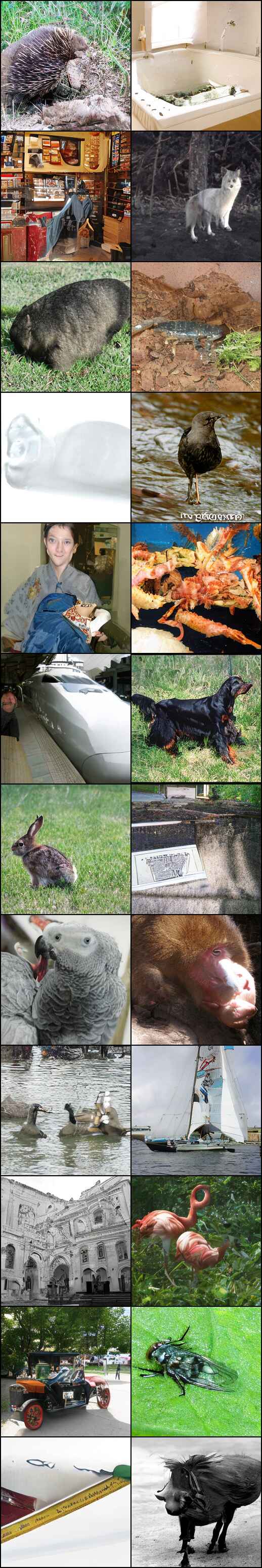} & 
         \includegraphics[scale=0.18]{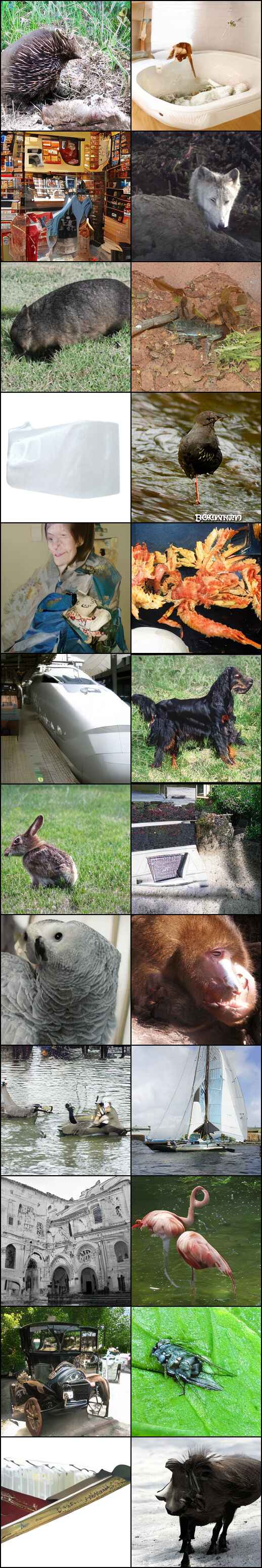} 
    \end{tabular}
    \caption{\textbf{Class-conditional generation on ImageNet}. Labels and noises are randomly sampled.\label{fig:gen_dit} }
\end{figure}

\begin{figure}[H]
    \centering
    \begin{tabular}{ccc}
        DiT + GTA & DiT + RoPE & DiT~\citep{Peebles2023ICCV} \\
        \midrule
         \includegraphics[scale=0.18]{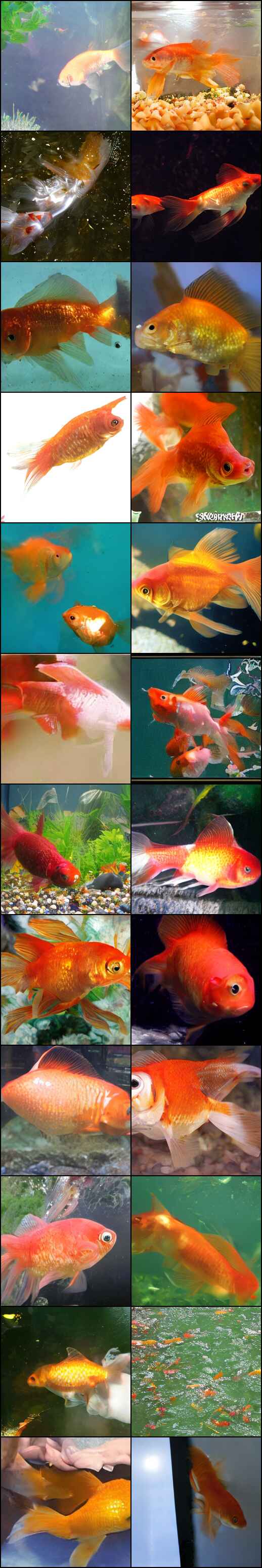} &
         \includegraphics[scale=0.18]{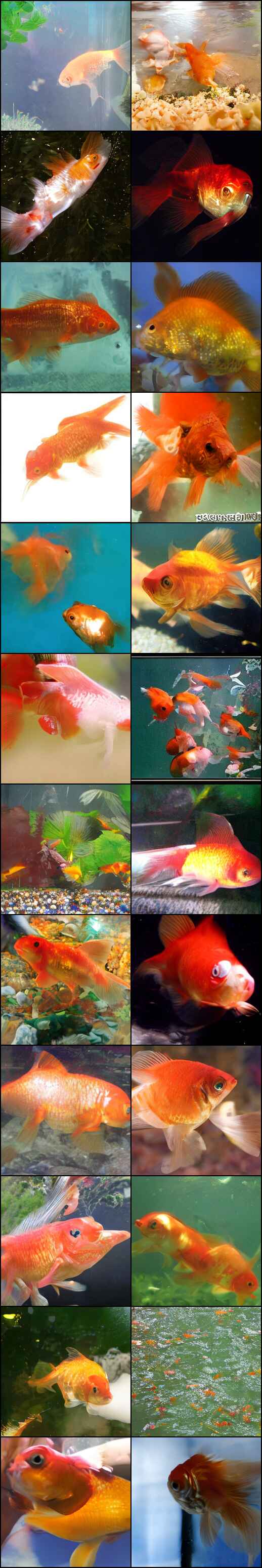} & 
         \includegraphics[scale=0.18]{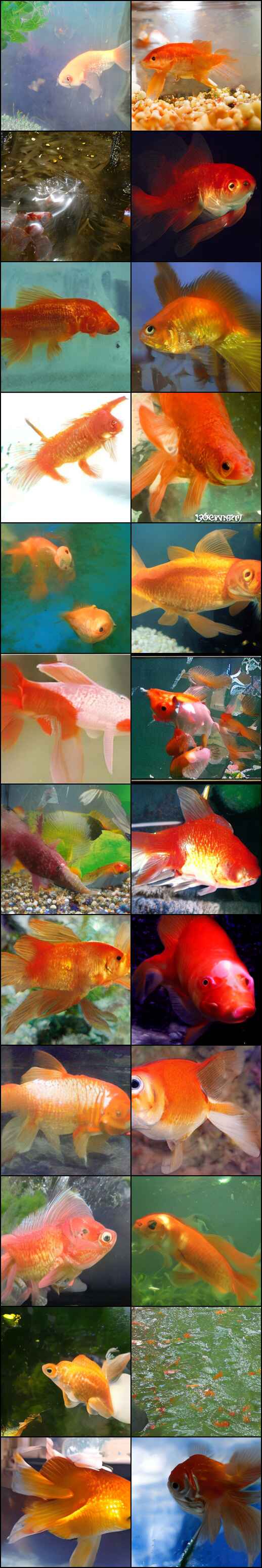}
    \end{tabular}
    \caption{\textbf{Generated images with class label `Goldfish'}\label{fig:gen_dit_1} }
\end{figure}

\begin{figure}[H]
    \centering
    \begin{tabular}{ccc}
        DiT + GTA & DiT + RoPE & DiT~\citep{Peebles2023ICCV} \\
        \midrule
         \includegraphics[scale=0.18]{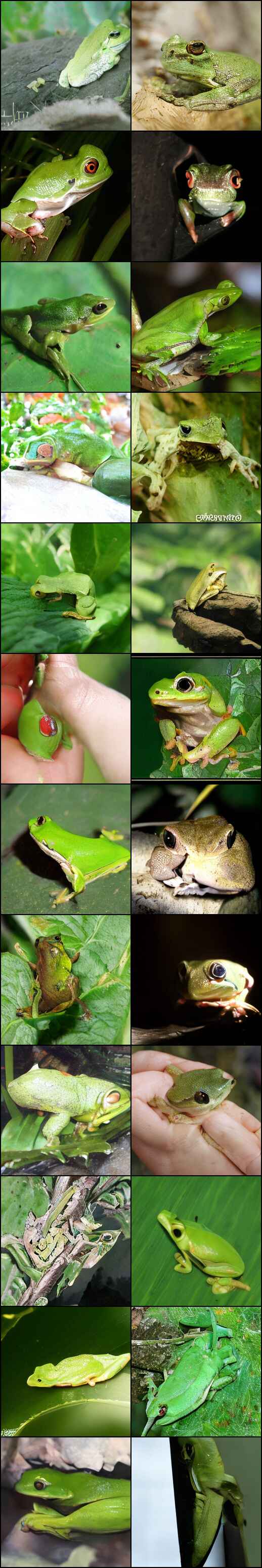} &
         \includegraphics[scale=0.18]{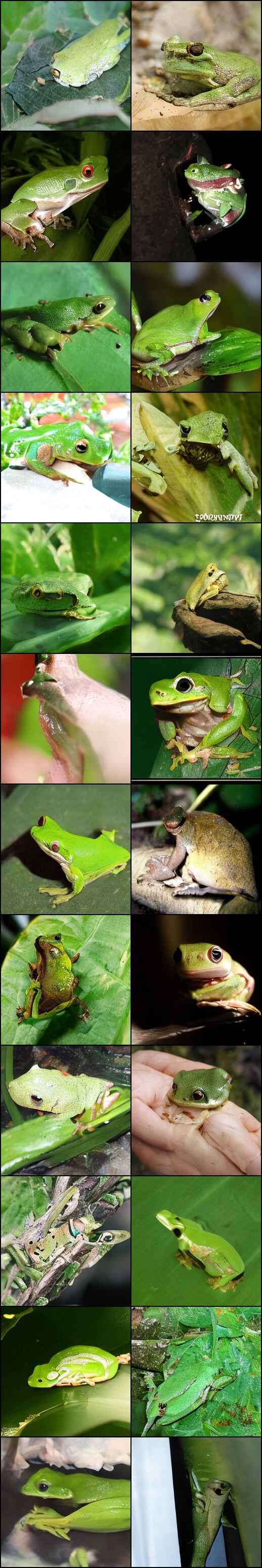} & 
         \includegraphics[scale=0.18]{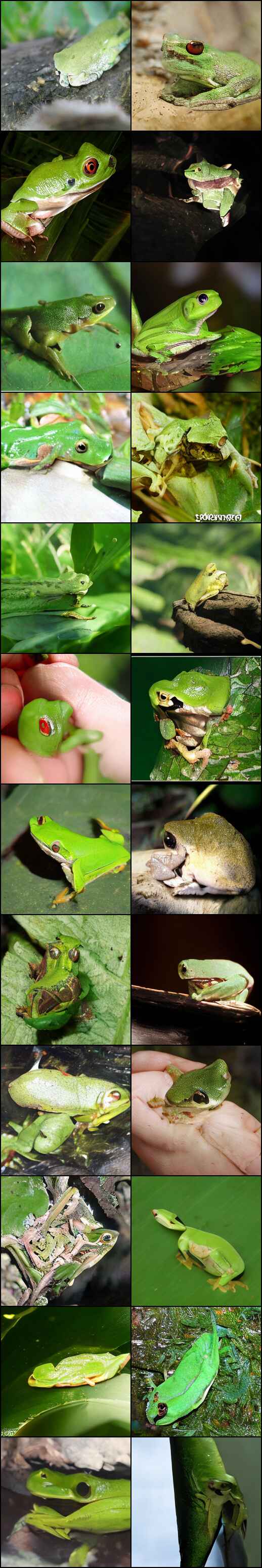}
    \end{tabular}
    \caption{\textbf{Generated images with class label `Tree frog'}\label{fig:gen_dit_31} }
\end{figure}

\begin{figure}[H]
    \centering
    \begin{tabular}{ccc}
        DiT + GTA & DiT + RoPE & DiT~\citep{Peebles2023ICCV} \\
        \midrule
         \includegraphics[scale=0.18]{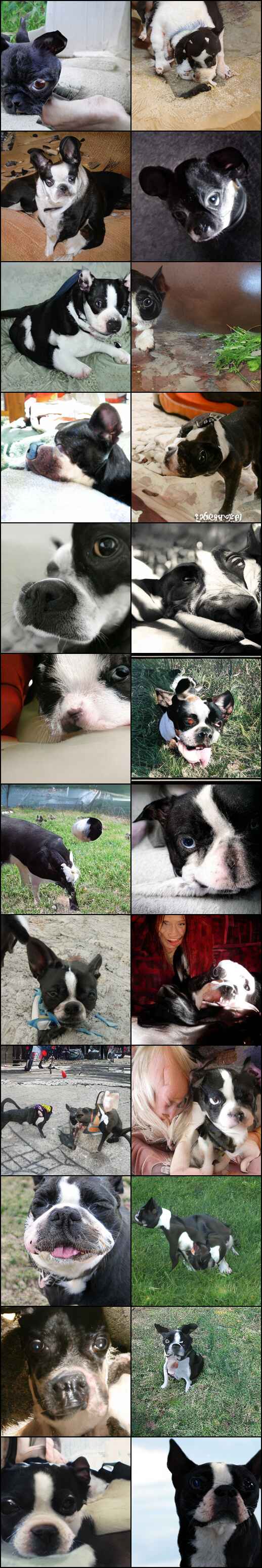} &
         \includegraphics[scale=0.18]{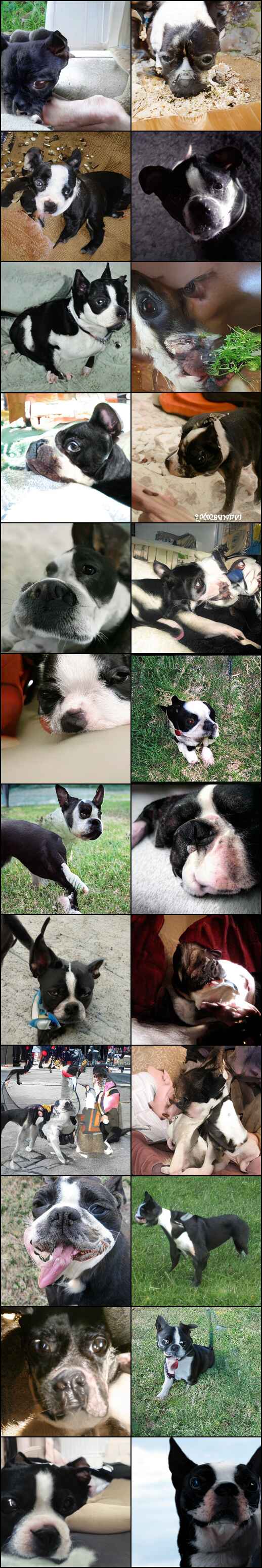} & 
         \includegraphics[scale=0.18]{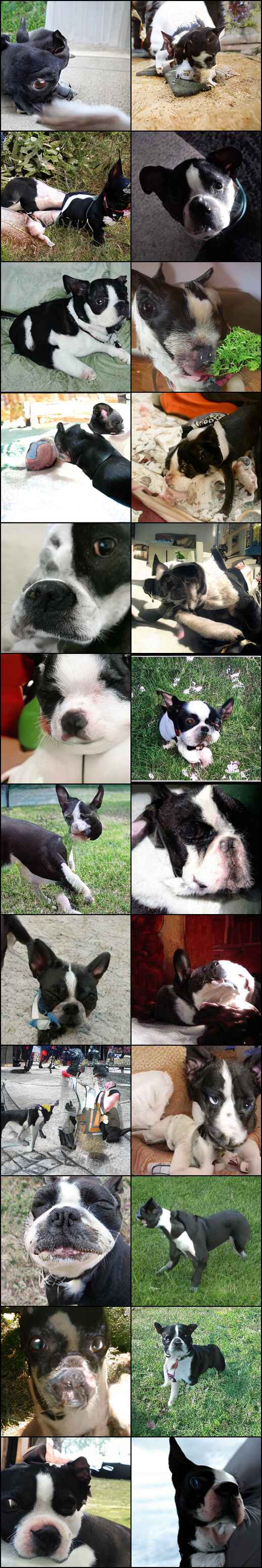}
    \end{tabular}
    \caption{\textbf{Generated images with label `Boston bull'}\label{fig:gen_dit_195} }
\end{figure}

\begin{figure}[H]
    \centering
    \begin{tabular}{ccc}
        DiT + GTA & DiT + RoPE & DiT~\citep{Peebles2023ICCV} \\
        \midrule
         \includegraphics[scale=0.18]{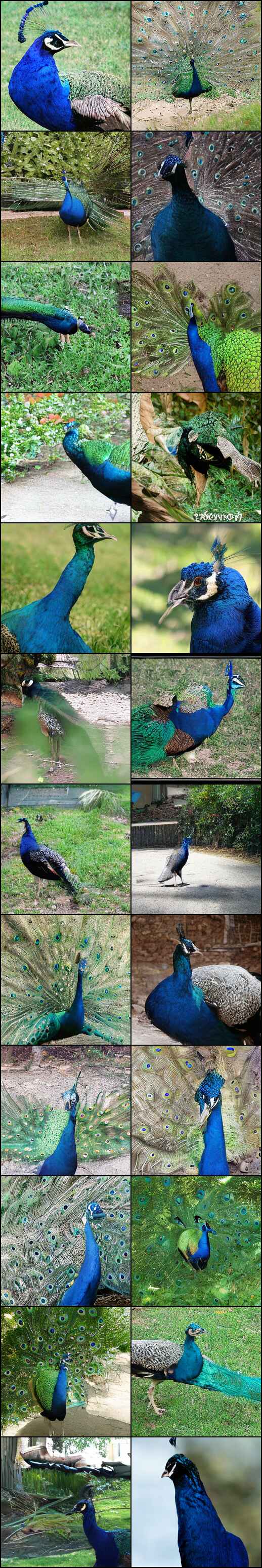} &
         \includegraphics[scale=0.18]{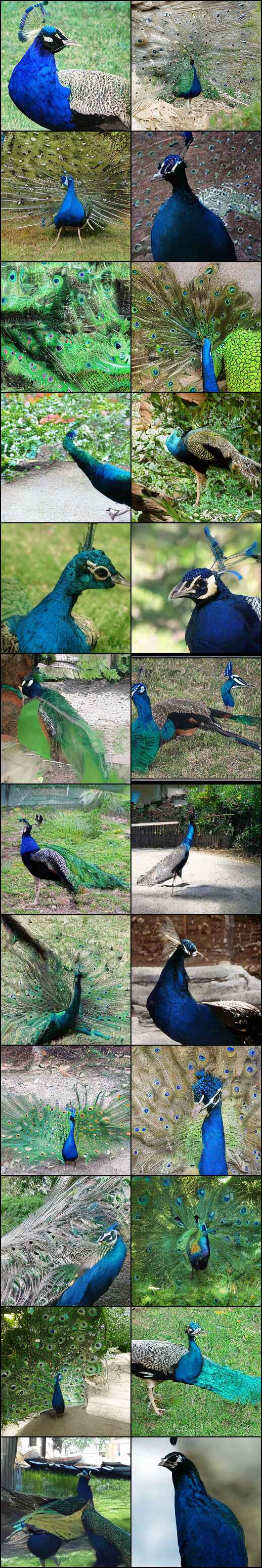} & 
         \includegraphics[scale=0.18]{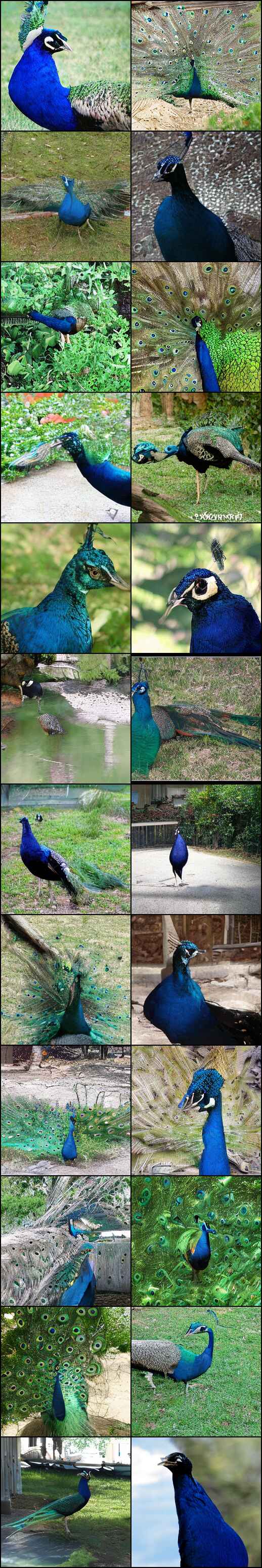}
    \end{tabular}
    \caption{\textbf{Generated images with label `Peacock'}\label{fig:gen_dit_84} }
\end{figure}
\end{document}